\documentclass{article}

\usepackage{arxiv}

\usepackage[utf8]{inputenc} 
\usepackage[T1]{fontenc}    
\usepackage{hyperref}       
\usepackage{url}            
\usepackage{booktabs}       
\usepackage{amsfonts}       
\usepackage{nicefrac}       
\usepackage{microtype}      
\usepackage{lipsum}		
\usepackage{graphicx}
\usepackage{natbib}
\usepackage{doi}

\usepackage{subcaption}
\usepackage{amsmath}
\usepackage{xcolor}
\usepackage{color}
\usepackage{verbatim}
\usepackage{mathabx}
\usepackage{tikz}
\usepackage{multirow}
\usepackage{floatrow} 

\usepackage{pifont}

\usepackage[noend]{algpseudocode}
\usepackage{algorithm,algorithmicx}  

\DeclareMathOperator*{\argmin}{argmin} 

\usepackage[toc,page]{appendix}

\newcommand{\primkey}[1]{k_{#1}^p}
\newcommand{\seckey}[1]{k_{#1}^s}
\newcommand{\pairkeys}[1]{\langle k_{#1}^p, k_{#1}^s \rangle}

\algdef{SE}[SUBALG]{Indent}{EndIndent}{}{\algorithmicend\ }%
\algtext*{Indent}
\algtext*{EndIndent}

\DeclareMathOperator*{\argmax}{argmax}

\begin{document}

\title{Dory: Overcoming Barriers to Computing Persistent Homology}

\author{Manu Aggarwal\thanks{Corresponding author}\\
       \texttt{manu.aggarwal@nih.gov}  \\
       Laboratory of Biological Modeling, NIDDK\\
       National Insitutes of Health\\
       31 Center Dr, Bethesda, MD 20892
       \AND
       Vipul Periwal\\
       \texttt{vipulp@niddk.nih.gov} \\
       Laboratory of Biological Modeling, NIDDK\\
       National Insitutes of Health\\
       31 Center Dr, Bethesda, MD 20892
       }

\maketitle

\begin{abstract}

    Persistent homology (PH) is  an approach to topological data analysis (TDA) that computes
    multi-scale topologically invariant properties of high-dimensional data that are robust to
    noise. While PH has revealed useful patterns across various applications, computational
    requirements have limited applications to small data sets of a few thousand points. We present
    Dory, an efficient and scalable algorithm that can compute the persistent homology of large
    data sets. Dory uses significantly less memory than published algorithms and also provides
    significant reductions in the computation time compared to most algorithms. It scales to process
    data sets with millions of points. As an application, we compute the PH of the human genome at
    high resolution as revealed by a genome-wide Hi-C data set. Results show that the topology of
    the human genome changes significantly upon treatment with auxin, a molecule that degrades
    cohesin, corroborating the hypothesis that cohesin plays a crucial role in loop formation in
    DNA.

\end{abstract}


\section{Introduction}

The ever increasing availability of scientific data necessitates development of mathematical
algorithms and computational tools that yield testable predictions or give mechanistic insights into
model systems underlying the data. The utility of these algorithms is determined by the validity of
their theoretical foundations, the generality of their applications, their ability to deal with
noisy, high-dimensional, and incomplete data, and the computational scalability. Persistent homology
(PH), a mathematically rigorous approach to topological data analysis (TDA), finds patterns in
high-dimensional data that are robust to noise, providing a multi-scale overview of the topology of
the data. 

\begin{figure}[tbhp]
  \centering
\begin{subfigure}{.48\textwidth}
  \centering
  \includegraphics[width=\linewidth]{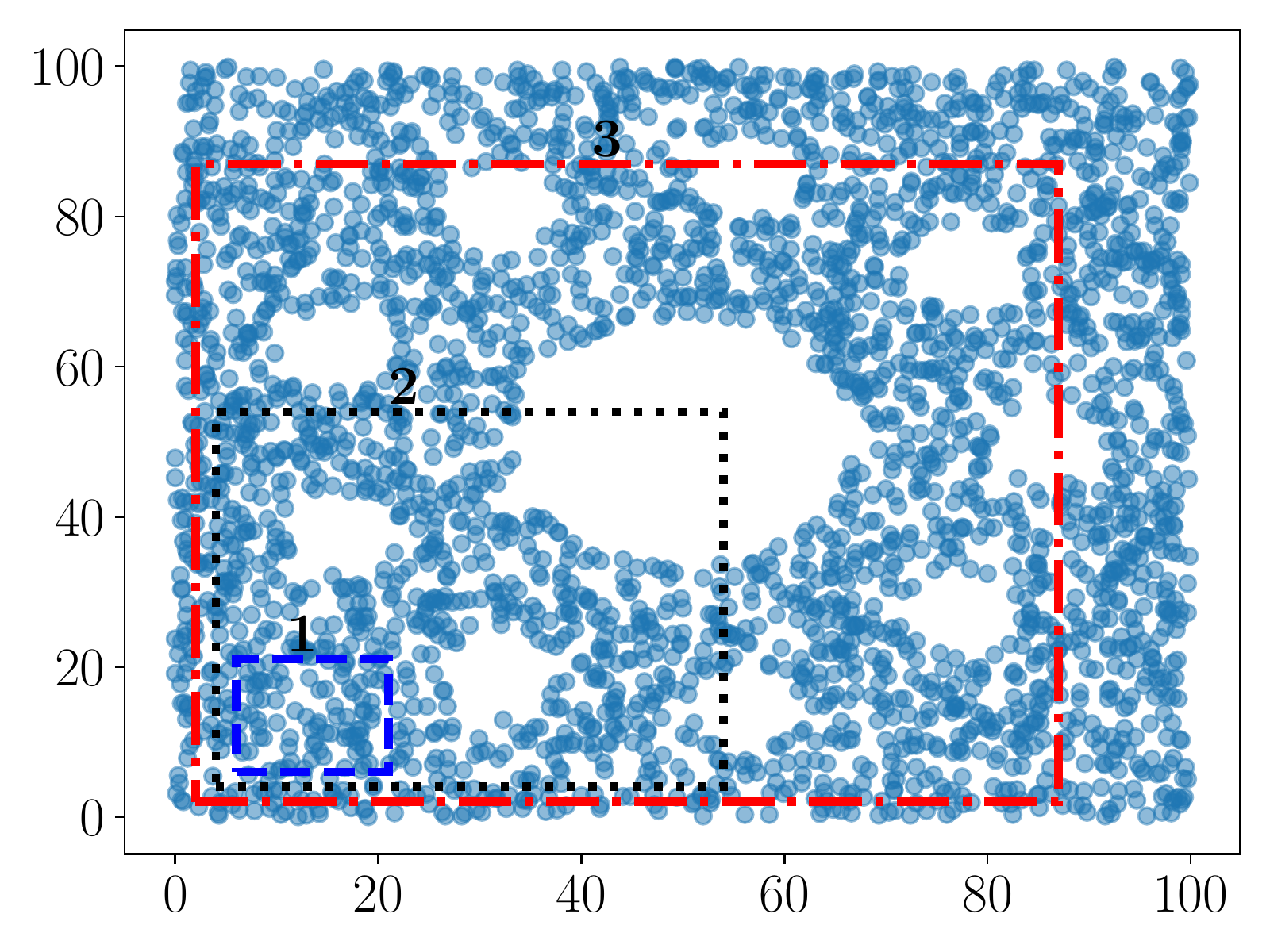}  
  \caption{We zoom in at three different scales---rectangle 1, 2, and 3.}
  \label{fig:rand3000_scatter}
\end{subfigure}
  \centering
\begin{subfigure}{.48\textwidth}
  \centering
  \includegraphics[width=\linewidth]{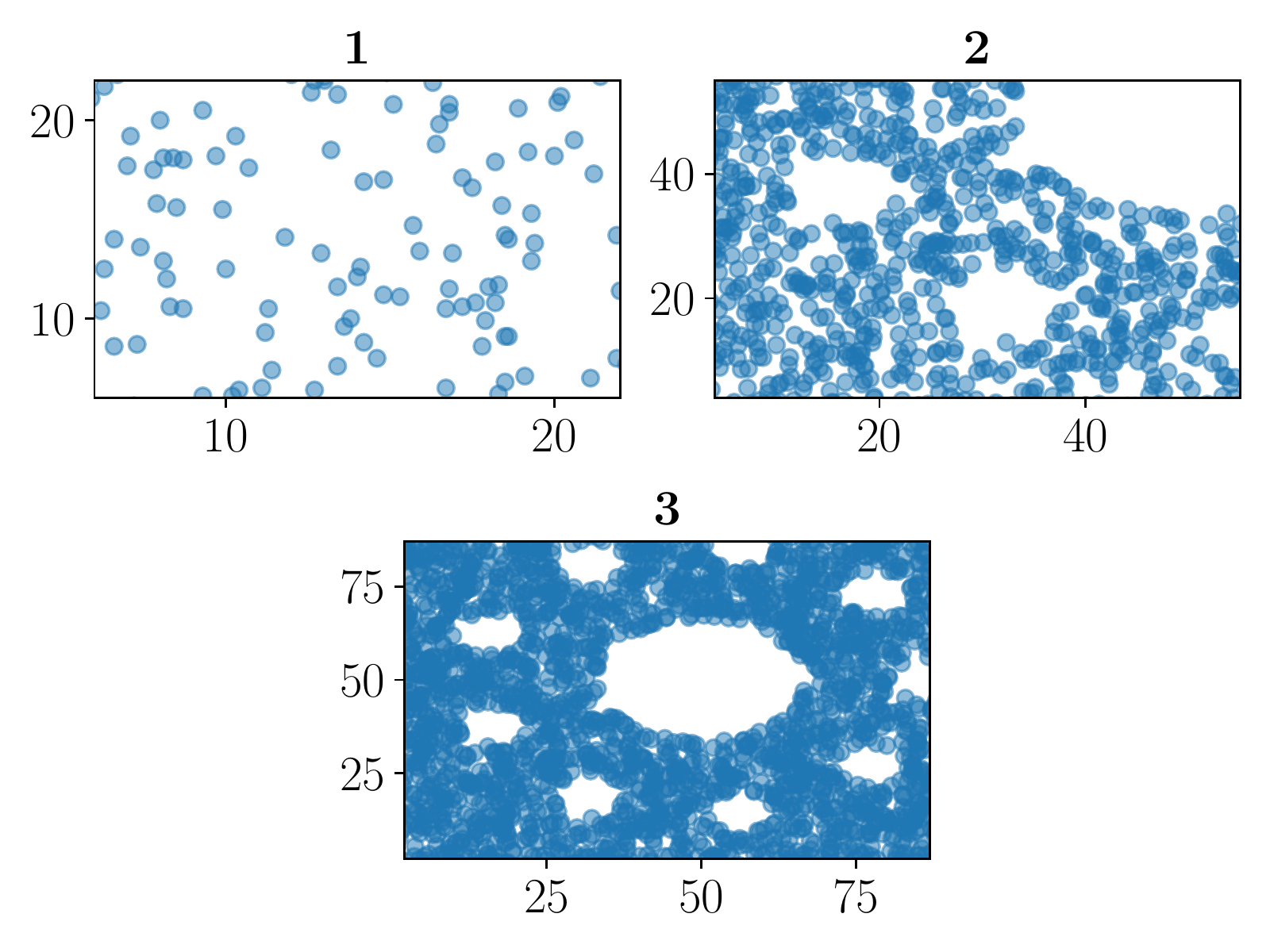}  
  \caption{New patterns emerge at different scales.}
  \label{fig:rand3000_scales}
\end{subfigure}
  \caption{A simulated data set with 3000 points.}
\label{fig:rand3000}
\end{figure}

For example, consider a point-cloud data set of 3000 points (Figure~\ref{fig:rand3000_scatter}). The
three rectangles (Figure~\ref{fig:rand3000_scales}) show different scales of observation of the
data. At the small spatial scale (rectangle 1), we do not see a discernible pattern, at a larger
scale (rectangle 2), two holes with a distinct pattern appear, and increasing the scale further
reveals a third distinct pattern, the large hole at the center (rectangle 3). For this data set, PH
will compute that there are three groups of topologically distinct features. It will also indicate
the scale at which they emerge. However, this analysis comes at a high computational cost that has
limited the applicability of PH to very small data sets. To introduce these extant computational
limitations, we briefly introduce some terminology. A general and detailed exposition on persistent
homology can be found in~\citet{edelsbrunner2008persistent}.

\begin{figure}[h]
\begin{subfigure}{0.48\textwidth}
  \centering
  \includegraphics[width=0.9\textwidth]{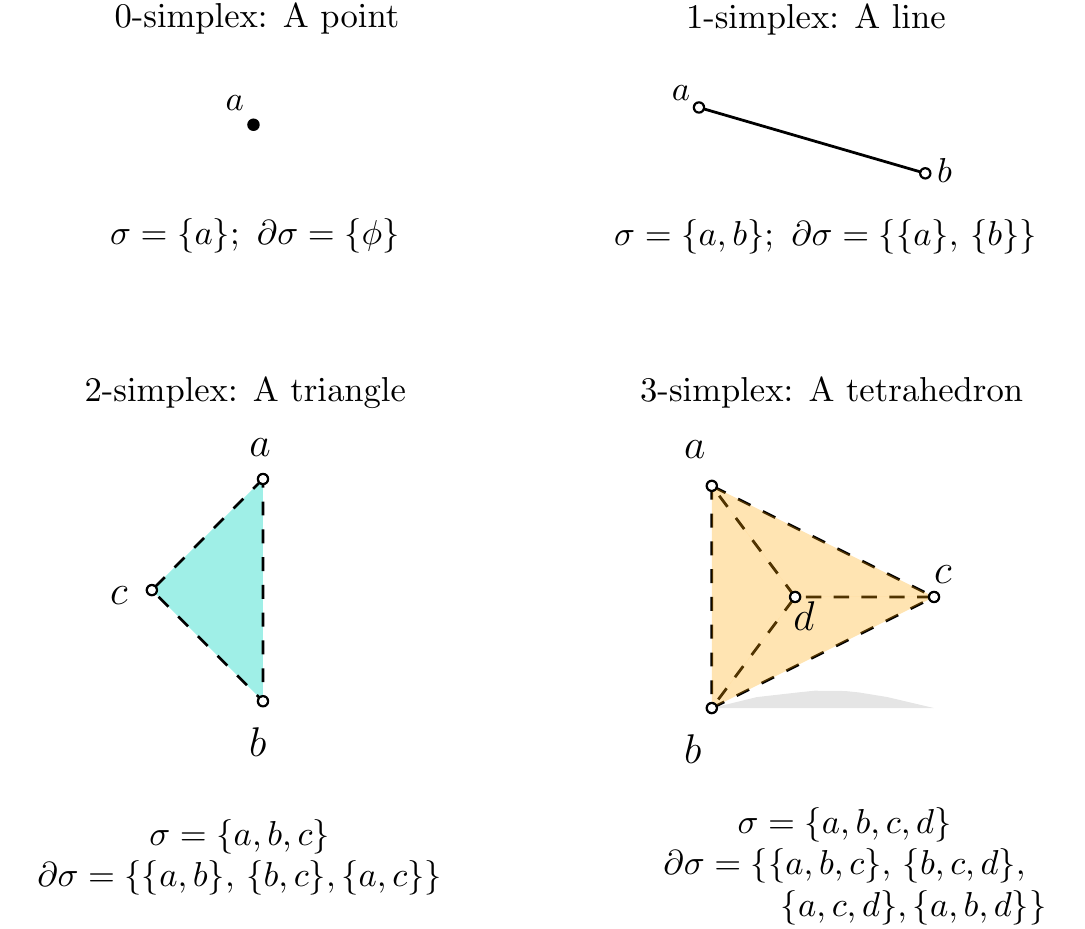}  
  \caption{Simplices}
  \label{fig:simplices}
\end{subfigure}
\begin{subfigure}{.48\textwidth}
  \centering
  \includegraphics[width=0.9\textwidth]{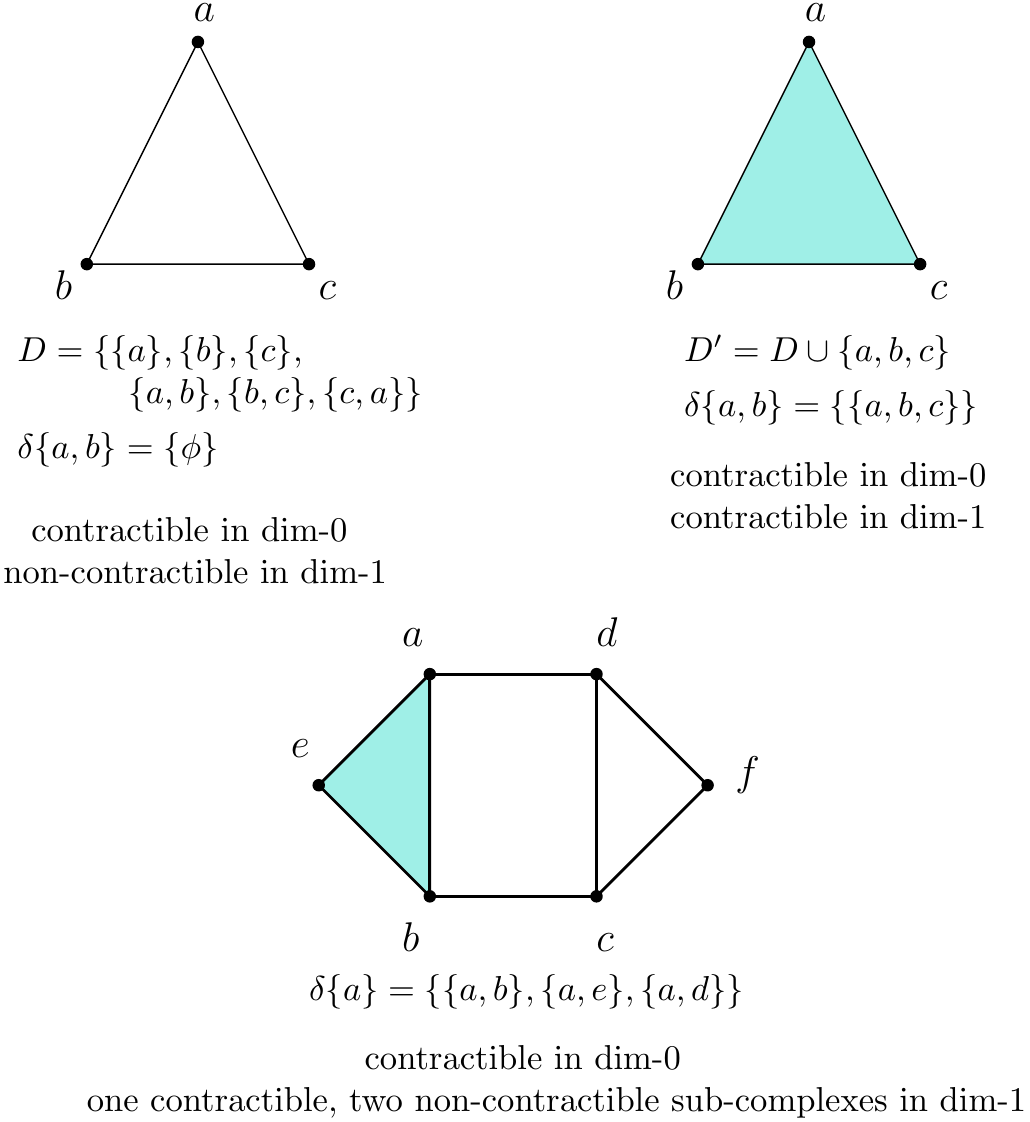}  
  \caption{Contractible and non-contractible structures in complexes.}
  \label{fig:homology_intuit_b}
\end{subfigure}
  \caption{Simplices form the building blocks for defining a topological space on discrete data set, simplicial
  topology.}
\label{fig:hom_intuit_examples}
\end{figure}

To formalize the notion of topology of a discrete data set,  a collection of so-called simplices is
defined from the data set as follows: An $n$-simplex is a set of  $n+1$ points and is said to have
dimension $n$ (dim-$n$). Figure~\ref{fig:simplices} shows different ways to interpret a simplex---as
a mathematical set, graphical object, or geometric object---in different dimensions.  The
\textit{boundary} of an $n$-simplex $\sigma,$ denoted  $\partial \sigma,$ is the set of all
$(n-1)$-simplices contained in the simplex.  The \textit{coboundary} of an $n$-simplex $\sigma,$
denoted by $\delta \sigma,$ is the set of all $(n+1)$-simplices $\omega$ such that $\sigma \in
\partial \omega.$  A collection of simplices is called a \textit{complex}.
Figure~\ref{fig:homology_intuit_b} shows examples of coboundaries and complexes. Contraction can be
visualized as a continuous deformation of a simplex to a point. Non-contractible topological
structures correspond to obstructions to such a contraction, suggesting the possible existence of a
feature in the data.

A hole in dim-$d$ is a complex that has a non-contractible boundary in dim-$(d-1).$ The complex $D'$
in Figure~\ref{fig:homology_intuit_b} contains the simplex $\{a,b,c\},$ and hence its boundary
$\partial\{a,b,c\}= \{\{a,b\}, \{b,c\}, \{c,a\}\}$ will contract in dim-1. On the other hand, since
the complex $D$ does not contain $\{a,b,c\},$ its boundary $\partial\{a,b,c\}$ cannot contract in
dim-1.  Hence, $D$ contains a non-contractible structure or a hole.  These non-contractible
structures in dim-$d$ define the \textit{homology group} for dim-$d,$ denoted by H$_d,$ partitioned
into equivalence classes that are related by contractible simplices. For example, the homology group
H$_1$ of the complex $D$ has one equivalence class. The non-contractible structures in H$_0$ can be
mapped to path-connected components when the complex is viewed as a discrete graph. Those in H$_1$
can be mapped to holes on the surface of the triangulation of the point cloud, but are more commonly
referred as \textit{loops}, indicative of one dimensional boundaries around the holes. In H$_2$ they
can be thought of as \textit{voids} in an embedding of the triangulation in a three-dimensional
metric space, and sets of triangular faces will define their boundaries.

\begin{figure}[h]
  \centering
  \includegraphics[width=\linewidth]{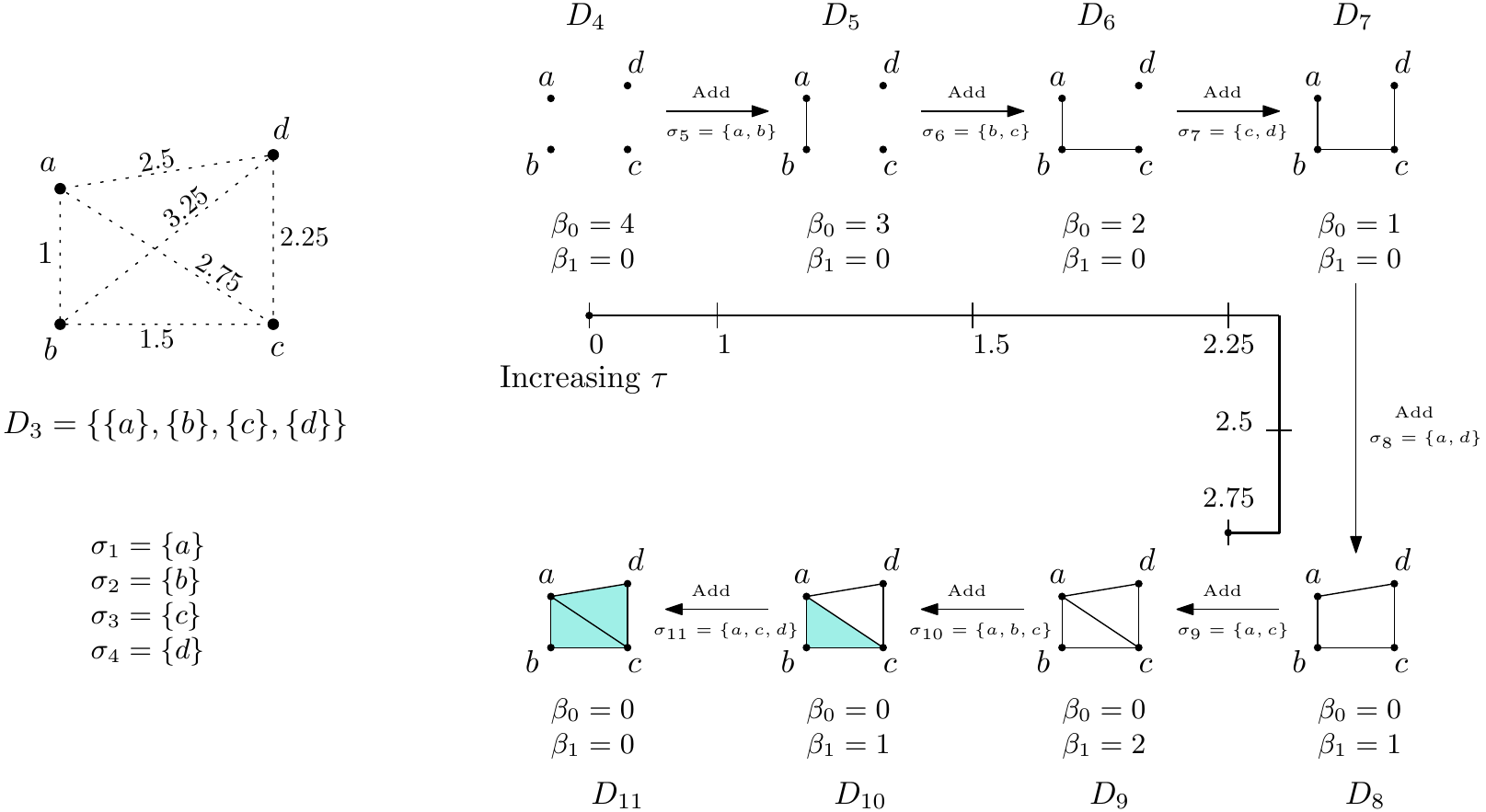}  

  \caption{Topological features---number of components ($\beta_0$) and holes ($\beta_1$)---change
  as the scale of observation ($\tau$) changes.}

  \label{fig:homology_intuit_a}
\end{figure}

To give a multi-scale overview, PH tracks changes in the homology groups as the scale of observation
changes. We see that the collection of simplices in the data set, otherwise known as the complex of
the data set, changes at different spatial scales as its construction is based on pairwise distances
in the discrete data set. The example (Figure~\ref{fig:homology_intuit_a}) shows 4 points in a
metric space, with the numbers between two points representing the spatial pairwise distances
($d(x,y)$). At any given scale of observation $\tau$ we build a complex with all of the 1-simplices
$\{a, b\}$ such that $d(x,y) \leq \tau.$ Additionally, to compute H$_d,$ all $(d+1)$-simplices whose
boundary is in the complex are also added to it. \textit{This results in a factorial increase in the
number of simplices to consider in the complex scaling as the number of points in the data set is
raised to a power two larger than} the dimension of the homology group. In
Figure~\ref{fig:homology_intuit_a}, when $\tau=0,$ there are 4 points or 0-simplices. These are
indexed arbitrarily from $\sigma_0$ to $\sigma_3.$ Starting from $D_0 = \{\sigma_0\},$ we define a
sequence of complexes $D_0 \subset D_1 \subset ... \subset D_n$ such that $D_i = D_{i-1} \cup
\{\sigma_i\}.$  At $\tau = 0,$ we have $D_3 = \{\sigma_0, \sigma_1, \sigma_2, \sigma_3\}.$ As $\tau$
increases, more simplices are added to the complex, and the sequence $(D_0, ..., D_{11})$ can be
computed for this example. Simplices that are added to the complex at the same value of $\tau$ can
be ordered arbitrarily relative to each other. This sequence of complexes is called the
\textit{Vietoris-Rips filtration}. 

\begin{figure}[h]
\begin{subfigure}{.48\textwidth}
  \centering
  \includegraphics[width=0.9\textwidth]{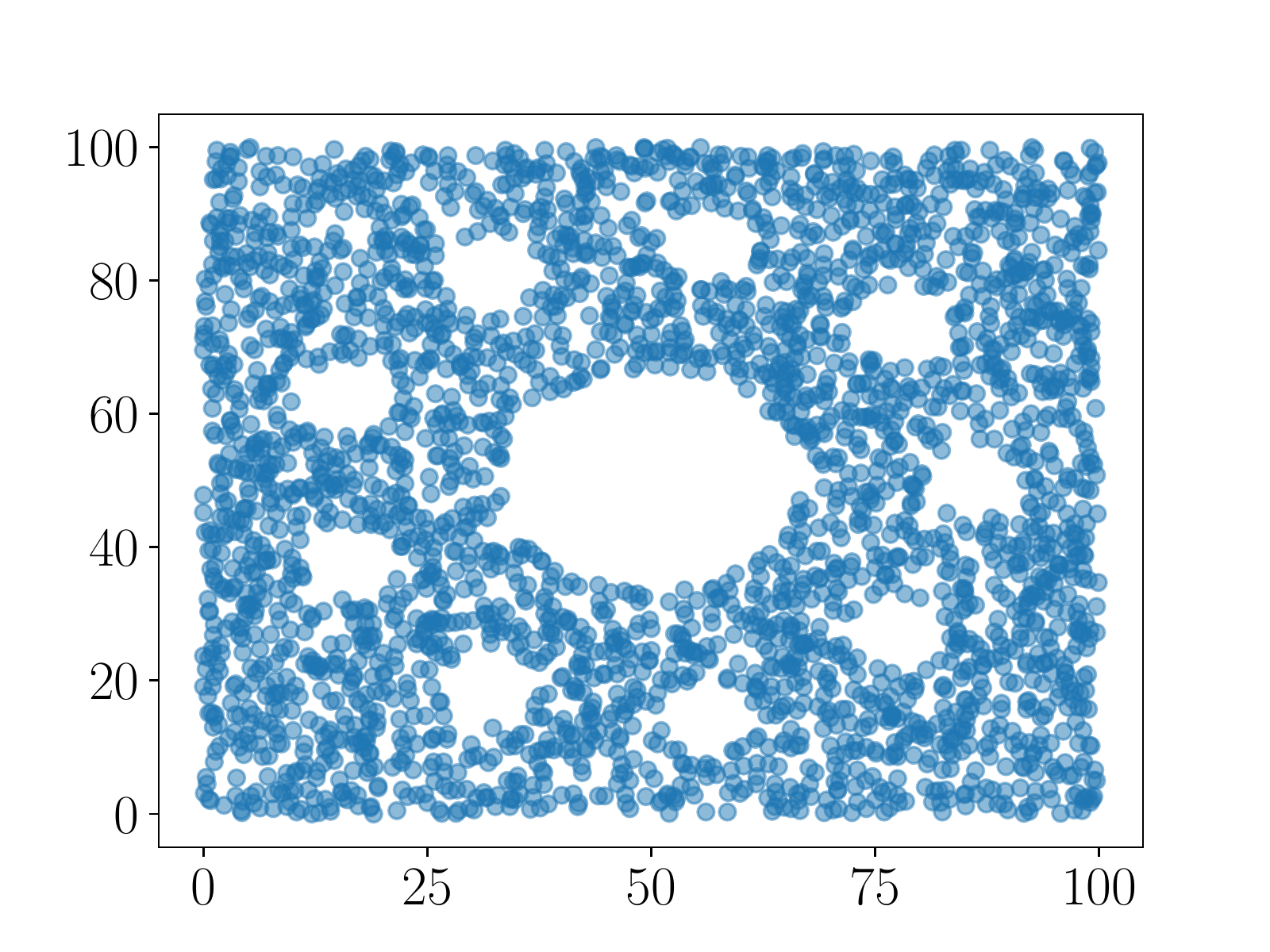}  
  \caption{Scatter plot from Figure~\ref{fig:rand3000_scatter}.}
  \label{fig:rand3000_only_scatter}
\end{subfigure}
\begin{subfigure}{0.48\textwidth}
  \centering
  \includegraphics[width=0.9\textwidth]{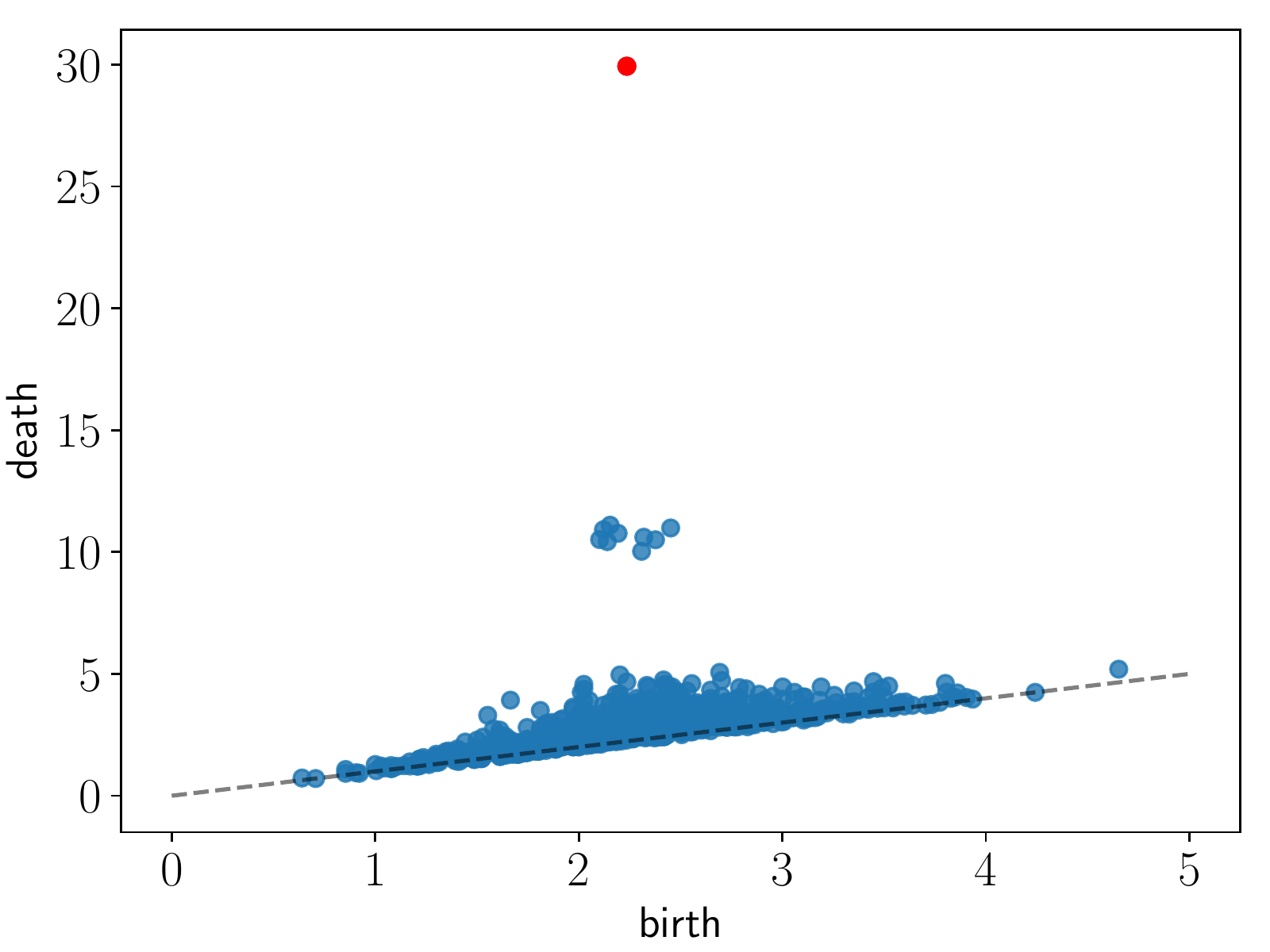}  

  \caption{H$_1$ PD shows that there are three topologically distinct groups of features that emerge
  across multiple scales. The red dot corresponds to the large hole at the center.}

  \label{fig:rand3000_PD}
\end{subfigure}
  \caption{Scatter plot of the data set and its H$_1$ persistence diagram.}
\label{fig:rand3000_scatter_PD}
\end{figure}

For every complex in the VR-filtration, we compute the homology groups and record changes in them as
we process complexes in the filtration. For example, there is the birth of a hole (equivalently, a
loop) in H$_1$ when $\sigma_8$ is added, which contracts or dies when $\sigma_{11}$ is added.
Birth-death pairs are called persistence pairs, and they are plotted as a function of the scale,
$\tau.$  In our example, the persistence pair is $(\sigma_8, \sigma_{11})$ in H$_1.$ These pairs
gives us \textit{persistence diagrams} (PD), one for every H$_d$ that is computed.  The persistence
diagram corresponding to H$_1$ for this example will contain exactly $(2.5, 2.75),$ since $\sigma_8$
is added at $\tau=2.5$ and $\sigma_{11}$ is added at $\tau=2.75.$ The persistence diagram for the
example in Figure~\ref{fig:rand3000_scatter} is shown in Figure~\ref{fig:rand3000_PD}.

It has been shown that the same persistence diagram can be obtained by computing cohomology groups,
denoted by H$^*_d.$ If $(\sigma, \tau)$ is a persistence pair in H$_d,$ then $(\tau, \sigma)$ is a
persistence pair in H$^*_d,$ and consequently, $\mid \text{H}_{d} \mid = \mid \text{H}^*_d \mid.$
Moreover, the algorithms that compute the persistence pairs of H$_d$ can also be used to compute the
persistence pairs of H$_d^*$ by applying them to the coboundaries of the complexes in the
filtration~\citep{de2011dualities}.

As discussed above, computing H$_2$ across all scales of a data set requires storing and processing
all 3-simplices, that is, $O(n^4)$ simplices. Even for a small data set with $n=500$ data points,
the number of 3-simplices is $500 \choose 4$ $=2573031125,$ indicative of the  memory required to
represent the filtration in the computer. Different methods have been developed for storing this
information. We compare our algorithm with three software packages---Gudhi, Ripser, and Eirene.
Gudhi represents the filtration using a simplex tree~\citep{boissonnat2014simplex}, and
Ripser~\citep{bauer2019ripser} represents a simplex using combinatorial indexing. For large data
sets, creating a simplex tree for the entire filtration a priori can require memory up to $O(n^4),$
and indexing simplices using combinatorial indexing overflows the bounds of integer data types in
most computer architectures. Therefore, both these methods cannot process data sets with large
numbers of points. Eirene uses matroid theory that takes more memory than Ripser and, in some cases,
more memory than Gudhi. All packages failed to compute PH for at least one data set in our
experiments. None were able to process the data set of interest to us, the conformations of the
human genome, because of such practical limitations. As PH computation requires processing a
combinatorially large number of simplices as well, any method for reducing memory requirements had
better not be accompanied by an inordinate increase in computation time either.

Due to these computational difficulties, published algorithms have been practically limited to
computing topological features up to and including H$_2.$ This still allows for applications of PH
to a large class of data sets in the physical sciences or to low-dimensional embeddings of
high-dimensional data. Therefore, we focus on computing topological features only up to and
including the first three dimensions. We take advantage of this restriction to devise a new way to
store information and new algorithms to process it, resulting in a reduction in memory requirements
by orders of magnitude accompanied with reduced computation time in almost all our test cases.

The two meter long human DNA fits into a nucleus with an average diameter of 10 $\mu$m by folding
into a complex, facilitated by many proteins which play functional and structural
roles~\citep{rowley2018organizational}. This folding is believed to have functional significance so
determining topological features like loops and voids in the folded DNA is of interest. Hi-C
experiments estimate pairwise spatial distances between genomic loci at 1 kilobase resolution
genome~\citep{lieberman2009comprehensive}. The resulting data set of around 3 million points is
analyzed by our algorithm in approximately ten minutes. 

Cohesin is a ring-shaped protein complex that has been shown to colocalize on chromatin along with a
highly expressed protein, CTCF, at anchors of loops in the folded chromosome~\citep{rao20143d},
indicative of its importance for loop formation in DNA. The H$_1$ PD corroborates the result
of~\citet{rao2017cohesin}---cohesin is crucial for loop formation in DNA because, upon addition of
auxin, an agent that is known to impair cohesin, the elimination of loops is observed. Additionally,
the H$_2$ PD reveals that auxin treatment leads to a significant reduction in the number of voids. 

The rest of this paper is structured as follows: Section~\ref{sec:algo_background} introduces the
algorithms that form the foundations of Dory. We summarize our contributions in
Section~\ref{sec:contribution}. The algorithm for Dory is explained in Section~\ref{sec:dory}. It is
then tested with pre-established data sets, and computation time and memory taken are compared with
published algorithms in Section~\ref{sec:experiments}. The analysis of human genome conformations
using Hi-C data is in Section~\ref{sec:human_gene}. We end with a discussion in
Section~\ref{sec:discussion}.

\section{Algorithmic Background}\label{sec:algo_background}

\begin{figure}[tbhp]
  \centering
\begin{subfigure}{.48\textwidth}
  \centering
  \includegraphics[width=0.9\linewidth]{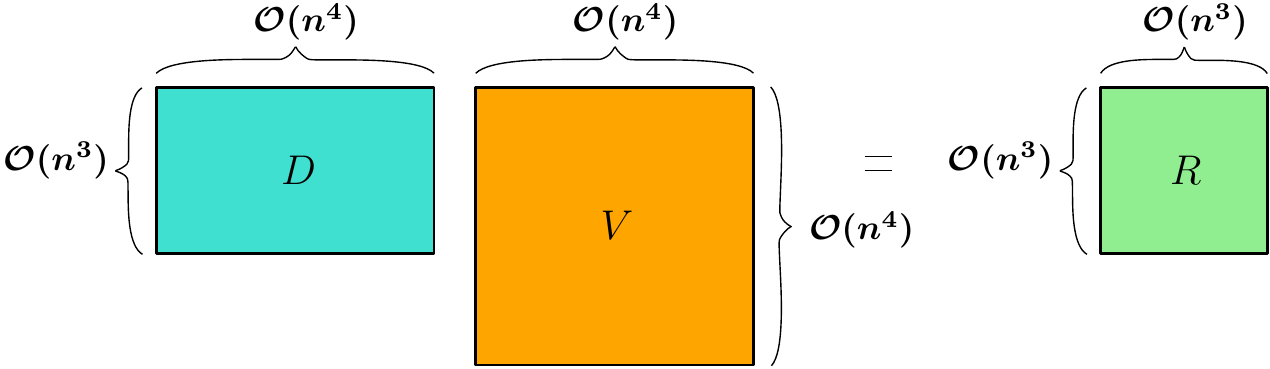}  
  \caption{Homology reduction}
  \label{fig:hom_matrix_size}
\end{subfigure}
  \centering
\begin{subfigure}{.48\textwidth}
  \centering
  \includegraphics[width=0.9\linewidth]{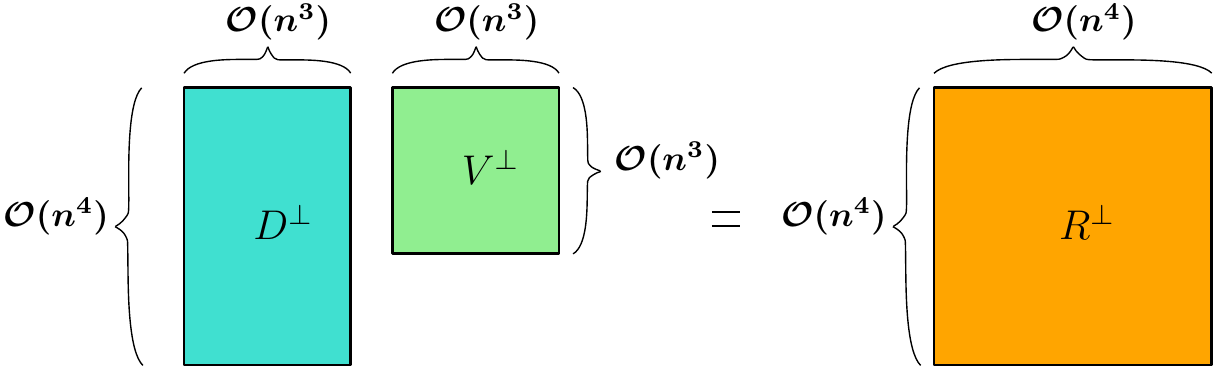}  
  \caption{Cohomology reduction}
  \label{fig:cohom_matrix_size}
\end{subfigure}
  \caption{Matrix representation and bounds of PH computation for a data set with $n$ points.}
\label{fig:matrix_sizes}
\end{figure}

An algorithm to compute the persistence birth-death pairs was given
by~\citet{edelsbrunner2000topological}, then reformulated as a matrix reduction
in~\citet{cohen2006vines}. Any given filtration, viewed as a set of sequences of simplices
$(\sigma_i)_{1\leq i \leq N},$ can be represented as a \textit{boundary matrix} $D = (d_{ij}) \in
\mathbb{R}^{N\times N},$ where  $d_{ij} = 1$ if $\sigma_i$ is a boundary element of $\sigma_j,$ and
is otherwise 0.  Consequently,  the indices of the columns and rows of $D$ represent the simplices
indexed according to their order in the filtration. We begin by defining a matrix $R = D.$ Then,
$\text{low}(j)$ is defined as the largest row index of the non-zero element in column $j$ of $R,$
that is, $\text{low}(j) = \argmax\limits_{i}\{d_{ij} = 1\}.$ The matrix reduction of $R$ is
formalized as adding (modulo 2) column $j$ with column $i,$ for $i < j,$ until $\text{low}(j)$ is a
\textit{pivot} entry---the first non-zero entry in the row with index $\text{low}(j)$ is at column
$j.$ This can be written as a matrix multiplication $DV = R$ (see Figure~\ref{fig:hom_matrix_size}),
where $V$ is the matrix that stores reduction operations and $R$ is the resulting matrix with all of
its columns reduced. In this work we specifically consider addition (modulo 2) of columns for
applications to spatial point-cloud data sets. This reduction can be carried out in two
ways---standard column algorithm (appendix~\ref{app:row_col}, algorithm~\ref{alg:std_col}) and
standard row algorithm (appendix~\ref{app:row_col}, algorithm~\ref{alg:std_row}).  When all columns
of $R$ have been reduced, the persistence pairs are given by $(\sigma_{\text{low}(j)}, \sigma_j)$
(born when $\sigma_{\text{low}(j)}$ is added and died when $\sigma_j$ is added). Further, if column
$j$ was reduced to $\mathbf{0}$ but $j$ is not a pivot of any column of $R,$ then there is a
non-contractible structure in the final complex that was born when $\sigma_j$ was added to the
filtration but it never contracted or died. We represent such a pair by $(\sigma_j, \infty).$

The same $V$ and, consequently, the same $R$ are obtained for the standard column and row
algorithms~\citep{de2011dualities}. Moreover, reduction of the coboundaries yields the persistence
pairs for the cohomology groups, H$^*_d,$ that are in one-to-one correspondence with the persistence
pairs of  H$_d.$ The coboundary matrix is denoted by $D^\bot = (d_{ij}) \in \mathbb{R}^{N\times N},$
where $d_{ij} = 1$ if $\sigma_{N-i+1}$ is in the coboundary of $\sigma_{N-j+1},$ and is otherwise 0.
In other words, the columns of $D^\bot$ are coboundaries, and the indices of the columns and rows of
$D^\bot$ are simplices ordered in the reverse order of the filtration sequence.  The matrix setup is
shown in Figure~\ref{fig:cohom_matrix_size}.  \citet{de2011dualities} observed empirically that
computing cohomology via the row algorithm provides improvements over homology computation in both
time taken and memory requirement.

Figure~\ref{fig:matrix_sizes} indicates the size of the matrices when computing PH up to and
including the first three dimensions for VR-filtration of a data set with $n$ points. If $\tau =
\infty,$ the filtration will admit $n \choose 0$ + $n\choose 1$ + $n\choose 2 $+ $n\choose 3$ +
$n\choose 4$ $\approx O(n^4)$ simplices (all of the possible 0-,1-,2-, and 3-simplices). For a data
set with as few as 500 points, the memory requirement just to store $D$ in a sparse format (storing
only the indices of the non-zero elements) is more than $41$ GB (presuming 4 bytes per
\texttt{unsigned int}). 

\section{Our Contribution}\label{sec:contribution}

\begin{itemize}

  \item A new way to index 2- and 3-simplices that significantly reduces memory requirement and also
    aids in reducing computation time.

  \item Algorithms to compute coboundaries for edges and triangles using our indexing method that
    optimize computation time of coboundary traversal during reduction.

  \item A fast implicit column algorithm to compute PH that can potentially reduce memory usage by a
    factor of the number of points in the data set without an inordinate increase in the
    computation time.

  \item Serial-parallel algorithm that distributes computation of PH over multiple threads without a
    significant increase in the total memory requirement.

  \item A computation of the PH of the human genome at high resolution, a data set with millions of
    points.

  \item Two versions of code---sparse and non-sparse. The sparse version requires memory
    proportional to the number of permissible edges in the filtration. The non-sparse version, while
    faster, requires memory proportional to the number of total edges possible in the data set.

\end{itemize}

\section{Our Algorithm}\label{sec:dory}

The aim is to compute persistence pairs up to three dimensions for VR-filtration. We will refer to
the 0-simplices as vertices ($v$), 1-simplices as edges ($e$), 2-simplices as triangles ($t$), and
3-simplices as tetrahedrons ($h$). Consider a point-cloud data set embedded in a metric space, that
is, there is a well-defined distance metric between any two points (vertices) in the space. We
denote the number of vertices by $n.$ The number of edges in the filtration, denoted by $n_e,$ will
depend upon the maximum permissible value of the filtration parameter, denoted by $\tau_m.$ If
$\tau_m$ is small compared to the maximum distance in the data set, we expect $n_e \ll {n \choose
2}$ and we call the filtration \textit{sparse}. Since the reduction operations are always between
complexes of the same dimension, we construct VR-filtrations for each dimension separately. Let
$\mathcal{S}^d$ be the set of all permissible simplices of dimension $d$ and $\mathcal{O}^d$ be the
corresponding set of orders in the filtration. We define bijective maps $f_d: \mathcal{S}^d \to
\mathcal{O}^d.$ Since all vertices are born at the same filtration parameter (zero), $f_0(v)$ for a
vertex $v$ can be arbitrarily assigned a unique whole number. For convenience, we will use $f_0(v)$
and $v$ interchangeably. For edges, the map $f_1(e)$ is the indexing defined by the sorting
algorithm applied to the lengths of the edges. The filtration for 0-simplices is then the list of
vertices ordered according to $f_0,$ and is denoted by $F_0.$ The filtration for 1-simplices is a
list of edges ordered according to $f_1,$ and is denoted by $F_1.$ Additionally, we will denote the
list of $d$-simplices in the reverse order of filtration by $F^{-1}_d.$ To define filtrations for 2-
and 3-simplices we introduce a new way to index triangles and tetrahedrons.

\subsection{Paired-indexing and Neighborhoods}

We first define the \textit{diameter} of a simplex $\sigma$ in the filtration as the maximum of the
orders of the edges in the simplex, denoted by $d(\sigma).$ The corresponding edge is denoted by
$d^{-1}(\sigma).$ Then, \textit{paired-indexing} uses a pair of keys, primary ($k^p$) and secondary
($k^s$), denoted by $\langle k^p, k^s \rangle.$ The primary key for both triangles and tetrahedrons
is their diameter and the secondary key is the order of the simplex defined by the remaining points.
For a triangle $t = \{a, b, c\},$ if $d^{-1}(t) = \{a, b\},$ then $f_2(\{a,b,c\}) = \langle
f_1(\{a,b\}), f_0(\{c\}) \rangle = \langle f_1(\{a,b\}), c\rangle.$  For a tetrahedron $h =
\{a,b,c,d\},$ if $d^{-1}(h) = \{a, b\},$ then, $f_3(\{a,b,c,d\}) = \langle f_1(\{a,b\}), f_1(\{c,
d\}) \rangle.$ We define an ordering on the paired-indexing as follows,

\begin{equation}\label{eqn:order_rule}
  \pairkeys{i} > \pairkeys{j} \,\, \text{iff either}\,\, \{\primkey{i} > \primkey{j}\} \,\,
  \text{or} \,\, \{\primkey{i} = \primkey{j} \, \text{and} \, \seckey{i} > \seckey{j}\}.
\end{equation}

This ordering on the paired-indexing preserves the order of the simplices in the VR-filtration since
a simplex with the larger diameter will have a greater order in the filtration. Simplices with the
same diameter can be ordered arbitrarily with respect to each other, which in paired-indexing is
based on the secondary key. The maps $f_2$ and $f_3$ then define filtrations $F_2$ and $F_3,$
respectively. We do not store these filtrations as lists in the algorithm and instead compute them
on the fly, reducing the memory required. We say that a simplex in $\mathcal{S}^d$ is greater than
another simplex in $\mathcal{S}^d$ iff its order in the filtration is greater.

The paired-indexing plays an important role in both reducing memory requirements and computation
time. Using the 4 byte $\texttt{unsigned int},$ this indexing will take exactly 8 bytes to represent
any triangle or tetrahedron in a filtration regardless of the number of points in the data set.
Further, since $k^p$ and $k^s$ are less than $n_e,$ the paired-indices are bounded by
$\mathcal{O}(n_e),$ rather than $\mathcal{O}(n^4).$ This specifically reduces memory requirement by
many orders of magnitude for sparse filtrations, where $n_e \ll {n \choose 2}.$

\subsection{Computing Coboundaries}\label{sec:compute_cob}

\begin{figure}[tbhp]

  \includegraphics{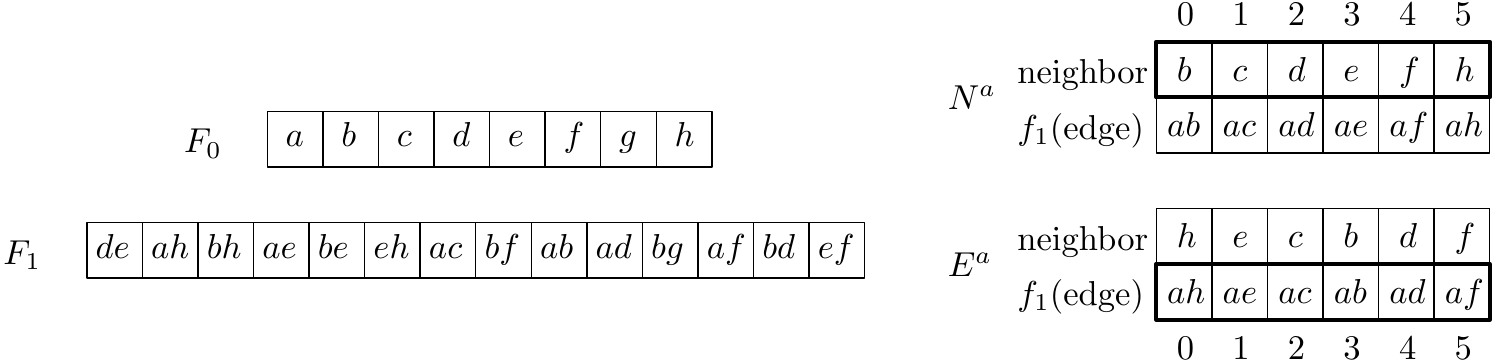}

 \caption{An example of a VR-filtration. $F_0$ is the list of vertices and $F_1$ is the list of
  edges, both sorted in the order of the filtration. The vertex- and edge-neighborhoods for vertex
  $a$ are shown as example. The dark border indicates that $N^a$ is sorted by the order of neighbors
  of $a$ and $E^a$ is sorted by the order of the corresponding edges.}

  \label{fig:neighborhood_figure}
\end{figure}

We show how paired-indexing can be used to compute coboundary of edges and triangles. We begin by
defining \textit{vertex-neighborhood} $(N^a)$ and \textit{edge-neighborhood} $(E^a)$ of a vertex
$a.$ The neighbor of the vertex $a$ is a vertex that shares an edge with $a.$ Both $N^a$ and $E^a$ are
lists of all neighbors of $a.$ Each element of these lists is a structure that contains a neighbor
of $a$ and the order of the edge between them. The vertex-neighborhood is sorted by the order of the
neighbors and the edge-neighborhood is sorted by the order of the corresponding edges (see
Figure~\ref{fig:neighborhood_figure}). The vertex and edge neighborhoods of all vertices are
computed using $F_0$ and $F_1,$ and they are stored in the memory. We will use $F_0$ and $F_1$ in
Figure~\ref{fig:neighborhood_figure} as an example to illustrate how paired-indexing can be used to
compute coboundary of an edge and a triangle. For convenience, the order of an edge $\{a,b\},$
$f_1(\{a,b\}),$ is denoted by $ab.$

\subsubsection{Coboundaries of Edges}\label{sec:cob_edges}

\begin{figure}[tbhp]
  \includegraphics[width=\textwidth]{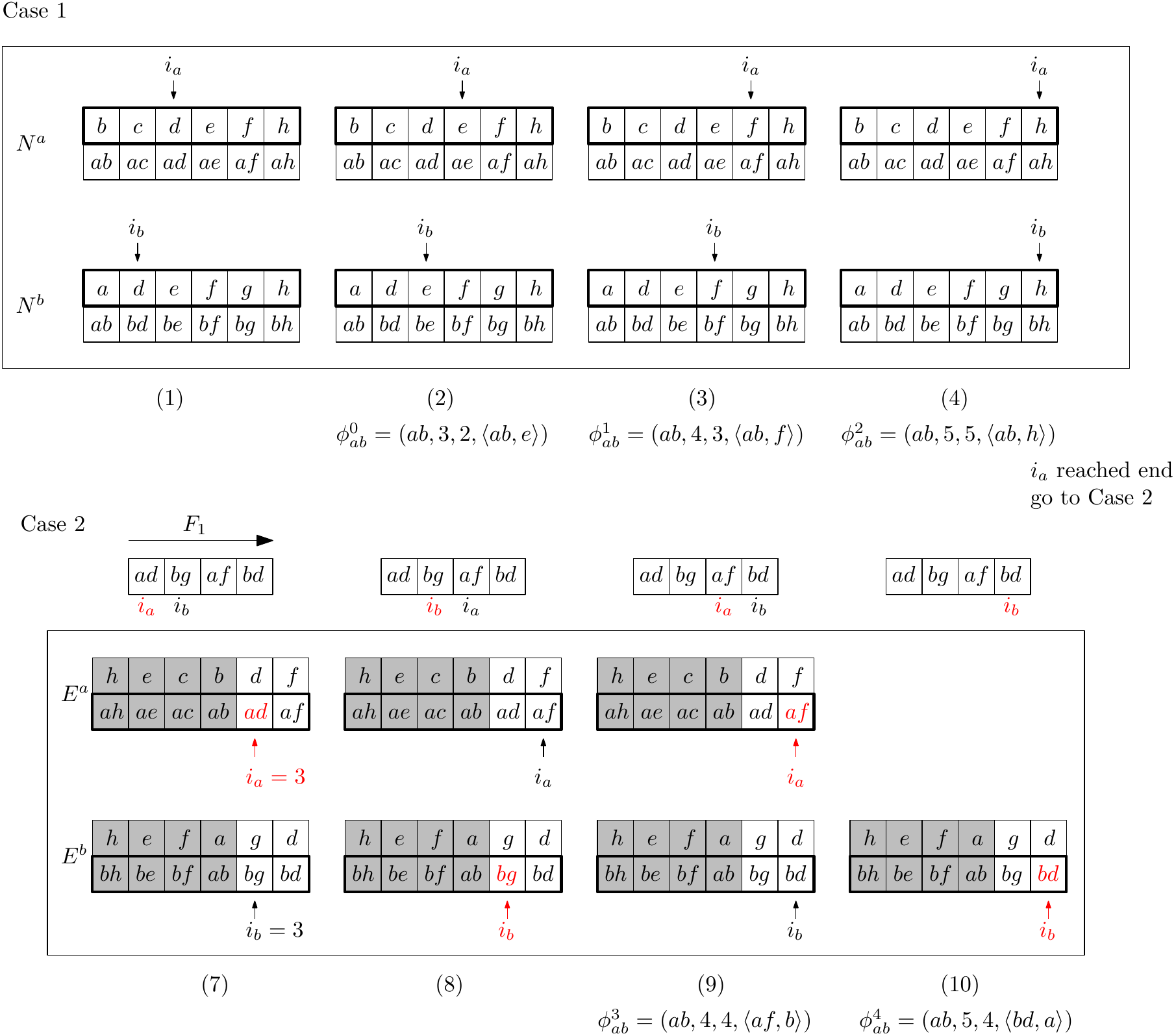}  

  \caption{Using vertex- and edge-neighborhoods to compute simplices in the coboundary of edge
  $\{a,b\}$ (see Figure~\ref{fig:neighborhood_figure} for $F_1$), such that they are in the order of
  $F_2.$ The valid triangles are stored using $\phi$-representation.}

\label{fig:edge_cob}
\end{figure}

Consider the edge $\{a,b\}$ in Figure~\ref{fig:neighborhood_figure}. Any simplex in its coboundary
is a triangle $t = \{a,b,v\} = \left< k_1, k_2 \right>,$ where $v$ is a common neighbor of $a$ and
$b$, and the diameter of $t = k_1 \geq ab.$ We consider two cases: case 1, triangles with diameter
equal to $ab,$ and case 2, triangles with diameter greater than $ab.$ The triangles in case 1 are
smaller than those in case 2. Hence, to compute the simplices in the coboundary as ordered in $F_2,$
we start in case 1---the primary key is $ab$ and the order of the simplices is decided by the vertex
in $t$ that is not $a$ and $b,$ that is, $v.$ We traverse along the vertex-neighborhoods using
indices $i_a$ and $i_b$ to find common neighbors in an increasing order. Initially, both are set to
0. As shown in Figure~\ref{fig:edge_cob}, we increment the index pointing to the lower ordered
vertex till both indices point to the same vertex $v$, in which case, $v$ is a common neighbor and
$\{a,b,v\}$ exists in the filtration. However, we have to ensure that its diameter is $ab$ since we
are in case 1.  For example, in (1) in Figure~\ref{fig:edge_cob}, we skip $\{a,b,d\}$ because $ad >
ab$ (see $F_1$ in Figure~\ref{fig:neighborhood_figure}). Note that the structures in
vertex-neighborhoods give direct access to $av$ and $bv$ at $i_a$ and $i_b$ for comparison with
$ab.$ Otherwise, if the diameter of $\{a,b,v\} = ab,$ then $\left< ab, v\right>$ is the next greater
triangle in $\delta \{a,b\}$ in $F_2.$ This triangle is recorded as $\phi = (ab, i_a, i_b,
\delta_*)$ where $\delta_* = f_2(\{a,b,v\}) = \left < ab, v\right>.$ We call this the
$\phi$-representation of $\delta_*$ as a simplex in the coboundary of $\{a,b\}.$ If either of the
indices reach the end of the corresponding neighborhood, we go to case 2.

In case 2, the triangles in $\delta \{a,b\}$ have diameter greater than $ab.$ Hence, their order in
the filtration is defined by the diameter of the primary key. To compute the primary keys as ordered
in $F_2,$ we define $i_a$ and $i_b$ to be indices of the respective edge-neighborhoods since they
are sorted by the order of the edges. We set $i_a$ and $i_b$ to point to the smallest edges greater
than $ab$ (binary search operations in $E^a$ and $E^b,$ respectively). We then consider the index
that points to the lower ordered edge. For example, in (7) in Figure~\ref{fig:edge_cob}, $i_a=3$
points to $ad$ and $i_b=3$ points to $bg.$ We consider $i_a=3$ because $ad < bg.$ Now, we have to
check that $\{a,b,d\}$ exists and its diameter is $ad.$ A single binary search operation over $N^b$
finds that $d$ is a neighbor of $b,$ confirming the existence of $\{a,b,d\},$ and it also gives us
$bd.$ Since $ad < bd,$ the diameter of $\{a,b,d\}$ is not $ad$ and we do not record it as a
coboundary.  We proceed by incrementing $i_a$ by one. Now $i_b$ points to the smaller edge, and the
corresponding triangle is $\{a,b,g\}.$ A binary search over $N^b$ finds that $g$ is not a neighbor
of $a,$ hence, this triangle does not exist, and $i_b$ is incremented by 1 to continue. The index
$i_a$ now points to the smaller edge ($af$), and the corresponding triangle is $\{a,b,f\}.$ Since,
this triangle exists and $af$ is its diameter, $\left < af, b\right>$ is the next simplex in the
coboundary and it is recorded as $\phi = (ab, i_a, i_b, \left< af, b\right>).$ This process is
continued till both $i_a$ and $i_b$ reach the end of the respective edge-neighborhood.

The above computation yields the coboundary of $\{a,b\}$ as a list of tuples $\phi^i_{ab}$ (see
Figure~\ref{fig:edge_cob}) in an increasing order. This is an inefficient method to compute the
coboundary of a simplex if it needs to be computed only once. A faster method would be to create the
entire list of simplices in the coboundary and sort it by the order. However, reduction can require
the computation of the coboundary of a simplex multiple times.  Further, it is not feasible to store
the entire coboundary matrix a priori due to memory limitations---even the size of the coboundary of
one simplex can be up to $O(n^4)$(see Figure~\ref{fig:matrix_sizes}).  Using our
$\phi$-representation, we implement three algorithms to address both these issues.  First, the
smallest simplex in the coboundary of $\{a,b\}$ can be found simply by starting in case 1 with $i_a$
and $i_b$ initialized to 0, and we proceed as outlined previously. The first valid simplex
encountered is the answer.  This is implemented as \texttt{FindSmallestt}
(algorithm~\ref{alg_findsmallest_edge}). Second, given a tuple $\phi = (ab, i_a, i_b, \left< k_1,
k_2\right>),$ the next greater simplex in $\delta \{a,b\}$ can be computed as follows. If $k_1 =
ab,$ then $i_a$ and $i_b$ are indices of the vertex-neighborhoods (case 1) and if $k_1 > ab,$ they
are indices of edge-neighborhoods (case 2). After determining the scenario, we proceed as outlined
previously. As a result, given the $\phi$-representation of a triangle $t$ in the coboundary of an
edge $\{a,b \},$ we can compute the next greater triangle in its coboundary without always having to
traverse entire neighborhoods of $a$ and $b.$ We implement this as \texttt{FindNextt}
(algorithm~\ref{alg_findnext_edge}). Third, given a triangle $\delta_{\#} = \left< k_1, k_2
\right>,$ the smallest simplex greater than or equal to $\delta_{\#}$ in the coboundary of an edge
$\{a,b\}$ can be computed by considering three scenarios.  (1) If $k_1 <  ab,$ then the triangle
with the least order in the coboundary of $\{a,b\}$ is the answer. (2) If $k_1 = ab,$ then we start
in case 1 at indices $i_a$ and $i_b$ of vertex-neighborhoods that point to the neighbor with
smallest order greater than or equal to $k_2$. If no such neighbor exists, then we initialize $i_a$
and $i_b$ to the beginning of case 2.  (3) If $k_1 > ab,$ then we start in case 2 at the indices
$i_a$ and $i_b$ of edge-neighborhoods that point to the edges with smallest order greater than or
equal to $k_1.$ This is implemented as \texttt{FindGEQt} (algorithm~\ref{alg_findgeq_edge}).  Note
that the search operations in all of these algorithms are binary search operations since vertex- and
edge-neighborhoods are sorted. We will show in~\ref{subsubsec:imp_row_alg} how these three
algorithms are utilized to reduce the coboundaries to compute persistence pairs.  See
appendix~\ref{app:cob_edge} for all pseudocode related to coboundaries of edges.

\subsubsection{Coboundaries of Triangles}\label{subsec:cob_triangles}

\begin{figure}[tbhp]
  \includegraphics[width=\textwidth]{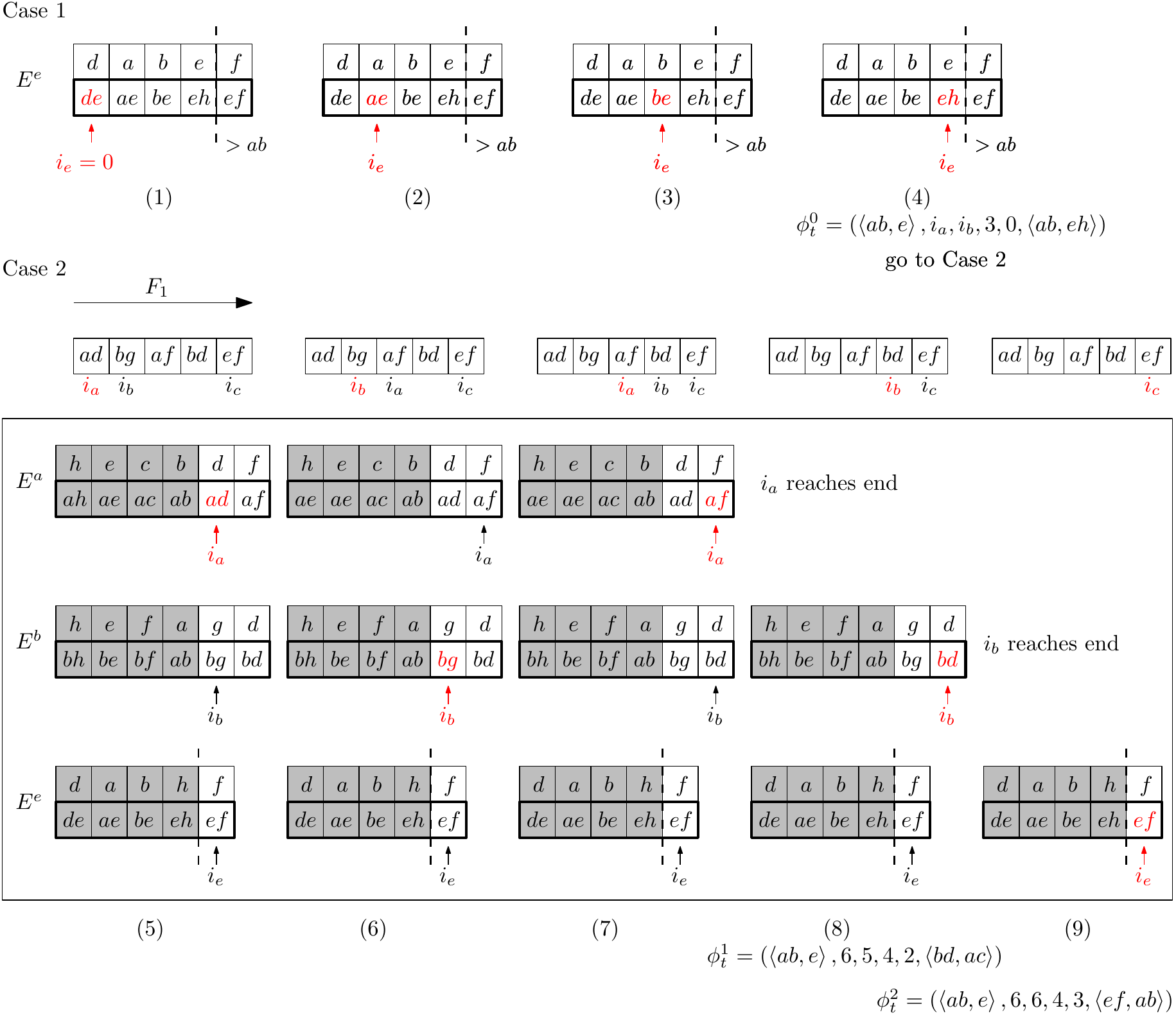}  

  \caption{Using edge-neighborhoods to compute simplices in the coboundary of triangle $\{a,b, e\}$
  (see Figure~\ref{fig:neighborhood_figure} for $F_1$), such that they are in the order of $F_3.$
  The index marked in red in every iteration, points to the edge under consideration. The valid
  tetrahedrons are stored using $\phi$-representation.}

\label{fig:triangle_cob}

\end{figure}

Consider the triangle $t = \{a,b,e\} = \left < ab, e\right >$ in
Figure~\ref{fig:neighborhood_figure}.  Any simplex in its coboundary is a tetrahedron $t = \{a,b,e,
v\} = \left< k_1, k_2 \right>,$ where $v$ is a common neighbor of $a,b,$ and $e$, and the diameter
of $h = k_1 \geq ab.$ We will use three indices, $i_a, i_b,$ and $i_e,$ to keep track of
neighborhoods of $a, b, $ and $e,$ respectively. We consider two cases: case 1, tetrahedrons with
diameter equal to $ab,$ and case 2, tetrahedrons with diameter greater than $ab.$  The tetrahedrons
in case 1 are smaller than the triangles in case 2, so we start with case 1---the primary key is
$ab$ and the order of the simplices is decided by the order of the edge $\{e,v\}.$ To compute such
edges according to their order in $F_1,$ we traverse through the edge-neighborhood of $e.$
Initially, $i_e$ is set to 0. For every edge $\{e,v\}$ that $i_e$ points to, we check whether $v$ is
a neighbor of $a$ and $b$ to confirm existence of $\{a,b,e,v\},$ and we check that $ab$ is its
diameter. These checks require two binary search operations, one each for $N_a$ and $N_b.$ If both
conditions are satisfied, then we record the tetrahedron. We proceed by incrementing $i_e$ by one.
For example, when $i_e=0$ in (1) in Figure~\ref{fig:triangle_cob}, the resulting tetrahedron exists
but $ab$ is not its diameter, and we do not record it. We skip when $i_e$ points to $ae$ or $be$
because they are not valid tetrahedrons. The first tetrahedron with diameter $ab$ is when $i_e = 3,$
and hence, $\{a,b,e,h\}$ is the smallest tetrahedron in $\delta t.$  We define the
$\phi$-representation for tetrahedrons as the tuple $\phi = (t, i_a, i_b, i_e, f, \left< ab,
eh\right>),$ where $f$ flags the diameter as follows---$f$ is $0,1,2,3$, and 4 correspond to
diameter being $ab$, $ah$, $bh$, and $eh$, respectively. This flag is implemented to determine which
index has to be incremented to find the next greater tetrahedron in $\delta t.$ For example, if $ab$
is the diameter, that is, $f=0,$ we are in case 1 and $i_e$ is to be incremented by one to proceed.
Note that, in general, for any given triangle $t = \left < k^p, k^s \right >,$ we compute $a \gets
\text{min}\{f^{-1}_1(k^p)\},$ $b \gets \text{max}\{f^{-1}_1(k^p)\},$ and $e \gets k^s.$ Case 1 is
continued till $i_e$ points to an edge greater than $ab,$ since all tetrahedrons henceforth will
have a diameter greater than $ab.$ In Figure~\ref{fig:neighborhood_figure}, $ef > ab$ (indicated by
a dashed line), hence we start with case 2 when $i_e = 3.$

All valid tetrahedrons in case 2 that are in $\delta t$ will have diameter greater than $ab.$ The
index $i_e$ already points to the entry in $E^e$ that has the smallest edge greater than $ab$. We
implement binary search to compute $i_a$ and $i_b$ such that they also point to the smallest edge
greater than $ab$ in the respective edge-neighborhood. All tetrahedrons in case 2 will have
different primary keys since they have different diameters. Among the indices $i_a, i_b,$ and $i_e,$
we consider the index that is pointing to the edge with the minimum order. As before, we have to
check the existence of the tetrahedron and that its diameter is the edge under consideration. For
example, in Figure~\ref{fig:triangle_cob}, we begin case 2 in (5) by considering $i_a$ because it
points to $ad = \text{min}\{ad, bg, ef\}.$  Binary searches for vertex $d$ over $N^b$ and $N^e$
confirm the existence of the tetrahedron $\{a,b,e,d\}$ and also show that $ad$ is not its diameter.
Hence, this tetrahedron is not recorded and $i_a$ is incremented by 1 to proceed. Now, $i_b$ points
to the smallest edge, but the resulting tetrahedron does not exist since $g$ is not a neighbor of
$e.$ The smallest tetrahedron in case 2 is $\{a,b,e,d\},$ recorded at (8) as $(\left < ab, e\right
>, 6,5,4,2,\left < bd, ac\right >).$ Note that $f$ is 2 since the diameter is $bd.$ This flag
dictates that $i_b$ is to be incremented by 1 to proceed because the current pointer being
considered is $i_b.$ Similarly, if $f=1$ then $ad$ is the diameter and $i_a$ has to be incremented
to proceed and if $f=3,$ then $ed$ is the diameter and $i_e$ has to be incremented by 1.  The index
$i_e$ is incremented to proceed when either $f=0$ or $f=3,$ but we are in case 1 in the former and
in case 2 in the latter. Case 2 ends when all three indices reach the end of their respective
edge-neighborhood.

The above computation yields the coboundary of $t=\{a,b,e\}$ as a list of tuples $\phi^i_t$ (see
Figure~\ref{fig:triangle_cob}). The case and the index to be incremented is determined by the flag
$f$ as discussed previously. As done for edges, this $\phi$-representation is used to implement
three algorithms---\texttt{FindSmallesth}, \texttt{FindNexth}, and \texttt{FindGEQh}. The smallest
simplex in the coboundary of $t$ can be found simply by starting in case 1, initializing $i_e$ to 0,
and proceeding as outlined previously. The first valid simplex encountered is the answer
(algorithm~\ref{alg:findsmallest_triangle}). For \texttt{FindNexth}
(algorithm~\ref{alg:findnext_triangle}), given the $\phi$-representation of a simplex $\left< k_1,
k_2\right>$ in $\delta t$ as $\phi = (t, i_a, i_b, i_e, f, \left< k_1, k_2\right>),$ the smallest
simplex greater than $\left<k_1, k_2 \right> $ in $\delta t$ can be computed by proceeding in case 1
if $f=0$ and proceeding in case 2 otherwise---incrementing the indices by one as dictated by the
value of $f.$ Finally, given a tetrahedron $\delta_{\#} = \left< k_1, k_2 \right>,$ the search for
the smallest simplex greater than or equal to $\delta_{\#}$ in $\delta t$ is optimized by
considering three scenarios (algorithm~\ref{alg:findgeq_triangle}). (1) If $k_1 <  ab,$ then the
smallest tetrahedron in $\delta t$ (\texttt{FindSmallesth}) is the answer. (2) If $k_1 = ab,$ then
we start in case 1 and search for $i_e$ that points to the smallest edge greater than or equal to
$k_2.$ If no such edge exists, then we start in case 2 with $i_e$ pointing to the end of $E^e.$ (3)
If $k_1 > ab,$ we are in case 2 and we find the indices that are greater than or equal to $k_1$ in
the respective edge-neighborhoods. See appendix~\ref{app:cob_triangle} for all the pseudocode
related to coboundaries of triangles.

\subsection{Cohomology Reduction}

To compute persistence pairs using cohomology reduction, we reduce one simplex at a time since
it is often not feasible to store the coboundary matrix $D^\bot.$ Suppose the reduced coboundaries
are stored in $R^\bot$ and the reduction operations are stored in $V^\bot.$ We will denote the
column in $R^\bot$ that contains the reduced coboundary of edge $e$ by $R^\bot(e)$ and the
corresponding column in $V^\bot$ by $V^\bot(e).$ Using this notation, the column reduction to reduce
coboundary of one edge at a time is shown in algorithm~\ref{alg:basic_red1}. Here, low$(R^\bot(e))$
is the simplex with the smallest order in column $R^\bot(e).$ The partial reduction of the edge is
stored in $r$ which initially is the coboundary of the edge. If low$(r) = $ low$(R^\bot(e'))$ for
some edge $e',$ then $r$ is reduced (sum modulo 2) with $R^\bot(e').$ If no such edge exists or if
$r$ is $\mathbf{0},$ then $r$ is completely reduced. A non-empty completely reduced $r$ is recorded
as the column $R^\bot(e)$ or $R^\bot,$ and (low($R^\bot(e)$), $e$) is a persistence pair.

\begin{algorithm}
  \begin{algorithmic}[1]

    \State \textbf{Input:} $F_1$

    \State $R^\bot$ is empty

    \For{$e$ in order of $F_1^{-1}$}

        \State $r \gets \delta e$

        \While{$r$ not empty AND there exists edge $e'$ s.t. low$(R^\bot(e'))= $ low$(r)$}
                \State $r \gets r \oplus R^\bot(e')$
        \EndWhile

        \If{$r$ not empty}
          \State $R^\bot(e) \gets r$
        \EndIf

    \EndFor

  \end{algorithmic}
  \caption{Starting with standard column reduction}
  \label{alg:basic_red1}
\end{algorithm}

\subsubsection{Improving Scalability of Memory Requirement}\label{subsubsec:mem_scale}

However, Figure~\ref{fig:cohom_matrix_size} shows that $R^\bot$ has the worst bounds on size.
Therefore, we store only the reduction operations, that is, $V^\bot.$ Then, $r$ can be implicitly
reduced with $R^\bot(e')$ by summing (modulo 2) it with the coboundaries of edges in $V^\bot(e')$
(see algorithm~\ref{alg:implicitV_red2}). This algorithm is implemented in Ripser. The smallest
simplex $t$ after reducing coboundary of an edge $e'$ is stored as a persistence pair $(t, e')$ in
$p^\bot.$ In this algorithm, the partial reduction operations are recorded in $v,$ the result of the
ongoing reduction is in $r,$ and the smallest simplex in $r$ is stored as $\delta_*.$ However,
keeping track of $r$ can require a lot of computer memory as its length is bounded by $O(n^3)$ (and
by $O(n^4)$ when reducing triangles). We begin by proposing an implicit row algorithm that does not
store $r$ at any stage of reduction, and it implicitly reduces $r$ using $v.$ Hence, its memory
requirement depends on the size of $v$---$O(n^2)$ for reducing edges and $O(n^3)$ for reducing
triangles, potentially reducing the memory requirement by a factor of $n.$

\begin{algorithm}
  \begin{algorithmic}[1]

    \State \textbf{Input:} $F_1$

    \State $V^\bot, p^\bot$ are empty

    \For{$e$ in $F_1^{-1}$}

        \State $r \gets \delta e$
        \State $\delta_* \gets$  low$(r)$
        \State $v \gets [e]$

        \While{$\delta_*$ is not empty AND there is a pair $(\delta_*, e')$ in $p^\bot$}
                \For{$e''$ in $V^\bot(e')$}
                    \State $r \gets r \oplus \delta e''$
                \EndFor
                \State $\delta_* \gets \text{low}(r)$\Comment{Update the low}
                \State $v \gets v \oplus V^\bot(e')$\Comment{Update reduction operations for $e$}
        \EndWhile

        \If{$\delta_*$ not empty}
          \State $V^\bot(e) \gets v$
          \State $p^\bot \gets (\delta_*, e)$
        \EndIf

    \EndFor

  \end{algorithmic}
  \caption{Do not store $R^\bot$: Implicitly compute from $V^\bot$}
  \label{alg:implicitV_red2}
\end{algorithm}

\subsubsection{Implicit Row Algorithm}\label{subsubsec:imp_row_alg}

The reduction of the coboundary of an edge requires the computation of the smallest triangle in $r$
with non-zero coefficient, denoted by $\delta_*$ in the algorithm. Implicitly, $\delta_*$ is the
smallest triangle with non-zero coefficient after coboundaries of all edges in the corresponding $v$
are summed.  The idea behind the algorithm is that we will store $\phi$-representations of only the
smallest triangle with non-zero coefficient in the coboundary of every edge in $v$, the function
\texttt{FindGEQt} will optimize reduction by eliminating certain redundant reductions, and the
function \texttt{FindNextt} will traverse along the coboundaries during reduction. We begin with
introducing a basic strategy to compute $\delta_*$ in this algorithm. To explain the algorithm, we
will walk through an example that reduces $r = \delta e_0$ with $R^\bot$ using only $v,$ $V^\bot,$
and $p^\bot.$

\begin{figure}
  \includegraphics[width=\textwidth]{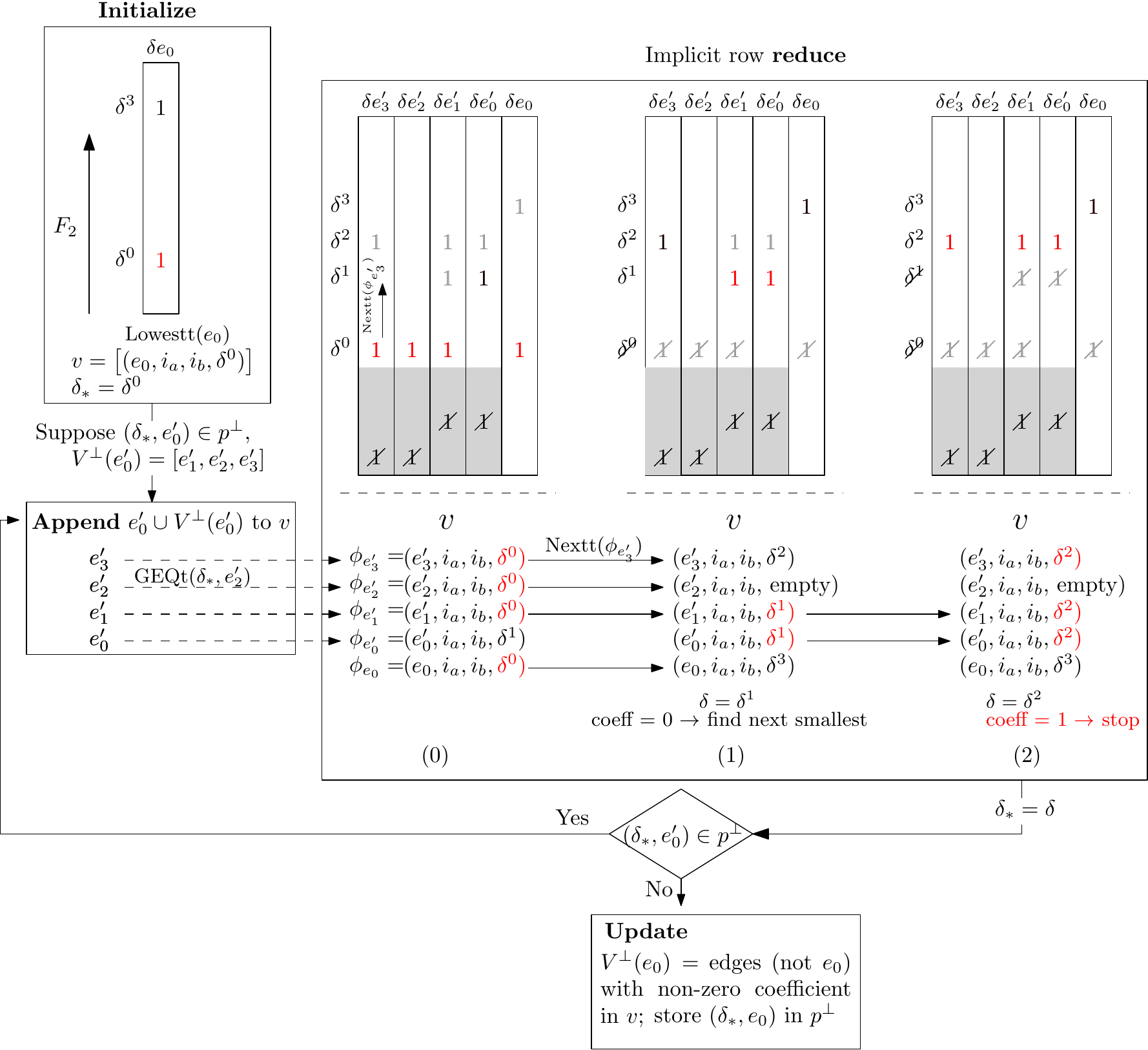}

  \caption{Example to show implicit reduction of $r = \delta e_0$ with $R^\bot$ using $V^\bot$ and
  $p^\bot.$ For every edge in $v,$ we record the smallest triangle in its coboundary with non-zero
  coefficient using the $\phi$-representation (marked in black). The smallest among these are marked
  in red. Simplices with zero coefficient are shown as crossed out.}

  \label{fig:implicit_row_reduce}
\end{figure}

Step 1 (`initialize' in Figure~\ref{fig:implicit_row_reduce}): As shown in the figure, $\delta^0$ is
the smallest simplex in $\delta e_0.$ Then, $v$ is initialized as a list with the single entry of
$\phi$-representation of $\delta^0$ in $\delta e_0,$ denoted by $\phi_{e_0} = (e_0, i_a, i_b,
\delta^0).$ This corresponds to the information that $r = \delta e_0$ initially and the smallest
triangle in $r$ is $\delta^0.$ Subsequently, $\delta_*$ is initialized with $\delta^0.$

Step 2 (`append' in Figure~\ref{fig:implicit_row_reduce}): Now, suppose there exists a pair
$(\delta_*, e')$ in $p^\bot$ and $V^\bot(e'_0) = [e_1', e_2', e'_3].$ Then, $r$ is to be summed with
$R^\bot(e').$ Since the smallest triangle with non-zero coefficient in $R^\bot(e')$ is $\delta_*,$
we know that when coboundaries of all $e'_k$ in $V^\bot(e')$ are summed with each other, then every
triangle that is smaller than $\delta_*$ will have a zero coefficient (shown as shaded area in the
figure). Hence, triangles in $\delta e'_k$ that are smaller than $\delta_*$ will have zero
coefficient and they do not matter. Therefore, for every edge $e'_k$ in $V^\bot(e'_0)$ and the edge
$e'_0,$ we use \texttt{FindGEQt} to compute the smallest triangle in their coboundary that is
greater than or equal to $\delta_*,$ and append its $\phi$-representation to $v$ (see (0) in
Figure~\ref{fig:implicit_row_reduce}). This eliminates unnecessary reductions of the
coboundaries of the new edges that are being appended to $v.$ Moreover, \texttt{FindGEQt} does so
efficiently by considering three cases and conducting binary searches as explained
in~\ref{sec:cob_edges}. Hence, paired-indexing not only improves the limits on memory requirements,
but also plays a crucial role in reducing the computation time of the reduction. The updated $v$
corresponds to the updated $r \gets r \oplus R^\bot(e'),$ therefore the coefficient of $\delta_*$
will be 0. To find the smallest triangle with non-zero coefficient in $r,$ we go to step 3.

Step 3 (`reduce' in Figure~\ref{fig:implicit_row_reduce}):  We first define $\delta = \text{MAX} =
\left < n_e, 0 \right >$ (since $k_1 < n_e$), and then we process every $\phi_e = (e, i_a, i_b,
\delta_e)$ in $v$ as follows.  If $\delta_e = \delta_*,$ then we find the next simplex greater than
$\delta_e$ in coboundary of $e$ using \texttt{FindNextt} since the coefficient of $\delta_*$ is 0.
If the updated $\delta_e$ is less than $\delta,$ then we set $\delta \gets \delta_e$ and set the
coefficient to 1, otherwise, 1 is added (modulo 2) to the coefficient. As a result, $\delta$ stores
the smallest triangle greater than $\delta_*$ in $r,$ and we also keep track of its coefficient. For
example, starting at (0) in Figure~\ref{fig:implicit_row_reduce}, $\delta_e = \delta_1$ for the
edges $e'_3, e'_2, e'_1, $ and $e_0.$ For these edges we compute the next greater triangle in their
respective coboundary using \texttt{FindNextt}. After iterating through all entries in $v,$ it is
determined that $\delta^1$ is the smallest simplex greater than $\delta_*,$ and it has a coefficient
of 0 (see (1) in figure). We update $\delta_* \gets \delta^1$ and $\delta \gets$ MAX, and we iterate
through all entries in $v$ as done previously. In (3) we get $\delta_* = \delta^2$ with non-zero
coefficient. This signals the end of implicit reduction of $r$ with $R^\bot(e'_0).$ To determine
whether $r$ is to be reduced with another column of $R^\bot,$ we check if there exists a persistence
pair $(\delta_*, e'')$ in $p^\bot$ and go back to step 2 to append $V^\bot(e'')$ to $v.$  Otherwise,
$r$ is completely reduced and we go to the next step to update $p^\bot$ and $V^\bot.$ In general, it
is possible that $\delta_*$ is empty, in which case $r$ was reduced to $\mathbf{0}$ and reduction
ends without requiring any update to $p^\bot$ and $V^\bot.$

Step 4 (`update' in Figure~\ref{fig:implicit_row_reduce}): If $\delta_*$ is not empty, then we add
the persistence pair $(\delta_*, e_0)$ to $p^\bot.$ To update $V^\bot,$ we first sum all $\phi_e$
modulo 2. This is because a $\phi_e$ with zero coefficient implies that the entire coboundary of $e$
will sum to 0 when coboundaries of all edges in $v$ are summed. For every $\phi_e$ with non-zero
coefficient, except for $\phi_{e_0},$ we append $e$ to the column $V^\bot(e_0).$ The edge $e_0$ is
not appended to $V^\bot(e_0)$ because $\phi_{e_0}$ will always have a coefficient 1 and we already
store $e_0$ in the persistence pair $(\delta_*, e_0).$ If the resulting $V^\bot(e_0)$ is empty, then
we do not make any record of it. This concludes the complete reduction of $r$ with $R^\bot.$

There are two pitfalls in the above strategy. First, if an edge $e_i$ occurs an even number of times
in $v$ (has zero coefficient in $v$), then we know that its coboundary will sum to 0 after complete
reduction of $r,$ and hence, calling \texttt{FindNextt} for each of the multiple entries of $e_i$
during reduction is redundant. The above algorithm trades the cost of computation of coefficient of
edges in $v$ with the cost of computing coboundaries.  This might be inefficient if there are many
edges that sum to 0 in $v.$  Second, computation of $\delta_*$ requires traversal across all entries
of $v$ at every step of reduction.  This does not scale well if $r$ requires a lot of reductions and
the corresponding $v$ has a large number of edges. We show next that both of these can be addressed
simultaneously when iterating over $v$ during reduction if we ensure that $\phi$-representations in
$v$ are always ordered appropriately. Further, we will use paired-indexing to optimize it, making it
an efficient strategy in terms of both memory requirement and computation time taken.

\subsubsection{Implicit Column Algorithm}\label{subsubsec:new_implicit}

\begin{figure}
  \centering
\begin{subfigure}{.20\textwidth}
  \centering
  \includegraphics[width=\linewidth]{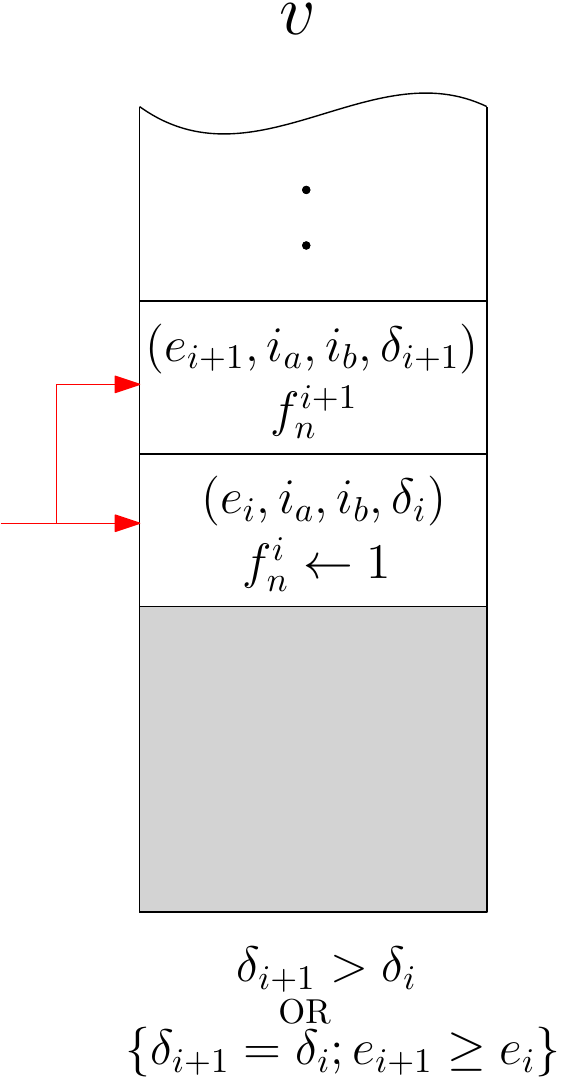}  

  \caption{New entries are now inserted in $v$ such that order is maintained as shown by the rule in
  the figure.}

  \label{fig:imp_col}
\end{subfigure}
  \centering
\begin{subfigure}{.78\textwidth}
  \centering
  \includegraphics[width=\textwidth]{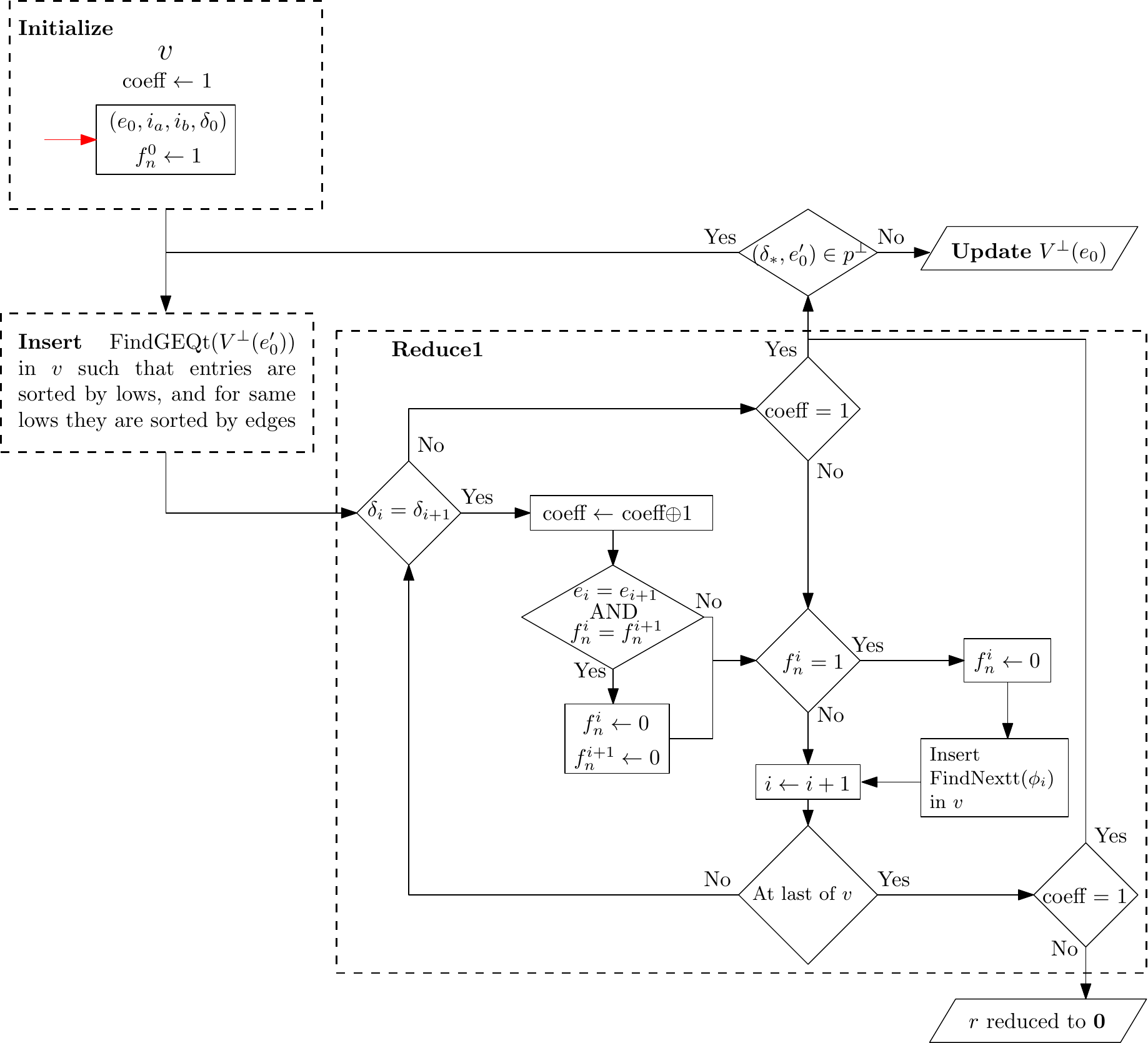}

  \caption{Flowchart to compute $\delta_*$ with non-zero coefficient.}

  \label{fig:imp_col_algo}
\end{subfigure}
  \caption{Implicit column algorithm.}
  \label{fig:implicit_new}
\end{figure}

Alternatively, we can view the $\phi$-representations in $v$ as a column of entries ordered first by
the lows ($\delta_i$) and by the edges ($e_i$) for the same lows (see
Figure~\ref{fig:implicit_new}). We say that $v$ is ordered by (low, edge). Then, entries in $v$ that
correspond to the same lows or that correspond to the same lows in the coboundaries of the same
edges, are adjacent to each other. As we move along $v,$ we sum adjacent entries and update the
coefficient accordingly. For every $\phi_i$ in $v,$ we define a flag-next, denoted by $f^i_n,$ that
is flagged if the next triangle in the coboundary of the corresponding edge has to be computed and
inserted in $v.$ We explain the algorithm below that can addresses the two pitfalls of implicit row
algorithm. The flowchart is shown in Figure~\ref{fig:implicit_new}.

The first two steps of `initialize' and `append' are similar to~\ref{subsubsec:imp_row_alg} with two
distinctions---new entries are inserted in $v$ such that it is ordered by (low, edge) and the
flag-next of every new entry is initialized with 1. To reduce $v,$ we move along it linearly using a
pointer, shown in red in Figure~\ref{fig:implicit_new}. It is initialized to point to the first
index of $v.$ The only entry in $v$ initially is $(e_0, i_a, i_b, \delta_0),$ hence, \texttt{coeff}
is initialized to 1 and $\delta_* = \delta_0.$ Suppose we point at index $i$ of $v$ at the beginning
of a reduction step. The \texttt{coeff} then has the coefficient of the entry at index $i.$ To
compute $\delta_*$ (as defined in implicit row algorithm), we compare the adjacent entries in $v$
and update the coefficient. If $\delta_i = \delta_{i+1},$ then the coefficient is summed by 1
(modulo 2). Additionally, if $e_i$ is also equal to $e_{i+1},$ we unflag $f^i_n$ and $f^{i+1}_n$ if
both are flagged. This eliminates edges $e_i$ and $e_{i+1}$ from subsequent computation of
$\delta_*,$ which is justified since coboundaries of equal edges will always sum to 0---resolving
the first pitfall discussed in~\ref{subsubsec:imp_row_alg}. Now, if $f^i_n$ is flagged, we compute
$\phi'_i =$ \texttt{FindNextt}$(\phi_i)$, unflag $f^i_n$, and insert $\phi'_i$ in $v.$ It is known
that the index of $\phi'_i$ in $v$ will be strictly greater than $i$ since it has a greater low than
$\phi_i.$ So, the insertion of an element in $v$ does not affect $v$ up to and including index $i,$
and we can by incrementing $i$ by 1. Otherwise, if $\delta_i \neq \delta_{i+1},$ we look at the
coefficient. If \texttt{coeff} $= 1,$ then $\delta_i$ has non-zero coefficient and, hence, $\delta_*
= \delta_i,$ and we are done with reduction. On the other hand, the value 0 of \texttt{coeff}
implies that $\delta_i$ has 0 coefficient. To proceed, we insert \texttt{FindNextt}$(\phi_i)$ in $v$
if $f^i_n$ is flagged, and we unflag $f^i_n.$ Additionally, we reset \texttt{coeff} to 1 to indicate
that the smallest triangle with non-zero coefficient is now $\delta_{i+1}.$ Then, we increment $i$
by one to continue computation of $\delta_*.$ A value of 0 for the coefficient of the last entry in
$v$ implies that $r$ is reduced to $\mathbf{0}.$

In implicit row algorithm, it was a requirement to traverse through entire $v$ every time the
coefficient of $\delta_*$ is 0. In implicit column algorithm also we have to traverse through $v$ to
maintain the order of the entries, but this traversal is not necessarily across all $v.$ This
potential improvement in computational efficiency is then canceled by the fact that the traversal
in implicit column algorithm has to be done every time \texttt{FindNextt} is called and a new entry
is inserted in $v.$ Further, since we insert entries in $v$ in implicit column algorithm in contrast
to replacing them in implicit row algorithm, the size of the data structure that stores $v$ is
theoretically bounded by the size of $r.$ As a result, this algorithm is not practically feasible
yet, and in the next section we will show next how paired-indexing will solve both of these issues.

\subsubsection{Fast Implicit Column Algorithm}\label{sec:hash_table_reduction}

\begin{figure}[tbhp]
  \centering
\begin{subfigure}{.40\textwidth}
  \centering
  \includegraphics[width=\linewidth]{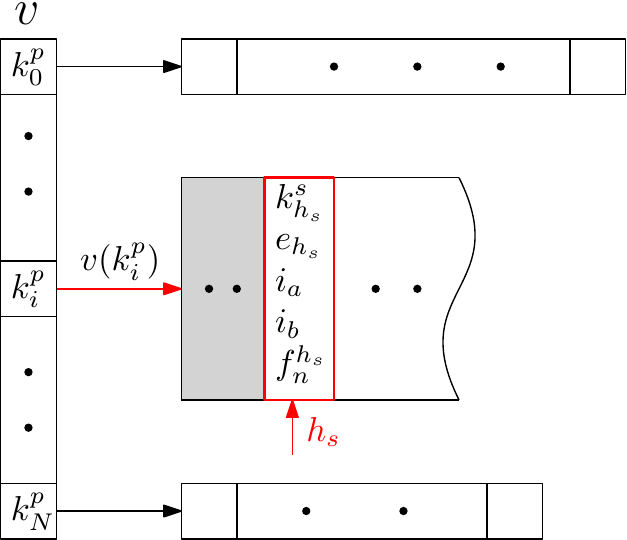}  
  \caption{$v$ is stored as a hash table with the primary keys of lows as its keys. The two red arrows
  collectively point to $\phi = (e_{h_s}, i_a, i_b, \left< k^p_i, k^s_{h_s} \right>)$}.
  \label{fig:hash1}
\end{subfigure}
  \centering
\begin{subfigure}{.55\textwidth}
  \centering
  \includegraphics[width=\linewidth]{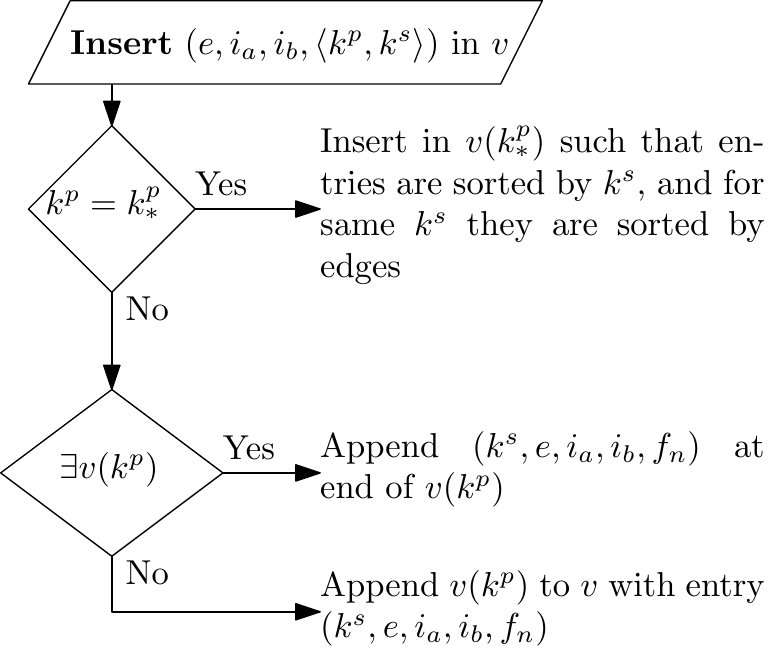}  
  \caption{Flowchart for inserting new entries in the hash table of $v.$}
  \label{fig:hash2}

\end{subfigure}
  \caption{Fast implicit column algorithm, using paired-indexing to make a hash table.}
\label{fig:hash_intro}
\end{figure}

To efficiently compute $\delta_* = \left < k^p_*, k^s_* \right>,$ the smallest triangle with
non-zero coefficient, we first observe that it will have the smallest primary key among all lows in
$v.$ Now, using paired-indexing, we can store $v$ as a hash table that is defined by a list of
unique primary keys that are in the lows of the entries in $v,$ and each entry in this list is
mapped to a list of lows in $v$ that have the same primary key. Figure~\ref{fig:hash1} shows an
example of such a hash map. The keys of this hash table are stored as a linear list of primary keys
$[k^p_i]$ (in no particular order), and each primary key is mapped to a linear list $v(k^p_*),$ with
elements of the form $(k^s, e, i_a, i_b, f_n).$ In Figure~\ref{fig:hash1}, the entry in the red box
is at index $h_s$ of $v(k^p_i)$, and it corresponds to $\phi = (e_{h_s}, i_a, i_b, \left < k^p_i,
k_{h_s}^s\right>)$ with the corresponding flag-next $f^{h_s}_n.$

\begin{figure}
  \includegraphics[width=\textwidth]{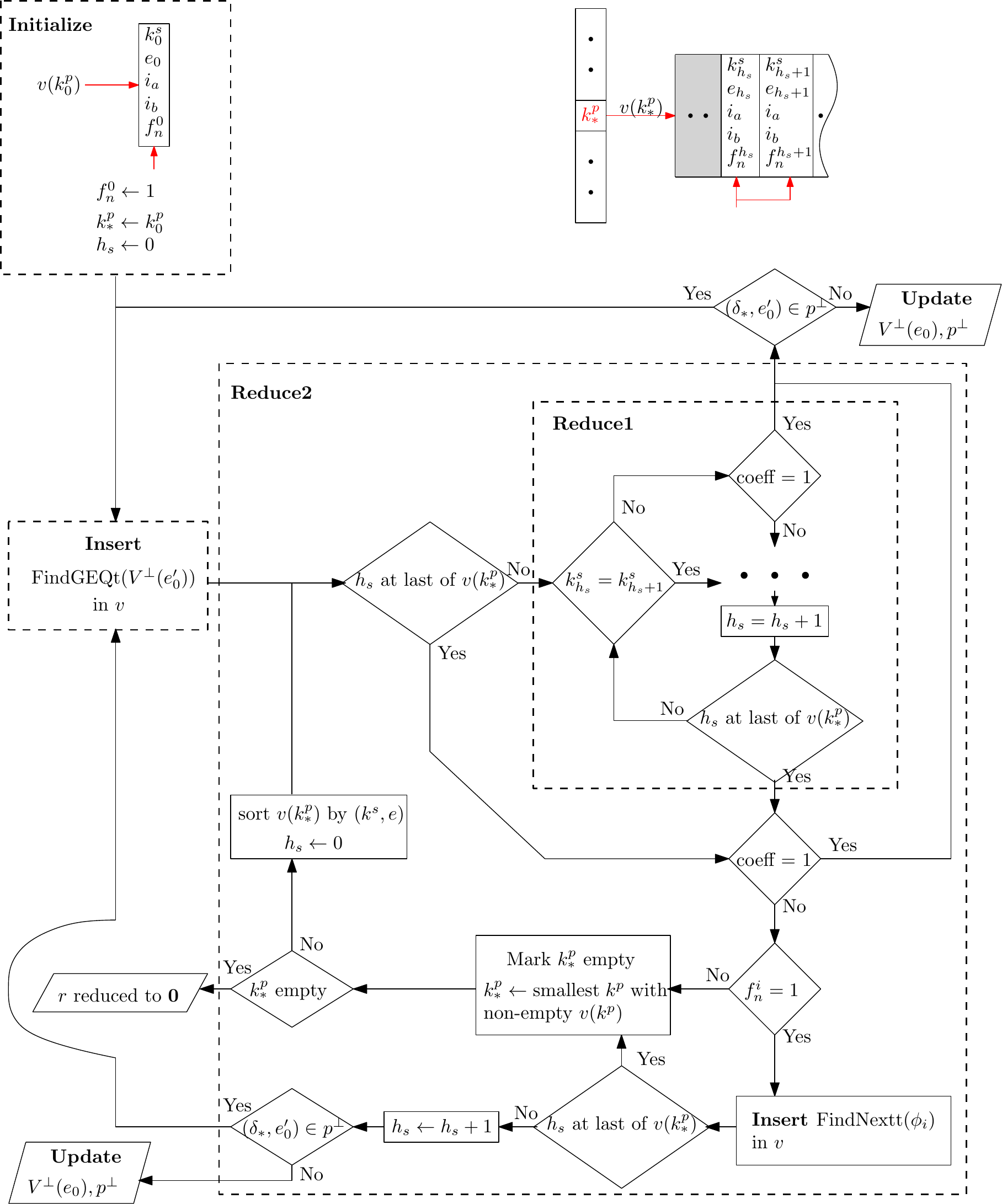}

  \caption{Flowchart for fast implicit column algorithm in which $v$ is stored as a hash table using
  paired-indexing. The three dots in box `Reduce1' denote that it is similar to the flowchart in
  Figure~\ref{fig:imp_col_algo}.}

  \label{fig:hash_reduce}
\end{figure}

To compute $\delta_*$ using this hash table, we first find the primary key with the smallest order,
denoted by $k^p_{*},$ and then compare adjacent entries in $v(k^p_{*})$ (see flowchart in
Figure~\ref{fig:hash_reduce}). That is why, the entries in $v(k^p_{*})$ have to be ordered by
secondary key and those with same secondary key have to be ordered by the edge. We denote this
ordering by (secondary key, edge). Note that, the order of elements of $v(k^p)$ for $k^p \neq
k^p_{*}$ does not matter. To ensure this, we define the insertion of a new entry in this hash table
as follows (see Figure~\ref{fig:hash2}). Suppose the new entry to be inserted is $\left (e, i_a,
i_b, \left < k^p, k^s\right > \right ).$ If $k^p = k^p_{*},$ then $(k^s, i_a, i_b, f_n=1)$ is
inserted in $v(k^p_{*})$ such that its entries are ordered according to (secondary key, edge).
Otherwise, we check whether $k^p$ exists in the list of keys of the hash table. If true, we append
$(k^s, i_a, i_b, 1)$ to the end of $v(k^p)$ since the order of its entries does not matter. If
$v(k^p)$ does not exist, then we append $k^p$ to the list of keys of the hash table, and we map
$v(k^p)$ to a list with a singular entry of $(k^s, i_a, i_b, 1).$ In this strategy, it is not
required for entries in $v(k^p)$ to be in a specific order when $k^p \neq k^p_{*},$ and we also do
not maintain the order of the list of keys of the hash table. Hence, we maintain the order only of
the entries that are crucial for computation of $\delta_*,$ that is, precisely those that have the
primary key as the smallest key of the hash table in the current reduction iteration. This optimizes
the computational overhead of maintaining the order during insertion of new entries in $v.$

Now, if $h_s$ reaches the last entry of $v(k^p_{*}),$ it does not imply that we have reached the
last entry of the hash table, and we proceed as follows. If \texttt{coeff}$ = 1,$ then $\delta_* =
\left< k^p_*, k^s_{h_s}\right>$ and we are done. Otherwise, if \texttt{coeff}$ = 0,$ then we first
check whether $f^{h_s}_n$ flagged. If it is flagged, we compute $\left (e_{h_s}, i_a, i_b, \left <
k^p_{\#}, k^s_{\#}\right >\right )=$ \texttt{FindNextt}$\left( \left(e_{h_s}, i_a, i_b, \left <
k^p_{*}, k^s_{h_s} \right > \right) \right)$ and insert it in the hash table.  If $k^p_{\#} =
k^p_{*},$ then $(k^s_{\#}, e, i_a, i_b, 1)$ is the last entry in the updated $v(k^p_{*}),$ and also,
$h_s$ does not point to the last entry in $v(k^p_*).$ We know that the next simplex in $\delta
e_{h_s}$ will be greater than $\left < k^p_{\#}, k^s_{\#}\right >,$ hence, it is not required to
compute \texttt{FindNextt} again, and $\left < k^p_{\#}, k^s_{\#}\right >$ will have a coefficient
of one. So, $\delta_* = \left < k^p_{*}, k^s_{h_s + 1} \right > = \left < k^p_{\#}, k^s_{\#}\right
>$ and we are done. Otherwise, if $k^p_{\#} \neq k^p_{*}$ or if $f^{h_s}_n$ was unflagged, we have
traversed through all entries in $v(k^p_{*})$ and we mark the memory occupied by $v(k^p_{*})$ as
free space that can be overwritten.  As a result, the memory used by the hash table will never
approach $r,$ addressing the second pitfall discussed in~\ref{subsubsec:new_implicit}. To proceed,
we iterate through rest of the keys of the hash table (that are not empty) and update $k^p_*$ with
the smallest primary key. Since $v(k^p_*)$ might not be sorted, we first sort it by (secondary key,
edge). Then, we update $h_s \gets 0$ and \texttt{coeff}$\gets 1$, and we proceed with the
computation of $\delta_*.$ If there are no keys left in the hash table, then $r$ was reduced to
$\mathbf{0}.$

The fast implicit column algorithm to compute $H_2^*$ similarly uses paired-indexing along with the
$\phi$-representation for tetrahedrons introduced in Section~\ref{subsec:cob_triangles} and the
functions \texttt{FindSmallesth}, \texttt{FindNexth}, and \texttt{FindGEQh}. It scales more
efficiently than the implicit row algorithm. See~\ref{tab:imp_row_vs_col} in
appendix~\ref{app:tab_benchmarks} for a comparison between the two algorithms for test data sets
that are used for benchmarking in this study (Section~\ref{sec:experiments}).

\subsubsection{Trivial Persistence Pairs}\label{sec:trivial_pair}

\begin{figure}[tbhp]
  \centering
\begin{subfigure}{0.49\linewidth}
  \centering
  \includegraphics[width=0.95\textwidth]{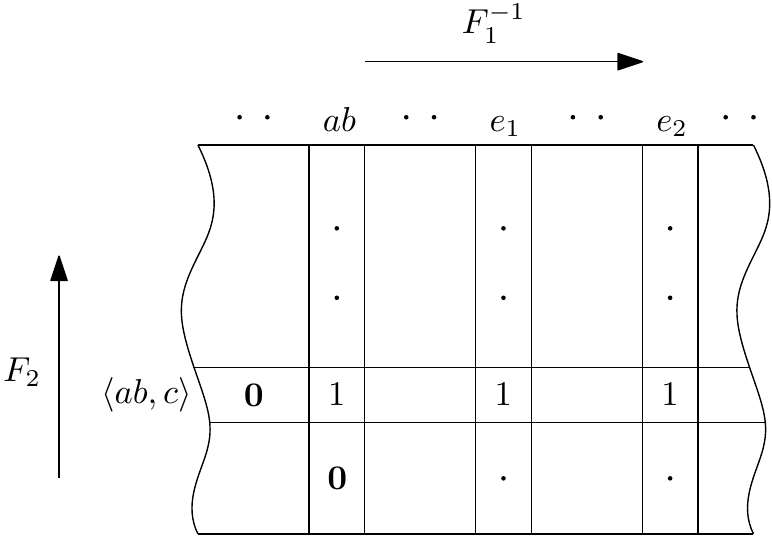}  
  \caption{If $\left <ab, c \right>$ is the smallest triangle in $\delta \{a,b\},$ then $(\left <ab,
  c \right>, ab)$ is a trivial persistence pair.}
  \label{fig:trivial_edge}
\end{subfigure}
  \centering
\begin{subfigure}{0.49\linewidth}
  \centering
  \includegraphics[width=0.95\textwidth]{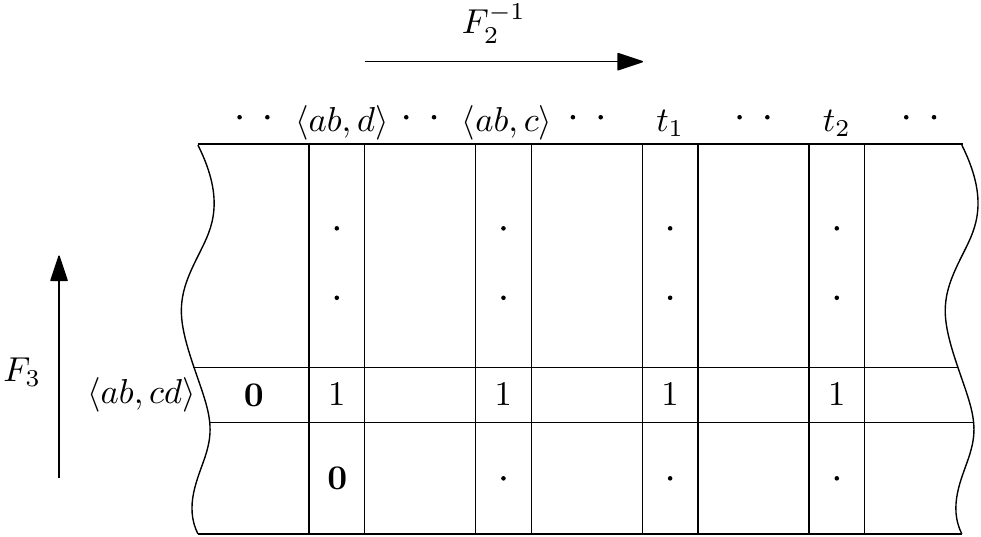}  

  \caption{If $\left <ab,d \right >$ is the greatest triangle in the boundary of $\left <ab, cd
  \right>,$ and if $\left <ab, cd \right>$ is the smallest tetrahedron in $\delta \{a,b,d\},$ then
  $(\left <ab, cd \right>, \left < ab, d\right >)$ is a trivial
  persistence pair.}

  \label{fig:trivial_triangle}
\end{subfigure}
  \caption{Trivial persistence pairs}
\label{fig:trivial_pairs}
\end{figure}

For further reduction in memory usage, we notice that there are specific persistence pairs that can
be computed on the fly and do not require storage in $p^\bot.$ Figure~\ref{fig:trivial_pairs} shows
an example of coboundary matrices for edges and triangles. The triangle $t = \left< ab, c\right>$
will be in the coboundary of exactly three edges (see Figure~\ref{fig:trivial_edge}).  Since the
diameter of $t$ is $\{a,b\},$ the row of $t$ will have all zeroes to the left of $(\left< ab,
c\right>, ab).$ If additionally, $t$ is the smallest simplex in the coboundary of $\{a,b\},$ then
there will be all zeroes below $(\left< ab, c\right>, ab).$ Consequently, $(\left< ab, c\right>,
ab)$ will be a persistence pair and will not require any reduction. We will call such pairs
\textit{trivial persistence pairs}, and we will not store them in $p^\bot.$ Instead, during the
reduction of $r,$ we check whether $\delta_* = \left< k_1, k_2 \right>$ is the smallest simplex in
the coboundary of the edge $e' = f_1^{-1}(k_1)$. If true, then $(\delta_*, e')$ is a trivial
persistence pair and the next reduction is to be with exactly the coboundary of $e'$ (since trivial
persistence pairs do not require any reductions). Conducting this check at every step of reduction
is computationally feasible because paired-indexing gives direct access to the diameter of $t,$
making these checks inexpensive. For further optimization, the smallest simplex in the coboundary of
each edge is stored a priori at the cost of $O(n_e)$ memory. 

Similarly, a tetrahedron, $h = \left< ab, cd\right>,$ will be in the coboundary of exactly four
triangles (see Figure~\ref{fig:trivial_triangle}). The greatest triangle in the boundary of $h$ will
be $t = \left< ab, \text{max}\{c, d\} \right>.$ Following similar reasoning, if $h$ is the smallest
simplex in the coboundary of $t,$ then we say that $(h, t)$ is a trivial persistence pair. During
reduction of any triangle, we check whether $\delta_* = \left< k_1, k_2 \right>$ is the smallest
simplex in the coboundary of the triangle $t' = \left<k_1, \text{max}\{f_1^{-1}(k_2)\}\right>$. If
true, $(\delta_*, t')$ is a trivial persistence pair and the next reduction is to be with the
coboundary of $t'.$ \texttt{FindSmallesth} is used to check whether $\delta_*$ is the smallest
simplex in $\delta t'.$

\subsection{Parallelizing: A General Serial-parallel Reduction Algorithm}

\begin{figure}
  \centering
  \includegraphics[scale=0.75]{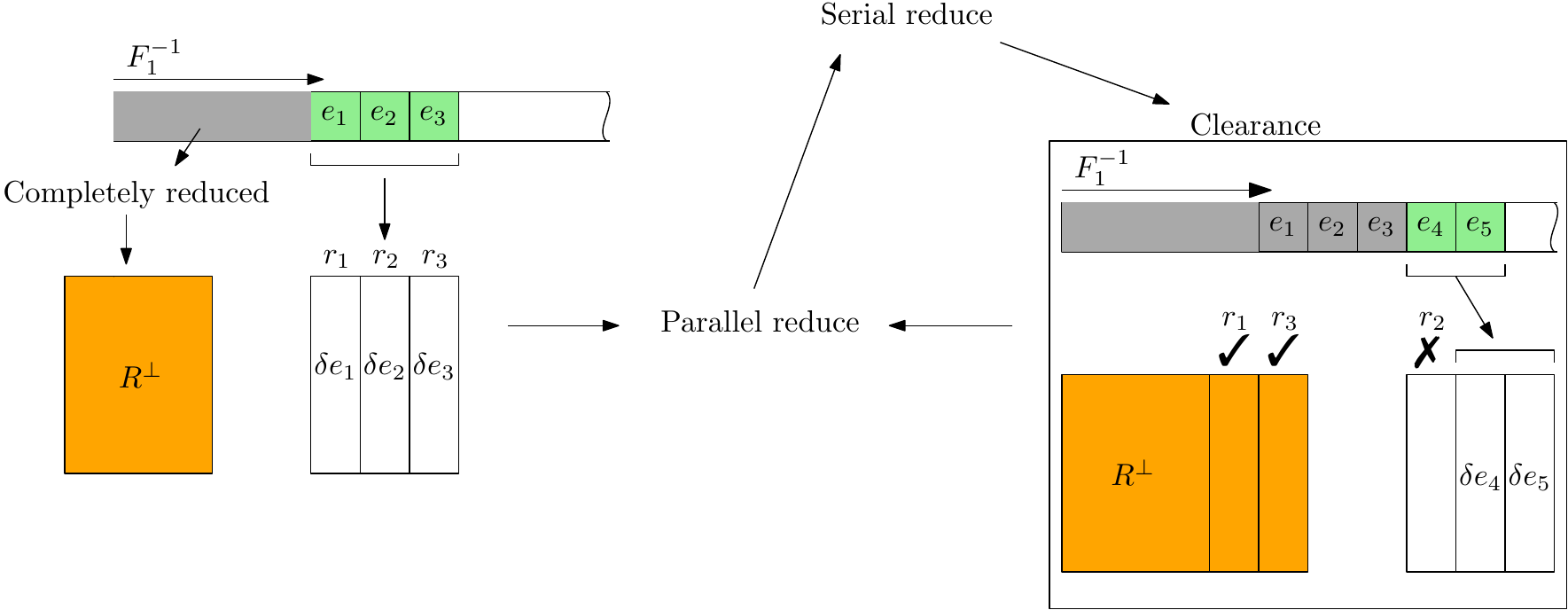} 

  \caption{Serial-parallel algorithm to reduce coboundaries of edges. Similar algorithm structure is
  used to parallelize reduction of coboundaries of edges (H$_1^*$) and triangles (H$_2^*$) and to
  reduce boundaries of edges (H$_0$).}

\label{fig:serial_parallel}
\end{figure}

To reduce the computation time required to process the large number of simplices in the
VR-filtration, we developed a novel serial-parallel algorithm that can reduce multiple simplices in
parallel. In essence, rather than reducing one simplex at a time, we will reduce a batch of
simplices. Any algorithm cannot be embarrassingly parallel because of the inherent order in
reduction imposed by the filtration. We introduce the serial-parallel algorithm by parallelizing the
standard column algorithm (algorithm~\ref{alg:basic_red1}) for cohomology computation (see
Figure~\ref{fig:serial_parallel}).

\begin{figure}[tbhp]
  \centering
  \begin{subfigure}{0.49\linewidth} 
    \centering
  \includegraphics[width=0.95\textwidth]{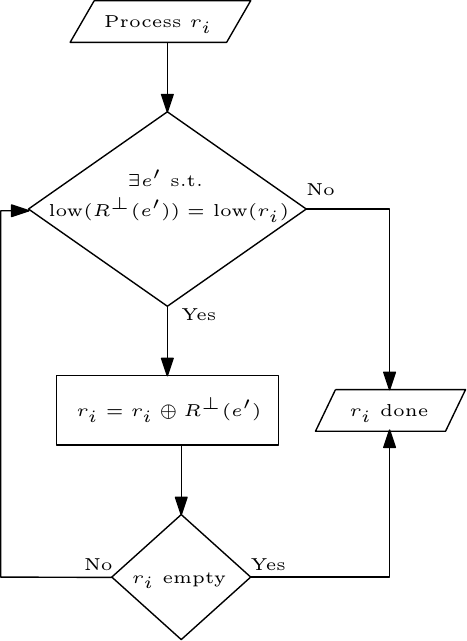}

  \caption{Parallel reduction: Each $r_i$ is reduced with $R^\bot$ independently, in parallel.}

\label{fig:parallel_flowchart}
\end{subfigure}
\centering
\begin{subfigure}{0.49\linewidth}
  \centering
  \includegraphics[width=0.95\textwidth]{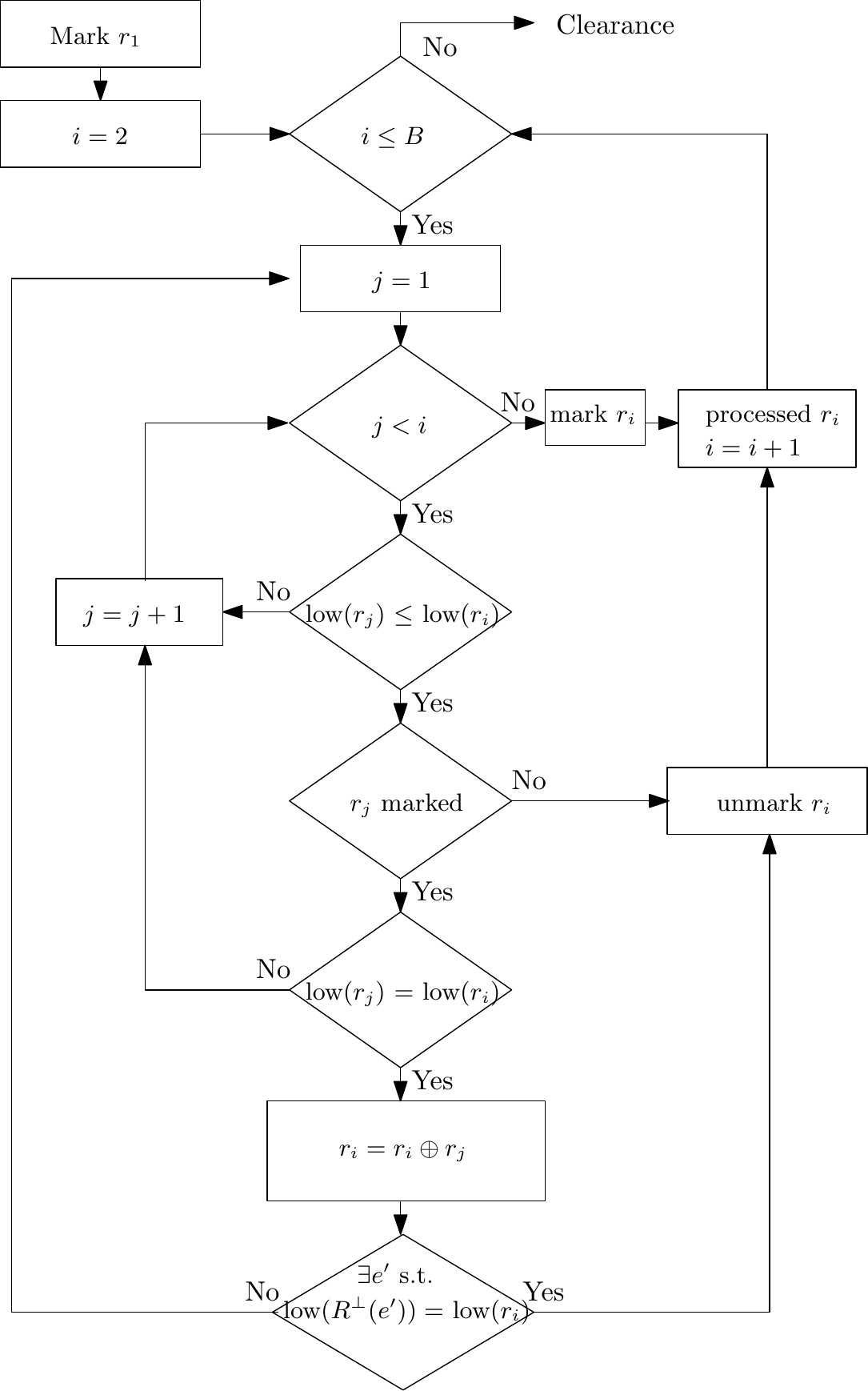} 

  \caption{Serial reduction: After parallel reduction, columns in $\mathbf{r}$ are reduced with each
  other.}

\label{fig:serial_flowchart}
\end{subfigure}

  \caption{Flowcharts for parallel and serial reductions that form the serial-parallel algorithm.}
  \label{fig:serial_parallel_flowcharts}
\end{figure}

Parallel (Figure~\ref{fig:parallel_flowchart}): Suppose $R^\bot$ contains reductions of the first
$n$ edges in $F^{-1}.$ Let $\mathbf{r} = [r_1, ..., r_B]$ be the batch of next $B$ edges in
$F_1^{-1}$ that have to be reduced. Initially, each $r_i$ is the coboundary of the corresponding
edge, and during reduction it contains the result of the ongoing reduction. Then, if the low of
$r_i$ and $r_j$ in $\mathbf{r}$ is equal to that of $R^\bot(e)$ for some edge $e,$ reducing $r_i$
and $r_j$ with $R^\bot(e)$ will take precedence over reducing them with each other. As a result,
each $r_i$ can be reduced with $R^\bot$ in parallel.

\begin{figure}
  \includegraphics[width=\textwidth]{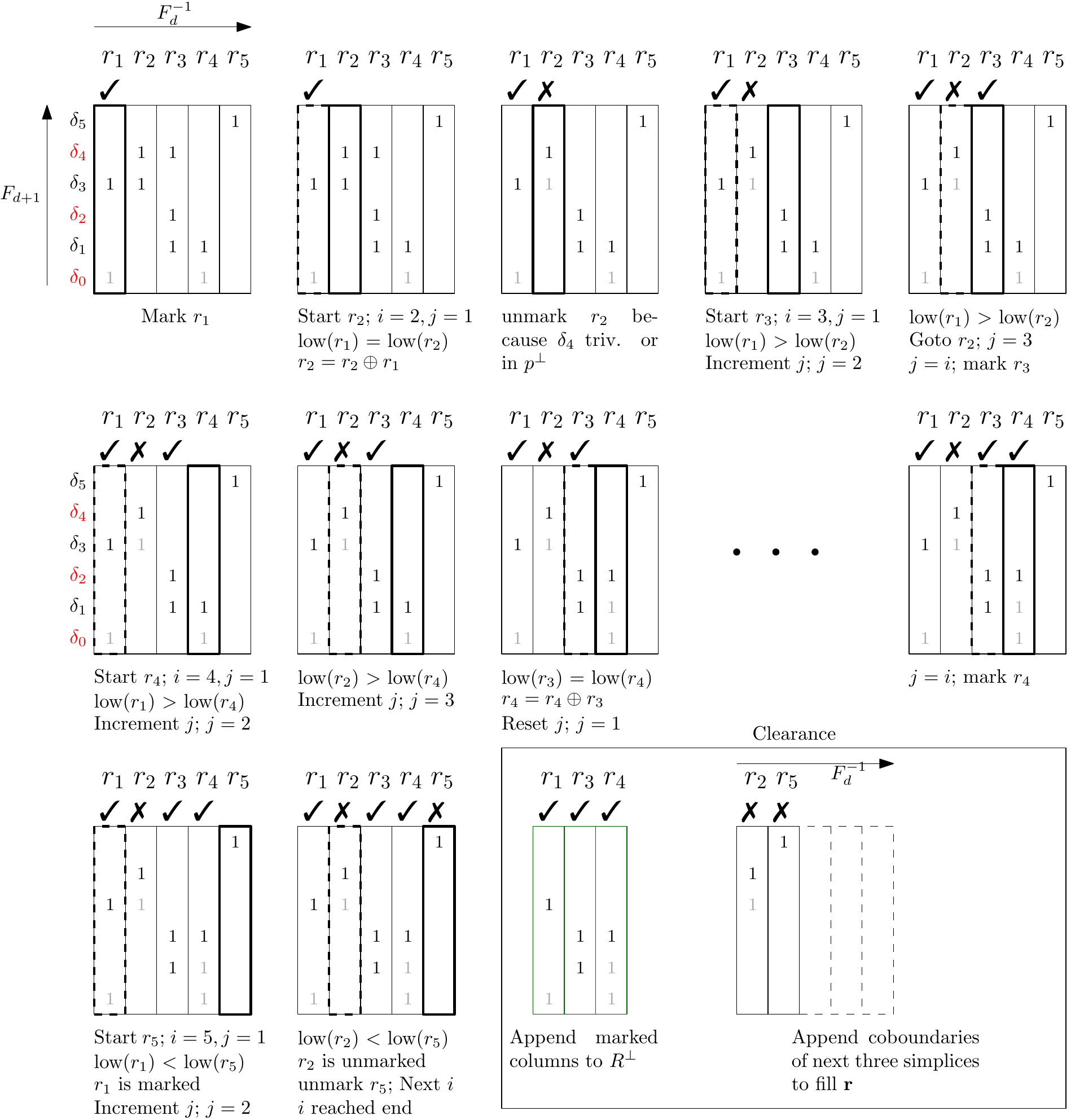}
  \caption{Hypothetical example to illustrate serial reduction in serial-parallel algorithm. The simplices marked by red indicate that they are in a persistence
  pair in $p^\bot.$}
  \label{fig:serial_stepbystep}
\end{figure}

Serial (Figure~\ref{fig:serial_flowchart}): After $\mathbf{r}$ has been reduced with $R^\bot,$ we
reduce the columns in $\mathbf{r}$ with each other using \textit{serial reduction}. We `mark' $r_i$
if it is completely reduced. After parallel reduction, none of the $r_i$ have the same low as any
column in $R^\bot.$ Hence, $r_1$ is completely reduced and we mark it. Then, carrying out the
standard column algorithm, the low of every $r_i$ ($i>1$) is compared sequentially with $r_j$ for
every $j <i .$ We consider the following cases for every $r_j.$ (1) If low($r_j$) $>$ low($r_i$),
then we skip it. (2) Otherwise, we check whether $r_j$ is marked. (2a) If it is unmarked, we cannot
continue reduction of $r_i$ before reducing $r_j$ so we unmark $r_i$ and increment $i$ by one.  (2b)
If $r_j$ is marked and low($r_j$) $=$ low($r_i$), then we reduce $r_i$ with $r_j,$ and we check
whether the new low of $r_i$ is also the low of a column in $R^\bot.$ If true, then $r_i$ has to be
reduced with $R^\bot$ before reducing with any $r_j$ in $\mathbf{r}$, and we unmark $r_i$ and
increment $i$ by one. Otherwise, if the updated low is not the low of any column in $R^\bot,$ then
$j \to 1.$ This process is repeated to reduce $r_i$ until either $j$ reaches $i$ or $r_i$ is
unmarked.  When all of $\mathbf{r}$ has been processed, that is, $i$ reaches the end of
$\mathbf{r},$ we go to clearance.  A detailed hypothetical example is shown in
Figure~\ref{fig:serial_stepbystep}.

Clearance: All marked $r_i$ are appended to $R^\bot,$ freeing up space in $\mathbf{r}$ to be filled
in by the coboundaries of the next edges from $F_1^{-1}.$ After filling $\mathbf{r}$ with the new
coboundaries, we go back to parallel reduction. This process is continued until all edges in
$F_1^{-1}$ have been reduced. The structure of the algorithm is shown in
Figure~\ref{fig:serial_parallel}.

\begin{figure}[tbhp]
  \centering
\begin{subfigure}{0.49\linewidth}
  \centering
  \includegraphics[width=0.95\textwidth]{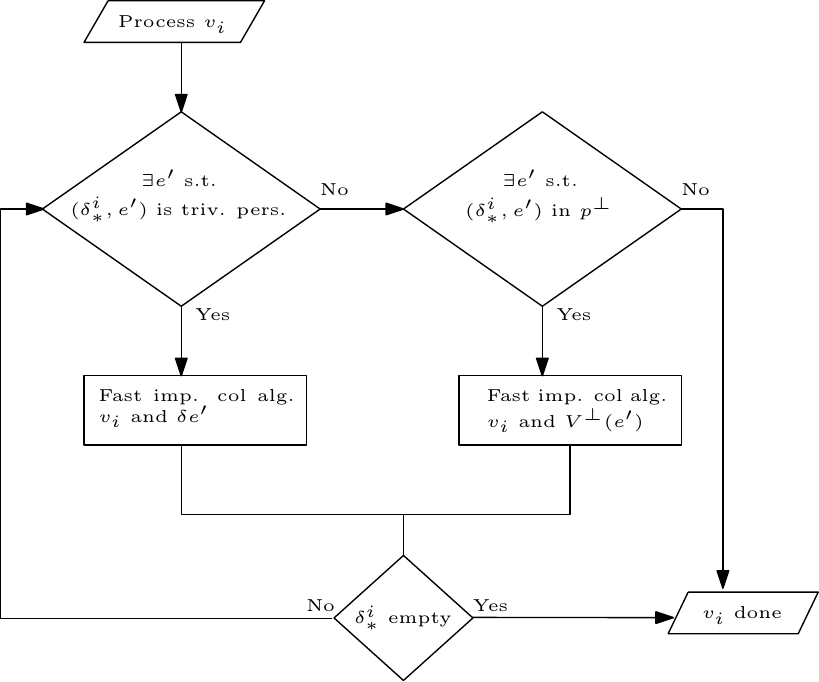}  
  \caption{Implicit reduction of a single $r_i$ in the parallel reduction.}
  \label{fig:parallel_flowchart_implicit}
\end{subfigure}
  \centering
\begin{subfigure}{0.49\linewidth}
  \centering
  \includegraphics[width=0.95\textwidth]{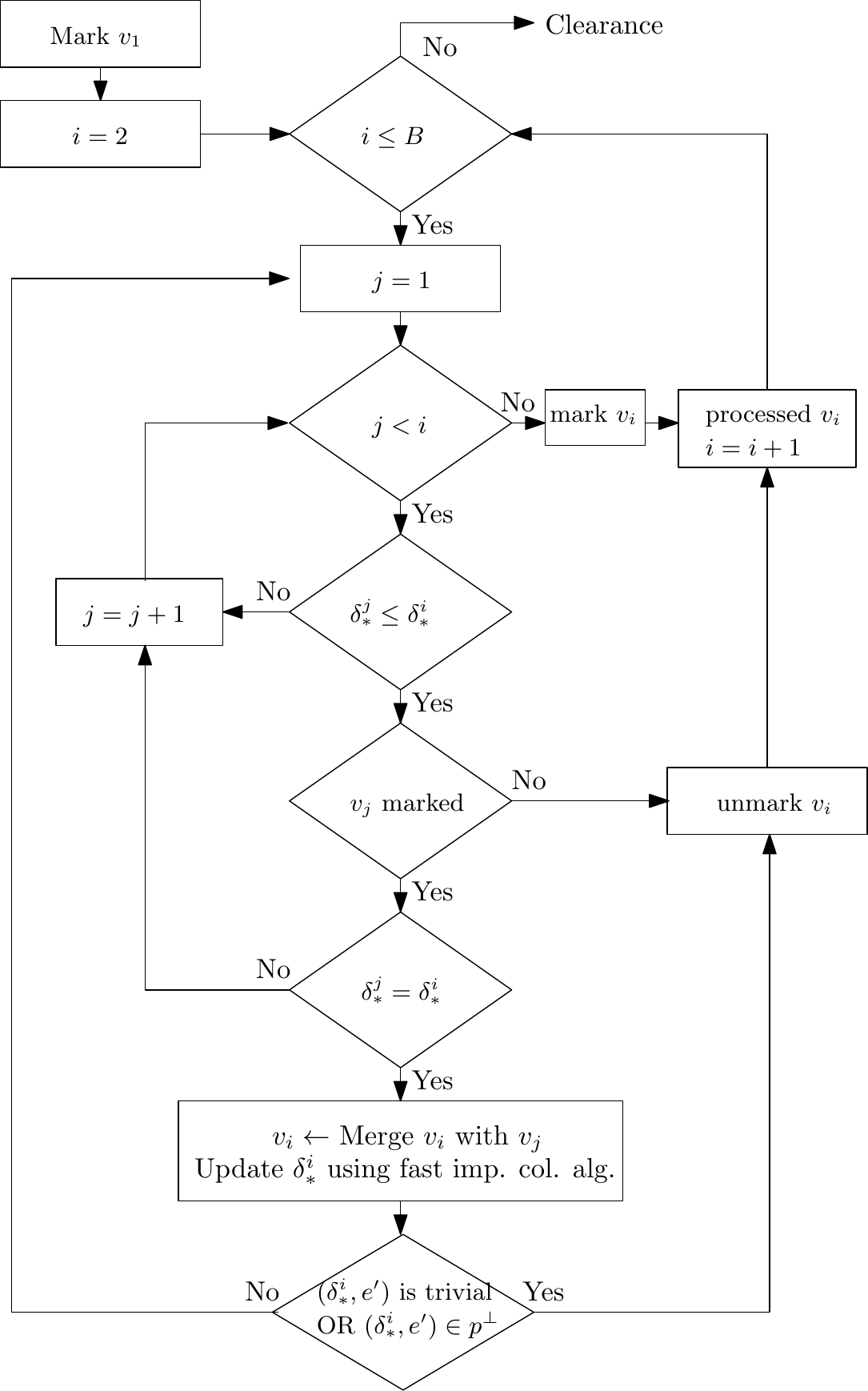}  
  \caption{Serial reduction.}
  \label{fig:serial_flowchart_implicit}
\end{subfigure}

  \caption{Flowcharts for algorithms in serial-parallel reduction using fast implicit column
  algorithm and incorporating trivial persistence pair computation.}

\label{fig:serial_parallel_flowcharts_implicit}
\end{figure}

In Dory, we implement the serial-parallel algorithm to compute H$_0,$ H$^*_1,$ and H$^*_2.$ We
discuss the implementation for cohomology computation in detail. The batch to be reduced is
represented by $\mathbf{v} = [v_1,..., v_B],$ and we store $V^\bot$ and $p^\bot.$ Then, reducing
$r_i$ with $R^\bot$ is implemented using the fast implicit column algorithm using $v_i$ and
$V^\bot,$ as shown earlier. The implicit reduction of $r_i$ with $r_j$ using $v_i$ and $v_j$ is
modified slightly because $v_j$ stores the $\phi$-representations and \texttt{FingGEQh} is not
required when appending $v_j$ to $v_i.$ Suppose $k^p_{j*}$ is the smallest primary key in $v_j$ and
$h^j_s$ is the index in $v(k^p_j)$ that points to the smallest low with non-zero coefficient. Then,
merging $v_i$ with $v_j$ is done by inserting every entry of $v_j$ in $v_i,$ except for the entries
that are in $v(k^p_{j*})$ before the index $h^j_s.$ Additionally, trivial persistence pairs are not
stored and are computed on the fly in the parallel reduction. The modified flowcharts are shown in
Figure~\ref{fig:serial_parallel_flowcharts_implicit}. For different scenarios during serial
reduction, we use four different flags for every $r_i$---$f_v$ flags whether $r_i$ has to be reduced
with a trivial persistence pair; $f_r$ flags whether $r_i$ is to be reduced with $R^\bot$; $f_a$ is
flagged if it is completely reduced; and  $f_e$ is flagged if $r_e$ is empty. Note that flagging
$f_v$ and $f_r$ is similar to unmarking $r_i,$ and flagging $f_a$ is similar to marking $r_i.$ The
basic flow of the serial-parallel algorithm to reduce triangles is shown in
algorithm~\ref{alg:cohom_reduce_serial_parallel}. See appendix~\ref{app:serial_parallel} for
pseudocode of serial (algorithm~\ref{alg:serialreduce}) and parallel
(algorithm~\ref{alg:parallelreduce}) reduction. The default value of the batch-size for
serial-parallel implementation of H$^*_2$ computation is chosen as 100 and for H$_0$ it is 1000.
These hyperparameters can be optimized for any data set. A smaller value of batch-size can increase
the computation time by lowering the number of simplices that are reduced in parallel, but a larger
value of batch-size can increase the computation time spent in the serial reduction. The
parallelized section of the serial-parallel algorithm will be called many times in the
serial-parallel reduction. To reduce the overhead of creating and destroying threads, we create
threads before the computation of PH. The jobs are allocated in fixed chunks to these threads and
the threads are woken up when they are required and destroyed after computaton of PH. We use
light-weight \texttt{POSIX} threads, or pthreads.

\subsection{All Together with Clearing Strategy}

We summarize the algorithm to compute the persistence pairs for groups H$_0, \text{H}^*_1, $ and
H$^*_2$ in algorithm~\ref{alg:computing_all}, including the clearing strategy suggested
by~\citet{chen2011persistent}. This strategy provides significant reduction in computation time of
cohomology reduction by eliminating the need for reduction of certain simplices. In essence, if
$(\tau, \sigma)$ is a persistence pair in H$_d$ (or $(\sigma, \tau)$ is in H$^*_d$), then there
cannot exist a persistence pair $(\sigma, \omega)$ in H$_{d+1}.$ Consequently, there cannot exist a
pair $(\omega, \sigma)$ in H$^*_{d+1},$ and there is no need to reduce the coboundary of $\sigma$
when computing H$^*_{d+1}.$

\begin{algorithm}
  \begin{algorithmic}[1]

    \State \textbf{Input:} $F_1$
    \State \textbf{Output:} Persistence pairs in $\text{H}_0, \text{H}^*_1, \text{H}^*_2$

  \item

    \For{$e$ in $F_1$}\Comment{Compute H$_0$}
      \State Serial-Parallel reduction
    \EndFor

  \item

    \For{$e$ in $F^{-1}_1$}\Comment{Compute H$^*_1$}
        \If{$e$ is in a persistence pair in H$_0$}\Comment{Clearing strategy}
            \State \textbf{continue}
        \EndIf
        \State Reduction
    \EndFor

  \item

    \For{$e$ in $F^{-1}_1$}\Comment{Compute H$^*_2$}
        \For{$t$ in $\delta e$ with $d(t) = e$}
            \If{$t$ is in a persistence pair in H$^*_1$}\Comment{Clearing strategy}
                \State \textbf{continue}
            \EndIf
            \State Serial-parallel reduction
        \EndFor
    \EndFor

  \end{algorithmic}
  \caption{Computing all}
  \label{alg:computing_all}
\end{algorithm}

\subsection{Sparse vs. Non-Sparse}

In the case of non-sparse filtrations, our experiments showed that most of the computation time was
being spent on the binary searches in edge- and vertex-neighborhoods to find orders of the edges
during cohomology computation. We implemented an alternate version in which we use combinatorial
indexing to store the orders of all edges in the filtration. This reduces computation time by
replacing binary search by an array access at the cost of using $O(n^2)$ memory instead of
$O(n^2_e).$ We call the non-sparse version DoryNS. In most cases it is advisable to use Dory because
the reduction in peak memory usage outweighs the computation cost. DoryNS should be considered when
computing H$_2$ for non-sparse filtrations.

\section{Computation and Benchmarks}\label{sec:experiments}

\begin{table}
\begin{center}
  \resizebox{0.7\textwidth}{!}{
  \begin{tabular}{|c|c|c|c|c|c|} 
 \hline
    Data set  & $n$ & $\tau_m$ & $n_e$  & $d$ &  $N$\\
    \hline

    dragon    & 2000  & $\infty$ & 1999000 & 1 & 1333335000\\ 
   \hline
   fractal   & 512   & $\infty$ & 130816  & 2 & 2852247168 \\
   \hline
   o3        & 8192  & 1        & 327614 & 2 & 33244954    \\
   \hline
   torus4(1) & 50000 & 0.15     & 2242206 & 1 & 41629821   \\
   \hline
 torus4(2) & 50000 & 0.15     & 2242206 & 2 & 454608895    \\
   \hline
    Hi-C (control) &3087941 & 400 & 51233398& 2 & 110946257 \\
   \hline
    Hi-C (auxin)&3087941 & 400 & 35170863& 2 & 36012219 \\
   \hline
  \end{tabular}}
\end{center}
  \caption{Data sets used for benchmarking in this study.}
\label{tab:datasets}
\end{table}

We tested Dory with six data sets---dragon, fractal, o3, torus4, Hi-C control, and Hi-C auxin
(Table~\ref{tab:datasets}). The data sets dragon and fractal were used in~\citet{otter2017roadmap}
to benchmark PH algorithms, o3 and torus4 are taken from the repository of
Ripser~\citep{bauer2019ripser}, and the Hi-C data sets are from~\citet{rao2017cohesin}. They are
briefly described as follows---dragon is a point-cloud in 3-dimensional space; fractal is the
distance matrix for nodes in a self-similar network; o$3$ consists of 8192 random orthogonal $3
\times 3$ matrices, that we consider as a point-cloud in nine dimensional space; torus4 is a
point-cloud randomly sampled from the Clifford torus $S^1 \times S^1 \subset \mathbb{R}$; and the
two Hi-C data sets are correlations matrices for around three million points (see
Section~\ref{sec:human_gene} for more details). The benchmarks for PH computation of torus4 data set
up to and including H$^*_1$ are shown as torus4(1) and up to and including H$^*_2$ are shown as
torus4(2). The data sets can be found at \texttt{https://github.com/nihcompmed/Dory}.  All
computations are done on a computer with a 2.4 GHz 8-Core Intel Core i9 processor and 64 GB memory.
We first comment on the computation time and memory taken by Dory.

\begin{table}
\begin{center}
  \resizebox{0.85\textwidth}{!}{
    \begin{tabular}{|c|c|c|c|c|c|} 
      \hline
      Data set & Creating $F_1$ & Creating $N^v, E^v$ & H$_0$ & H$^*_1$ &  H$^*_2$\\
      \hline
      dragon & 1.14 & 0.488 & 0.144 & 0.36 & NA\\
      \hline
      fractal & 0.09 & $\approx 0$ & 0.01 & 0.03 & 30.5 \\
      \hline
      o3 & 0.4 & 0.03 & 0.03 & 1.4 & 2.9 \\
      \hline
      torus4 & 5.74 & 0.66 & 0.25 & 1.95 & 30.68 \\
      \hline
      Hi-C (control) & 26.47 & 17.29 & 41.17 & 106.98 & 68.94 \\
      \hline
      Hi-C (auxin) & 17.64 & 10.29 & 21.57 & 20.35 & 15 \\
      \hline
  \end{tabular}}
\end{center}
  \caption{Time taken in seconds for different processes in Dory (4 threads and without \texttt{-D
  PRINT} compiler flag so that it does not print the persistence pairs in the terminal window).}
\label{tab:time_taken_each_process}
\end{table}

The computation time taken by Dory can be split into following processes---create $F_1,$ create the
vertex- and edge-neighborhoods, and compute the persistence pairs (see
Table~\ref{tab:time_taken_each_process} for results with Dory using 4 threads). In the dragon data
set, almost half of the computation time is spent in creating $F_1.$ We believe that this can be
significantly improved since we call a memory allocation for every edge in the current version of
Dory. The other process in Table~\ref{tab:time_taken_each_process} that takes a significant amount
of time is the computation of H$^*_2.$ This follows from the facts that there are generally more
simplices to be processed as compared to H$^*_1$, and that all computations of coboundaries of
triangles require binary searches, whereas, computation of coboundary of an edge does not require
binary search when in Case 1 (Section~\ref{sec:cob_edges}). We improved the latter by implementing
the non-sparse version, DoryNS at the cost of higher peak memory usage (see
Table~\ref{tab:all_benchmarks}).  The former might be improved by determining special classes of
persistence pairs a priori~\citep{bauer2019ripser} that have been shown to not require any
reductions, but their implementation and impact on computational feasibility within Dory's framework
is not yet clear and is an ongoing work.

The memory taken can be split into two parts---the base memory that is used by $F_0, F_1,$ and the
vertex- and edge-neighborhoods, and the PH-memory used for homology and cohomology computations. The
practical difference between these is that the base memory is known before computation of PH to be
exactly $(3n + 12n_e)\times 4$ bytes in Dory for a data set with $n$ points and $n_e$ permissible
edges in the filtration (see appendix~\ref{app:tab_benchmarks} for this calculation), but PH-memory
can be significantly more depending upon the reduction operations that need to be stored.  Hence,
the scaling of PH-memory can decide the feasibility of computation of PH for a data set.

Since Ripser outperforms both Gudhi and Eirene in computing PH for VR-filtrations, we compare time
taken and peak memory usage between Dory and Ripser in this section (see
Table~\ref{tab:all_benchmarks}), and the results with Gudhi and Eirene can be found in
appendix~\ref{app:tab_benchmarks}. The Ripser source file was downloaded from
\texttt{https://github.com/Ripser/ripser} and compiled using \texttt{c++ -std=c++11 ripser.cpp -o
ripser -O3}.  DoryNS was compiled using the additional flag \texttt{-D COMBIDX}. Both were executed
in the terminal of macOS (v 10.15.7). The computation time is the `total' as reported by the command
\texttt{time} and the peak memory usage was recorded using the application Instruments (v 12.2) in
macOS in a separate execution.

\begin{table}
\begin{center}
  \resizebox{\textwidth}{!}{
  \begin{tabular}{|c|c|c|c|c|c|} 
 \hline
    \multirow{2}{*}{Data set}     &\multirow{2}{*}{Ripser}
   &\multicolumn{2}{c|}{Dory} & \multicolumn{2}{c|}{DoryNS} \\
    \cline{3-6}& & 4 thds. & 1 thd.  & 4 thds. & 1 thd. \\
 \hline

    dragon    & \textbf{(2.57 s, 199 MB)} & (2.8 s, 262 MB )  & (3.3 s, 270 MB ) & (3.16 s, 277 MB)
    & (3.59 s, 269 MB)\\
   \hline
    fractal   & \textbf{(7.54 s, 775.13 MB)} & (31.9 s, 695 MB)  & (40.5 s, 687 MB) & (21.6 s, 810
    MB) & (23.8 s, 802.8 MB)\\
   \hline
    o3        & (6.47 s, 219 MB   ) & \textbf{(5.07 s,  157 MB)} & (6.43 s, 149 MB) & (4.86 s, 285
    MB) & (6.2 s, 277 MB)\\
   \hline
    torus4(1) & (31.54 s, 12 GB   ) & \textbf{(8.83 s, 328 MB)}  & (10.9 s, 321 MB) & (11.7 s, 5 GB
    ) & (13.6 s, 5 GB)\\
   \hline
    torus4(2) & (94 s, 12 GB      ) & \textbf{(40.74 s,  1 GB)}  & (59.3 s, 1 GB  ) & (42.7 s, 5.7
    GB) & (60 s, 5.7 GB)  \\
   \hline
    Hi-C (control) & NA &\textbf{(276 s, 6.23 GB)}  & (540 s, 6.21 GB) & NA & NA  \\
   \hline
    Hi-C (Auxin) & NA &\textbf{(123 s, 3.98 GB  )} & (230 s, 3.98 GB )& NA & NA  \\
   \hline

  \end{tabular}}
\end{center}

  \caption{Time taken (seconds) was measured using the command \texttt{time} in the terminal of macOS.
  Peak memory usage is recorded using the application Instruments in macOS. The highlighted results
  indicate efficient performance.}

\label{tab:all_benchmarks}
\end{table}

We first address the benchmarks in Table~\ref{tab:all_benchmarks} in which Ripser does better than
Dory. For the dragon data set, Dory is slower than Ripser by $\approx 0.2$ s and takes $80$ MB more
memory. The former can be attributed to the inefficiency of creating $F_1$ in Dory for which it
takes almost half the total runtime for this data set. The higher peak memory usage can arguably be
explained by the differences in the base memory because Dory additionally creates vertex- and
edge-neighborhoods. However, as the size of the data set increases, the PH-memory will generally
define the peak memory usage. For the fractal data set, Dory is slower than Ripser by a factor of 3
because Ripser identifies a large number of simplices that do not require any reduction during
H$^*_2$ computation of this data set.  Implementation of this strategy in Dory is an ongoing work.
DoryNS is faster than Dory, but it is still slower than Ripser for this data set. It is advisable to
use Ripser to compute H$^*_2$ for small data sets that are non-sparse.

For larger data sets with sparse filtrations, Dory outperforms Ripser in both computation time and
memory taken, and it extends PH computation to data sets with millions of points. For example, to
compute persistence pairs for dim-1 for the torus4 data set, Dory takes only $328$ MB as compared to
$12$ GB taken by Ripser, and it is also faster by a factor of 3. To compute persistence pairs up to
and including dim-2 for the same data set, Dory takes $1$ GB in contrast to $12$ GB taken by Ripser,
and it is also faster by a factor of 2. To show an application of computing PH of a data set with
millions of points, we compute PH up to and including dim-2 for the Hi-C, control and auxin, data
sets. These data sets are defined by a distance matrix that is stored in a sparse format. Ripser
crashed giving an overflow error, possibly due to combinatorial indexing of tetrahedrons in a data
set with millions of points.  Ripser-128bit (\texttt{https://github.com/Ripser/ripser/tree/128bit})
is an implementation of Ripser that uses 128-bit \texttt{int}, and hence, it is technically able to
encode the filtration on these data sets using combinatorial indexing. It did not give an overflow
error, but we stopped the simulation after waiting for two hours. Dory, on the other hand, took (276
s, 6.23 GB) for Hi-C control and (123 s, 3.98 GB) for Hi-C auxin data set. Also, Dory performs
consistently better with 4 threads across all data sets as compared to one thread, reducing the
computation time by up to a factor of 2 for the Hi-C data sets with negligible increase in the peak
memory usage.

\begin{figure}[tbhp]
\begin{subfigure}{.3\textwidth}
  \centering
  \includegraphics[width=\textwidth]{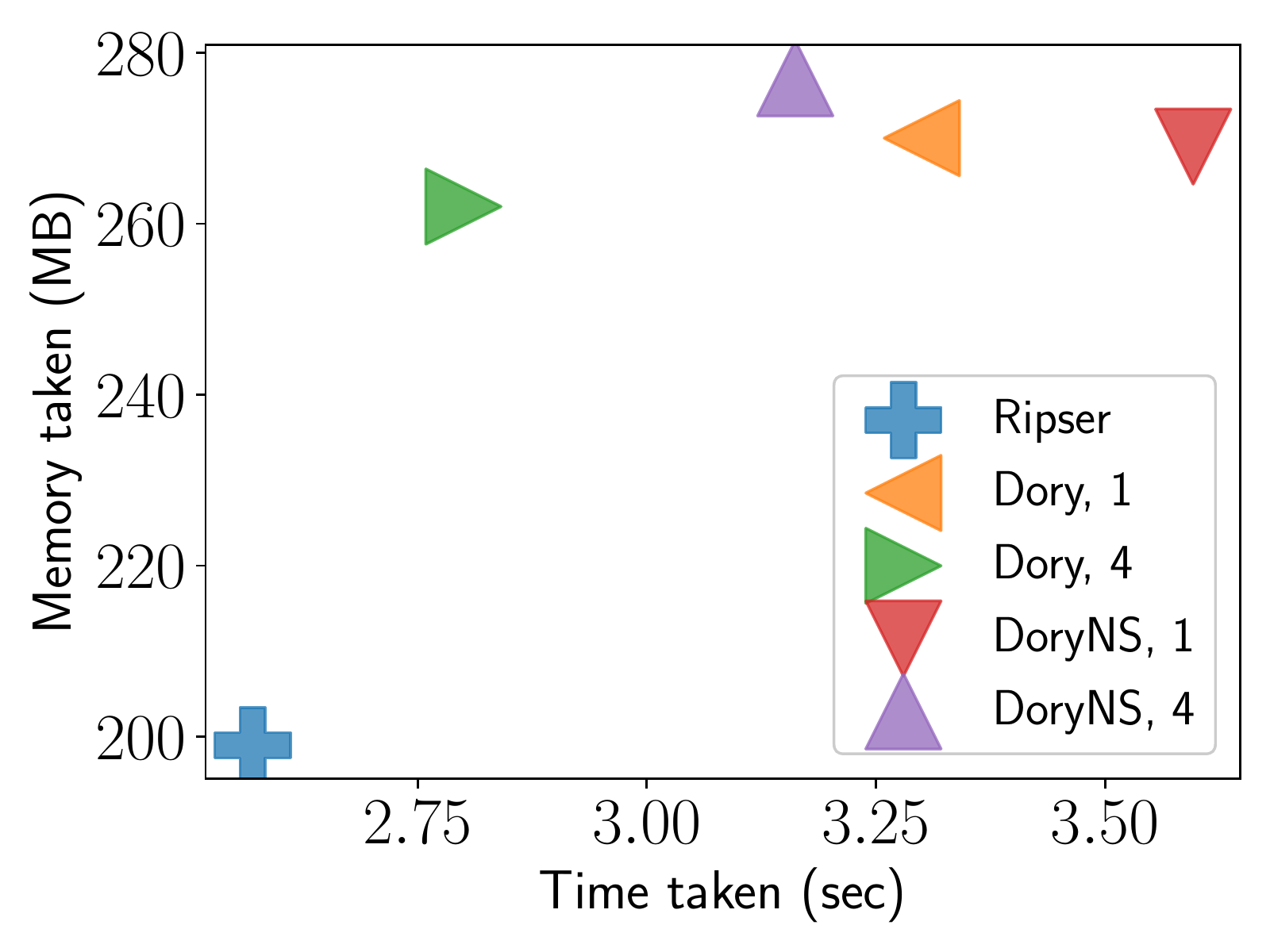}  
  \caption{dragon}
  \label{fig:dragon_bench}
\end{subfigure}
\hfill
\begin{subfigure}{.3\textwidth}
  \centering
  \includegraphics[width=\textwidth]{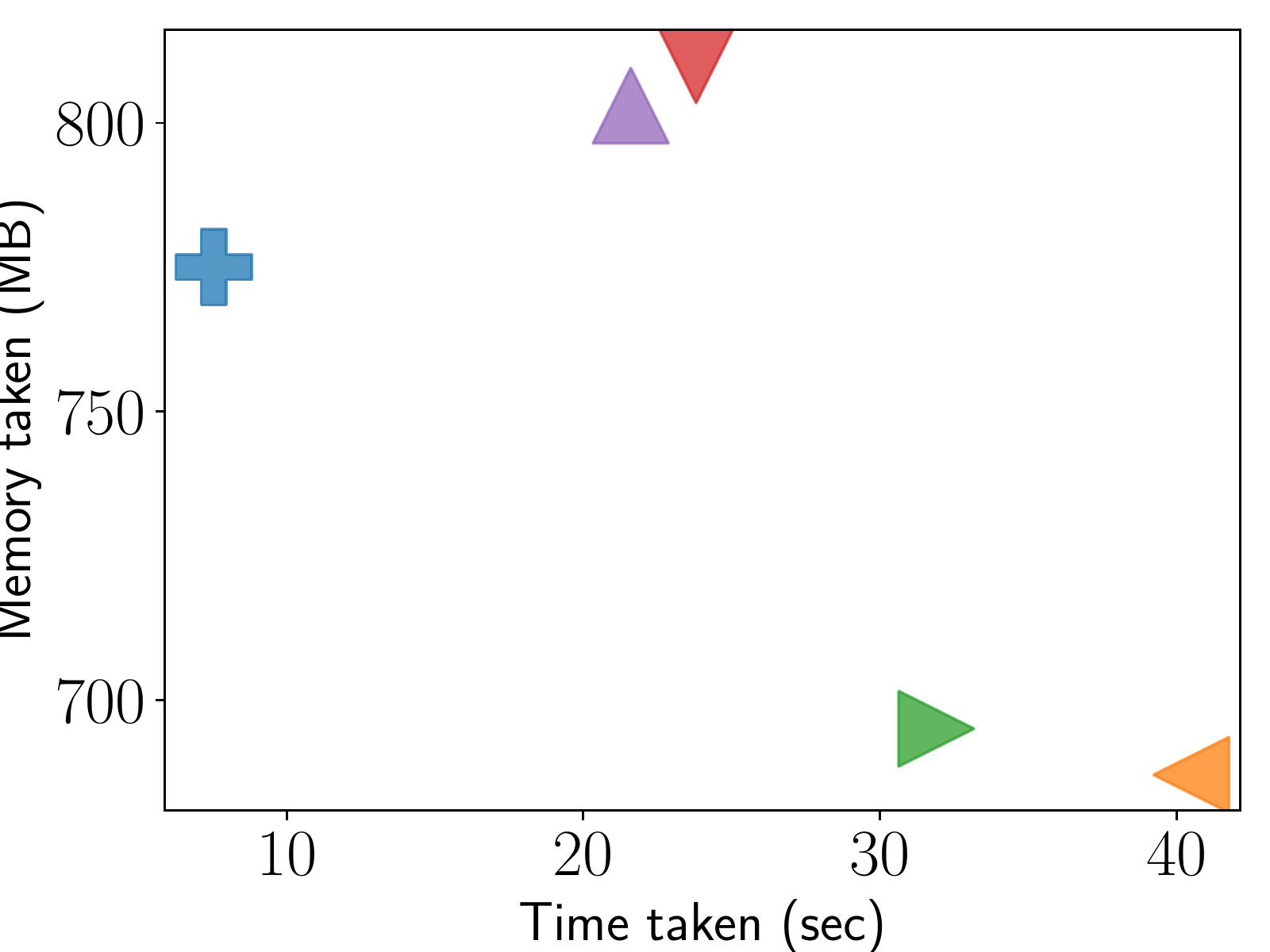}  
  \caption{fractal}
  \label{fig:fract_bench}
\end{subfigure}
\hfill
\begin{subfigure}{.3\textwidth}
  \centering
  \includegraphics[width=\textwidth]{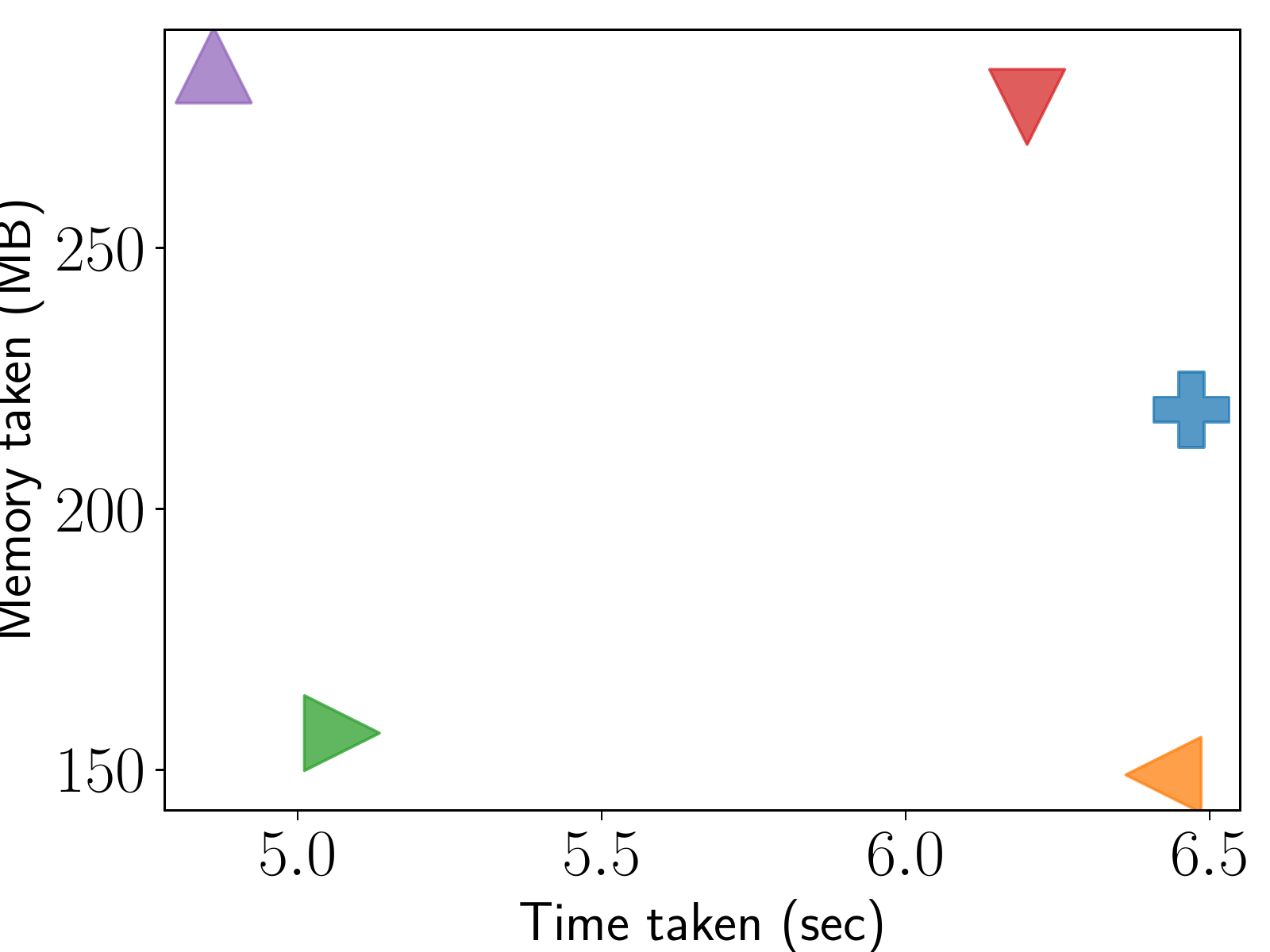}  
  \caption{o3}
  \label{fig:o3_bench}
\end{subfigure}
\begin{subfigure}{.3\textwidth}
  \centering
  \includegraphics[width=\textwidth]{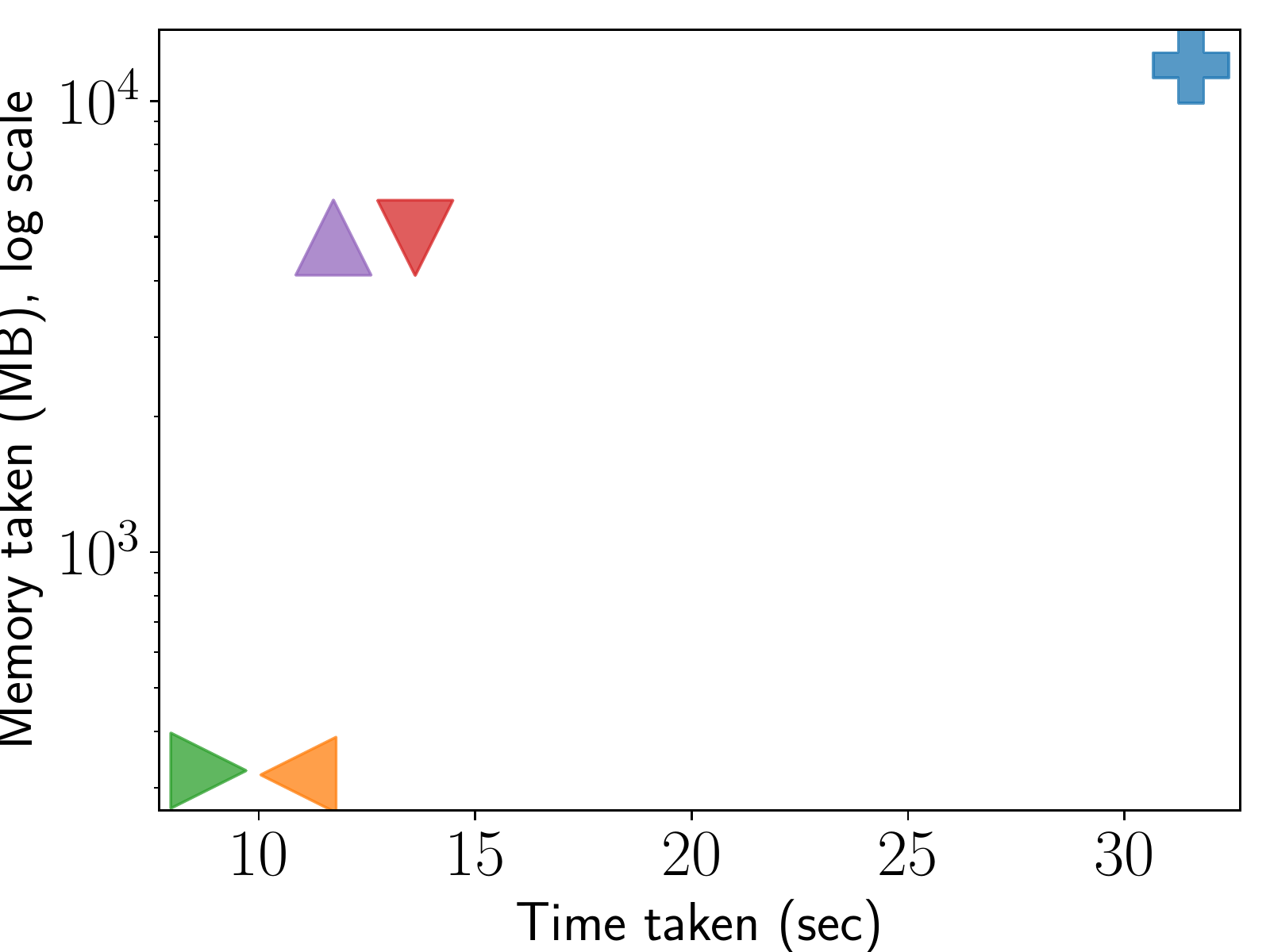}  
  \caption{torus4(1)}
  \label{fig:t41_bench}
\end{subfigure}
\begin{subfigure}{.3\textwidth}
  \centering
  \includegraphics[width=\textwidth]{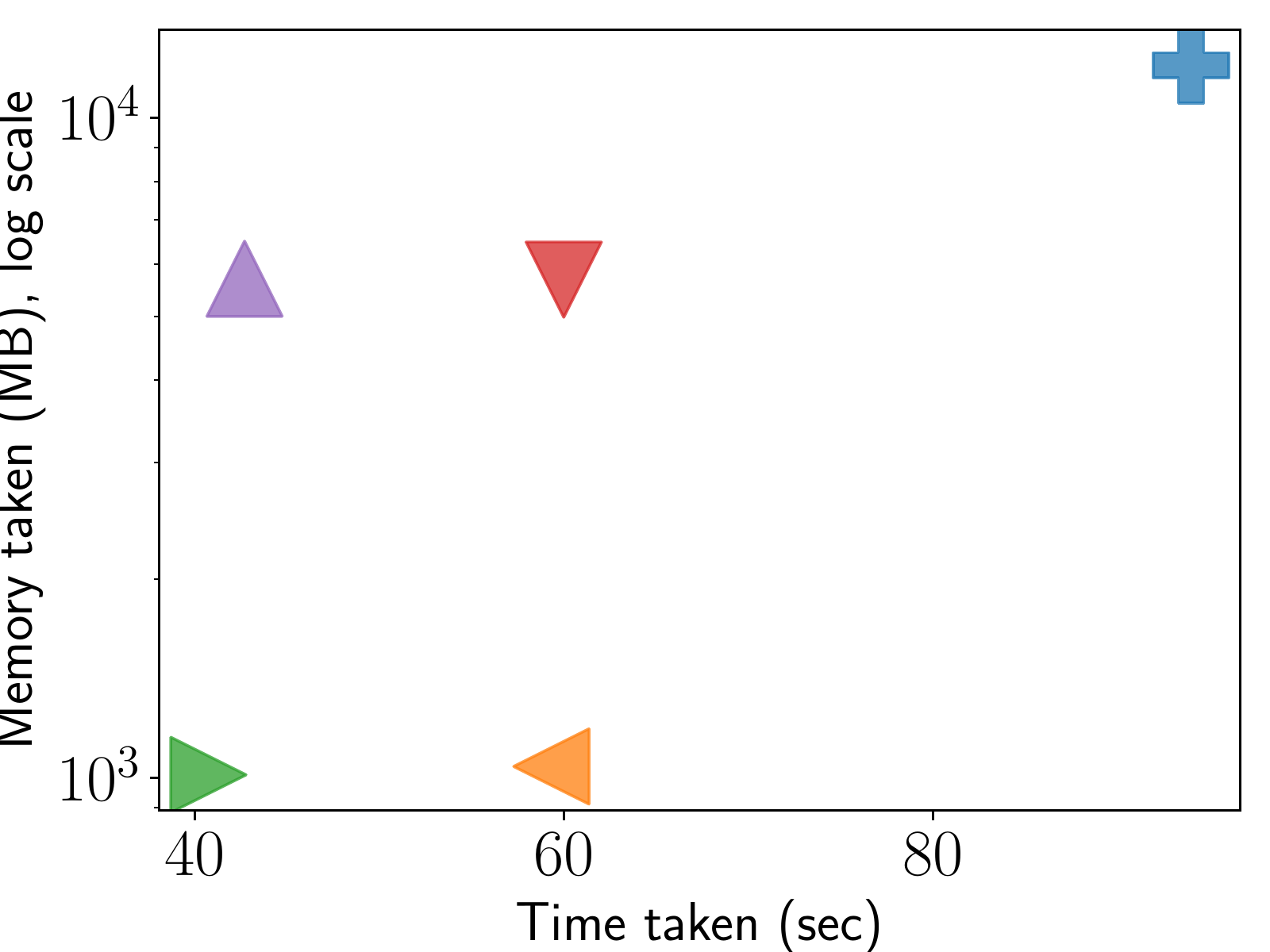}  
  \caption{torus4(2)}
  \label{fig:t42_bench}
\end{subfigure}

  \caption{ Computation time and peak memory usage by Dory, DoryNS, and Ripser for the data sets
  used for benchmarking. The peak memory usage for torus4 data set is shown in log scale. Results
  for Hi-C data sets are not shown here because only Dory was able to process it. }

\label{fig:benchmarks}
\end{figure}

\begin{figure}[tbhp]
  \centering
  \begin{subfigure}{.3\textwidth}
    \includegraphics[width=\linewidth]{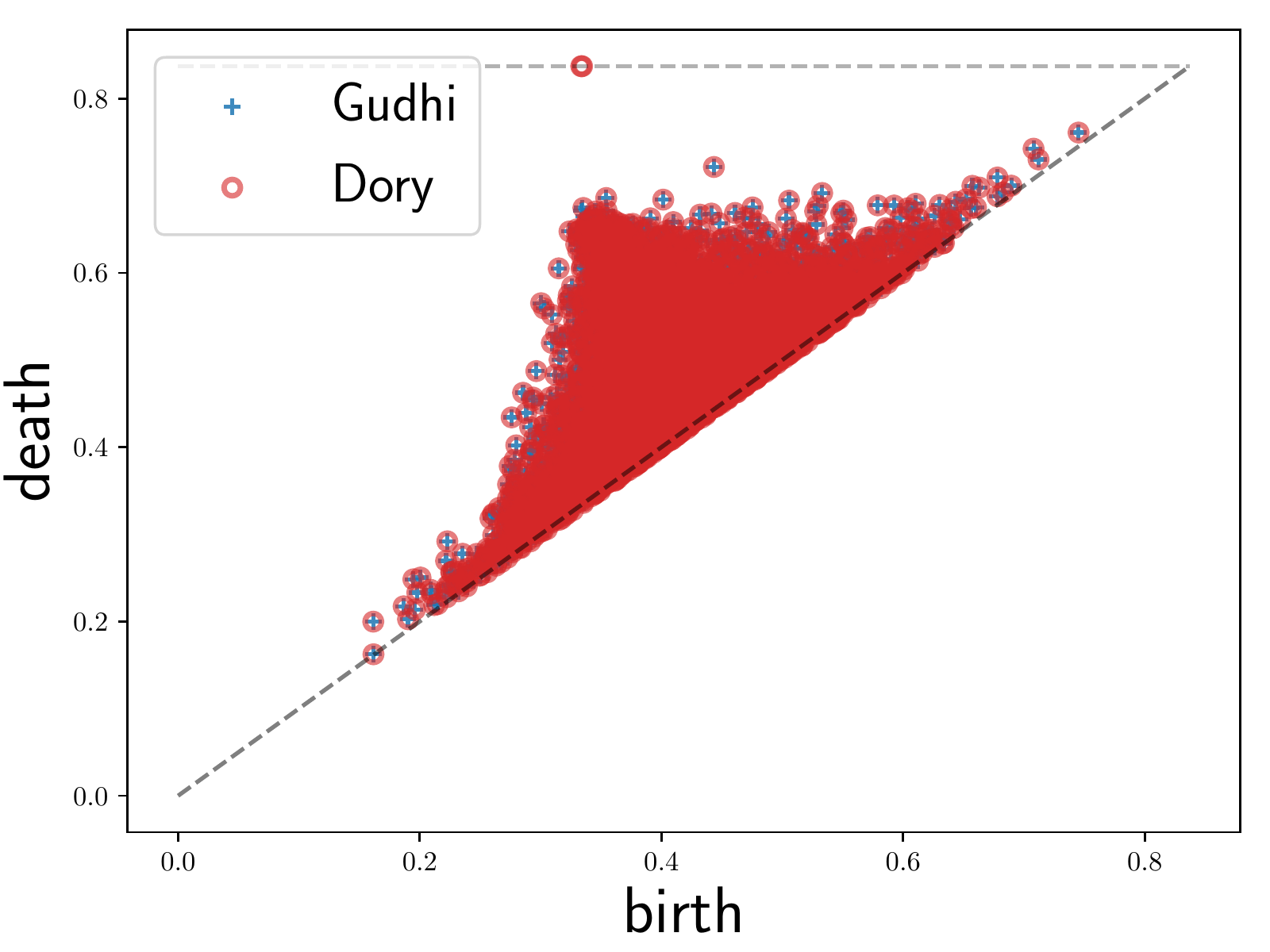}  
    \caption{Gudhi}
    \label{fig:o3_H1_gudhi_main}
  \end{subfigure}
  \centering
  \begin{subfigure}{.3\textwidth}
    \includegraphics[width=\linewidth]{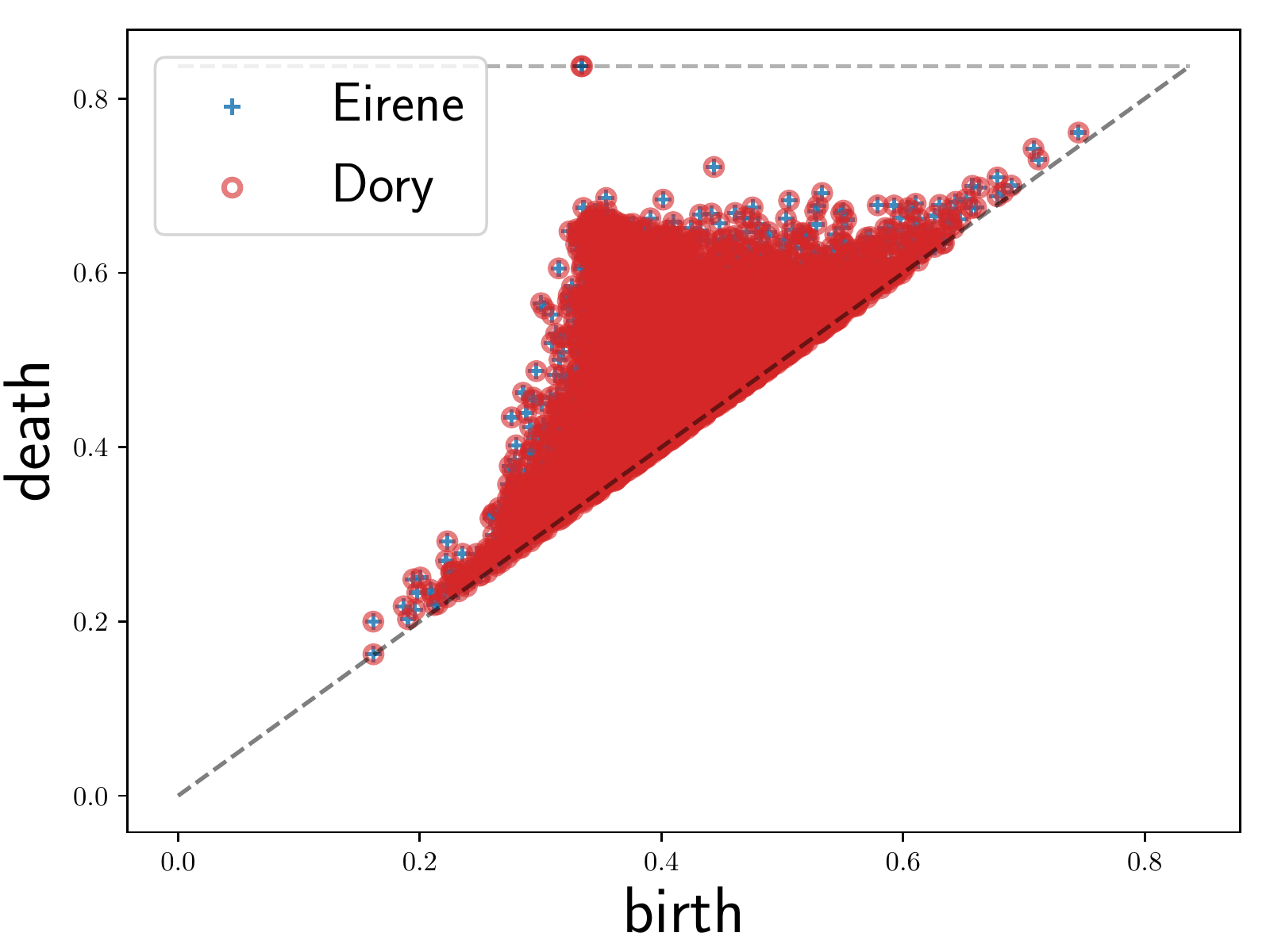}  
    \caption{Eirene}
    \label{fig:o3_H1_eirene_main}
  \end{subfigure}
  \centering
  \begin{subfigure}{.3\textwidth}
    \includegraphics[width=\linewidth]{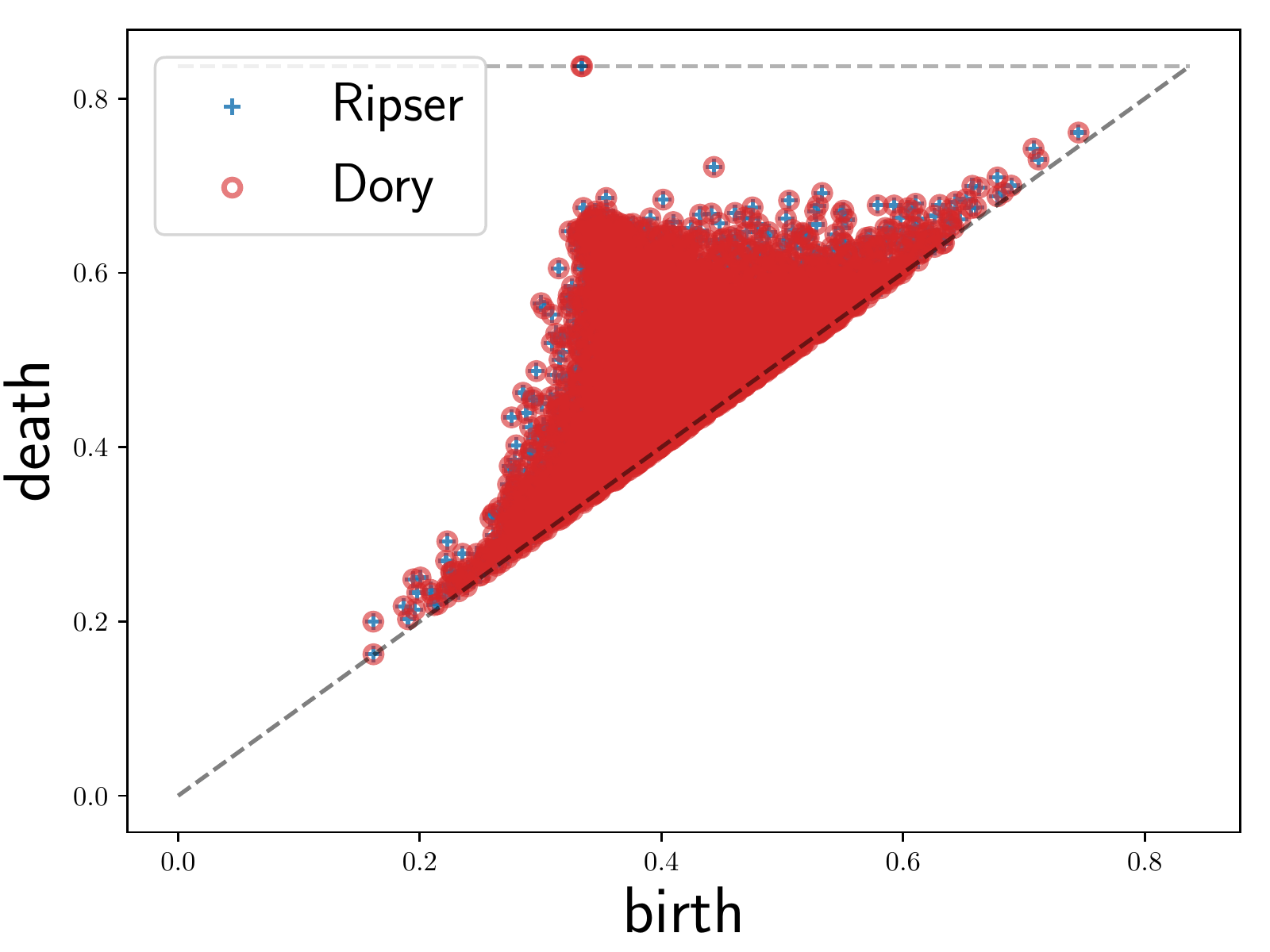}  
    \caption{Ripser}
    \label{fig:o3_H1_ripser_main}
  \end{subfigure}
    \caption{o3 H$_1$ PD: Gudhi does not report one feature that does not die and is reported by
    other algorithms.}
    \label{fig:o3_H1_PD_main}
\end{figure}

\begin{figure}[tbhp]
  \centering
  \begin{subfigure}{.3\textwidth}
    \includegraphics[width=\linewidth]{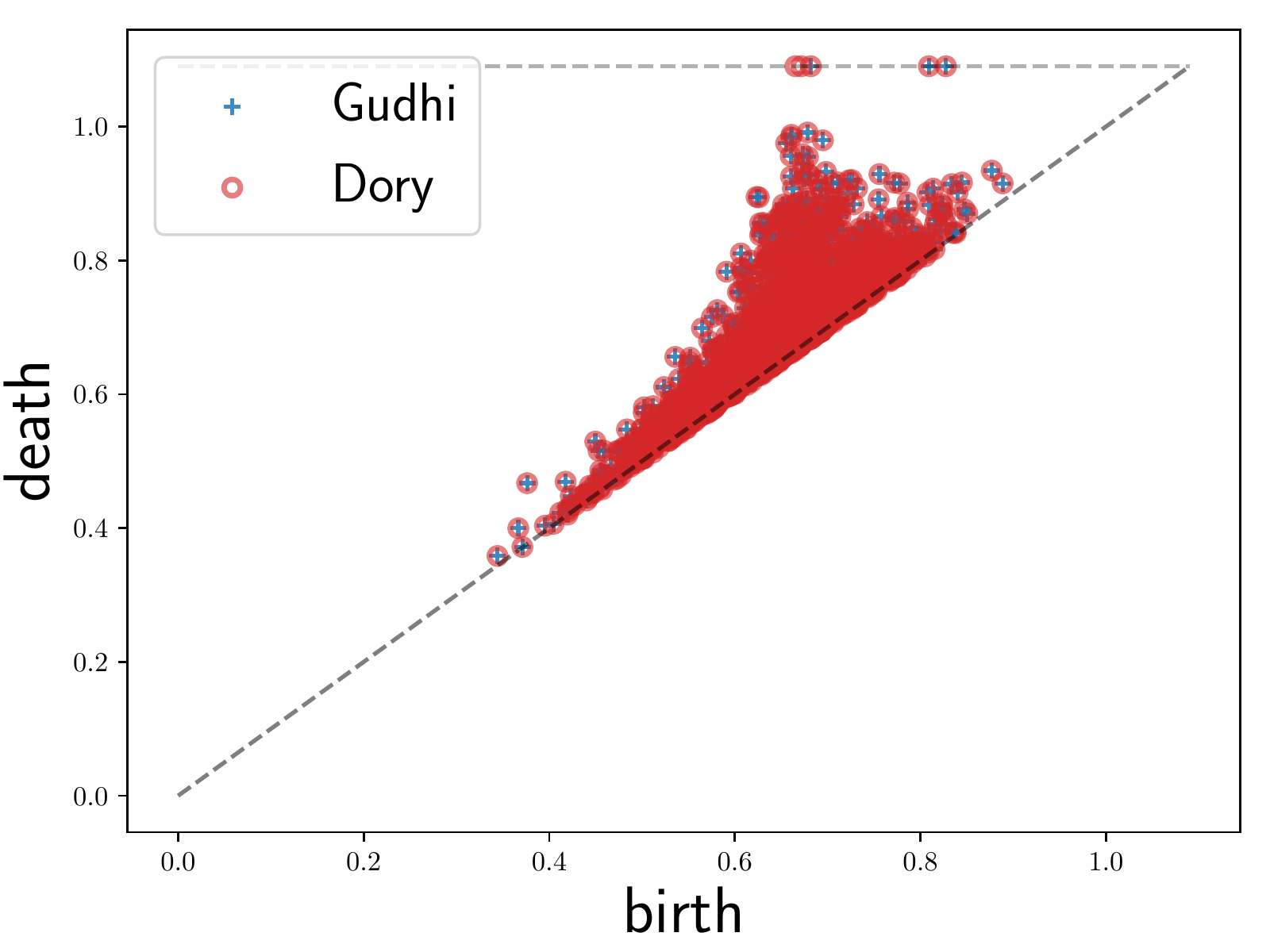}  
    \caption{Gudhi}
    \label{fig:o3_H2_gudhi_main}
  \end{subfigure}
  \centering
  \begin{subfigure}{.3\textwidth}
    \includegraphics[width=\linewidth]{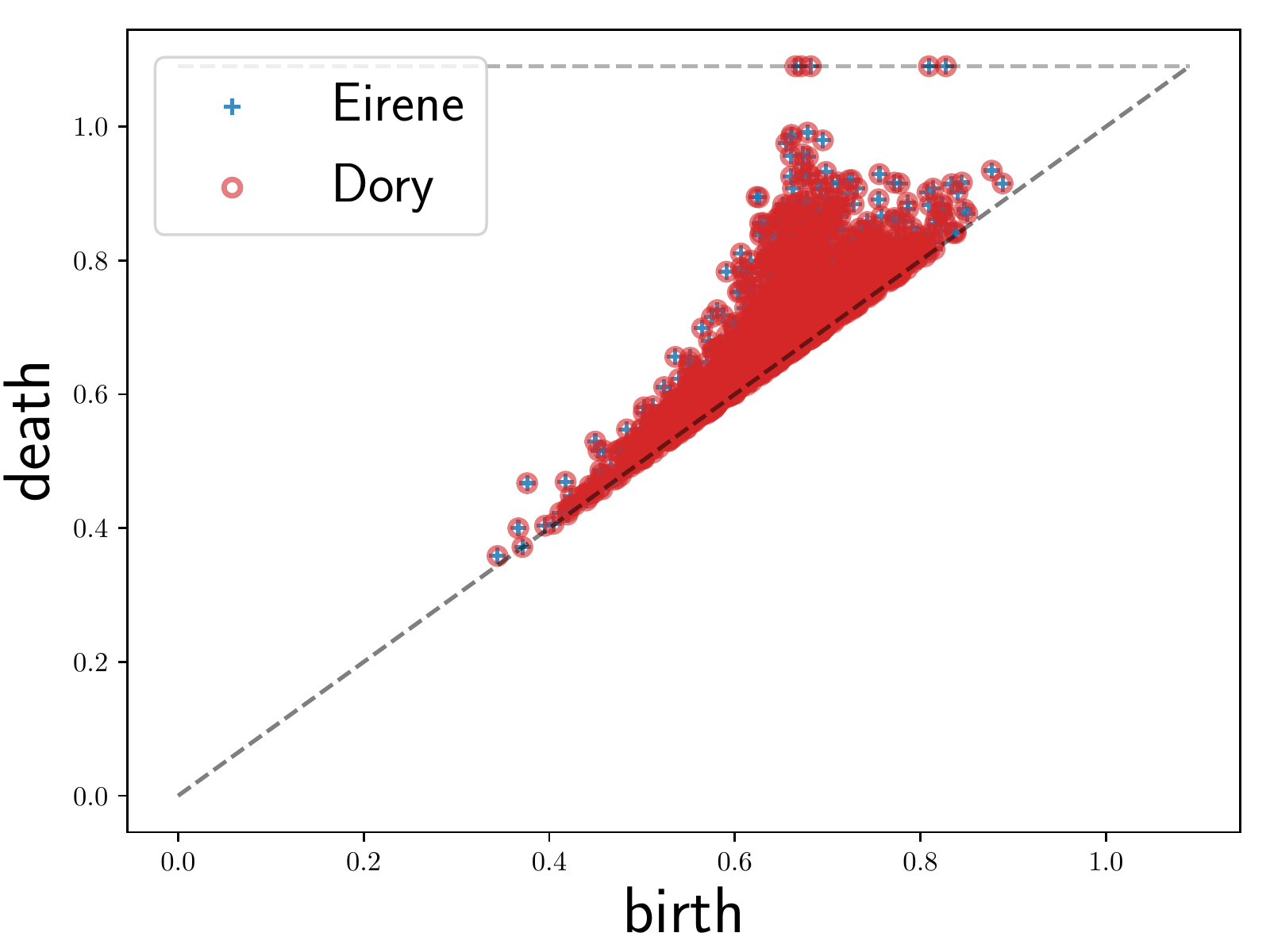}  
    \caption{Eirene}
    \label{fig:o3_H2_eirene_main}
  \end{subfigure}
  \centering
  \begin{subfigure}{.3\textwidth}
    \includegraphics[width=\linewidth]{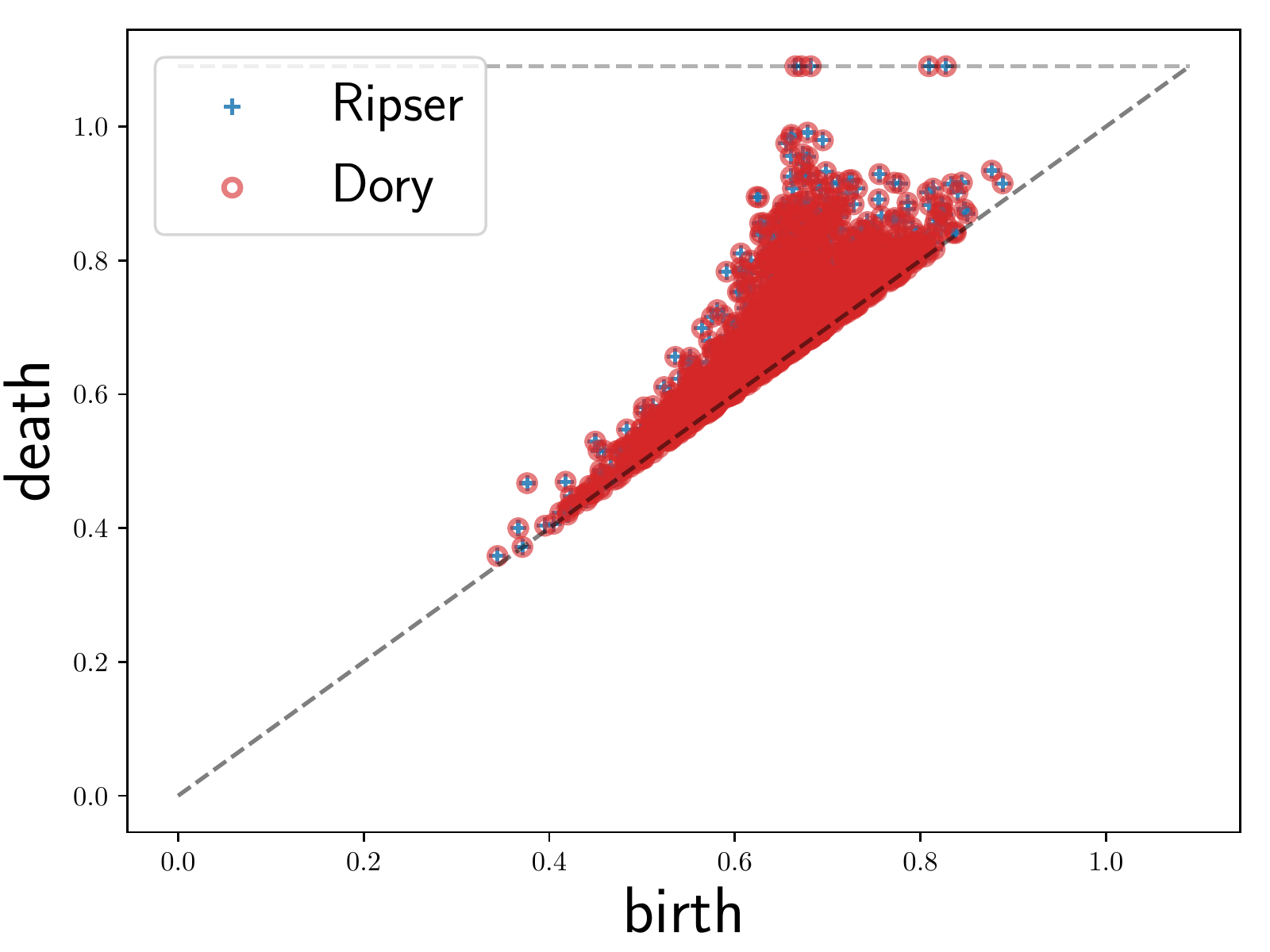}  
    \caption{Ripser}
    \label{fig:o3_H2_ripser_main}
  \end{subfigure}
    \caption{o3 H$_2$ PD: Gudhi does not report two features that do not die and are reported by
    other algorithms.}
    \label{fig:o3_H2_PD_main}
\end{figure}

All PDs are shown in appendices~\ref{app:PDs} and~\ref{app:pd_hic}. We observed a discrepancy in the
PDs of o3 data set for Gudhi (see Figures~\ref{fig:o3_H1_PD_main} and~\ref{fig:o3_H2_PD_main}),
specifically in topological features that do not die. The Gudhi PDs were produced using parameters
\texttt{points = data, max\_edge\_length = 1} in the method \texttt{RipsComplex} (Python v 3.8.5).
In the resulting file, \texttt{inf} entries were replaced by -1 before plotting. The benchmarking
and plotting codes can be found at \texttt{https://github.com/nihcompmed/Dory}. We did not explore
features of Gudhi, for example, the edge collapse option, that might improve its efficiency or give
a consistent PD.

\section{Topology of Human Genome}\label{sec:human_gene}

Among different techniques to quantify chromatin structure, Hi-C experiments allow relatively
unbiased measurements across an entire genome~\citep{lieberman2009comprehensive}.  Hi-C is based on
chromosome conformation capture (3C) which attempts to determine spatial proximity in the cell
nucleus between pairs of genetic loci. The experiments measure the interaction frequency of every
pair of loci on the genome, and this is believed to correlate with spatial distance in the cell
nucleus~\citep{lieberman2009comprehensive, dixon2012topological}.

Using Hi-C experiments,~\citet{rao20143d} identified chromatin loops in mouse lymphoblast cells
which are orthologous to loops in human lymphoblastoid cells. To highlight the functional importance
of chromatin loops they provide multiple sources of evidence that associate most of the detected
loops with gene regulation. \citet{rao2017cohesin} showed that treatment of DNA with auxin
removes most loop domains. 

We test this result by computing PH to compare the number of loops (and voids) across the Hi-C data
sets from two different experimental conditions---with and without auxin treatment, provided
in~\citet{rao2017cohesin}. To compute PH, the DNA is visualized as a point-cloud where each point
represents a contiguous segment of a 1000 base pairs on a chromosome, a so-called genomic bin. The
relative pairwise spatial distances between genomic bins are then estimated from the Hi-C data set
at this 1 kilobase resolution. The functionally significant loops in this point-cloud are most
likely the ones with spatially close genomic bins on their boundary to allow for biological
interaction via physical processes such as diffusion. Therefore, we compute PH up to a low $\tau_m$
resulting in a sparse filtration.

\begin{figure}[tbhp]
  \centering
  \includegraphics[width=0.65\linewidth]{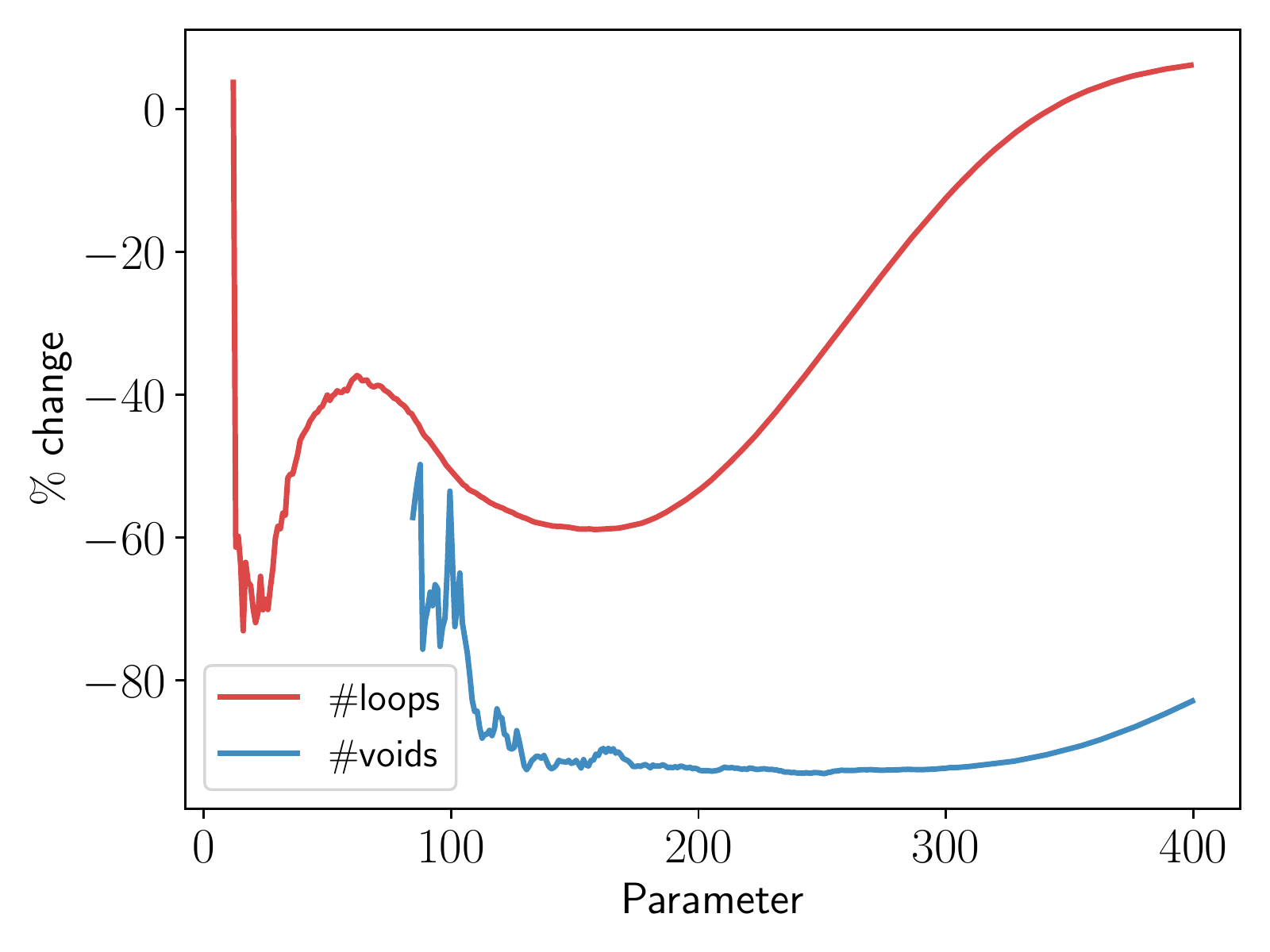}  
  \caption{PH of the entire human genome at 1 kilobase resolution shows that the number of loops and
  voids decreases significantly upon addition of auxin.}
\label{fig:HiC_num_diff}
\end{figure}

In Figure~\ref{fig:HiC_num_diff}, we plot the percentage change in the number of loops and voids
upon addition of auxin ($(\beta_i^{\text{auxin}} -
\beta_i^{\text{control}})/\beta_i^{\text{control}}*100$). It shows that there is a significant
decrease in the number of loops, corroborating previous results~\citep{rao2017cohesin}. The analysis
using PH additionally shows that the percentage of reduction in the number of loops is greater for
thresholds less than 50 and between 100 and 200, and it also shows that most voids are not born when
auxin is added.  These results warrant an investigation into possible biological implications. The
PDs are shown in appendix~\ref{app:pd_hic}.

\section{Discussion}\label{sec:discussion}

In this paper we introduced a new algorithm that overcomes computational limitations that have
prevented the application of PH to large data sets. Compared with pre-existing algorithms, Dory
provides significant improvements in memory requirements, without an impractical increase in the
computation time. Dory is able to process the large Hi-C data set for the human genome at high
resolution, corroborating the expected topology changes of  chromatin in different experimental
conditions. 

While our algorithm is limited to computing PH for VR-filtrations up to three dimensions, a large
class of real-world scientific data sets requires only such dimensional restrictions. The current
implementation of the algorithm computes PH modulo 2, but it can be easily extended to any prime
field.

An alternate class of methods deals with large data sets by approximating their PDs. For example,
SimBa~\citep{dey2019simba} reduces the number of simplices in the filtration by approximating it to
a sparse filtration such that the PDs of the sparse filtration are within a theoretical error of
margin when compared to those of the original filtration. Another method, PI-Net~\citep{som2020pi},
uses neural networks to predict persistence images, that are pixelated approximations of PDs in
$\mathbb{R}^2.$ The former method uses Gudhi to compute PDs of the approximated sparse filtrations
and the latter uses Ripser to compute true PDs when training the neural network. Since Dory can
handle larger data sets compared to both Gudhi and Ripser, SimBa and PI-Net can expand their scope
by using Dory instead.

In this paper we have focused on computing PDs.  However, our algorithm can also be extended to
compute representative boundaries of the holes and voids in the data set. For scientific
applications, these representative boundaries of topological features in the data set are critical
for connecting topology to structural properties of the data that may be linked to functional
properties of the underlying system. For instance, they might yield insights into the biological
implications of a reduction in the number of voids in human genome upon treatment with auxin.
Computation of representative boundaries faces the hurdle of high memory cost. We are currently
working on developing a scalable algorithm.

\section*{Acknowledgement}
This research was supported by the Intramural Research Program of the NIH, the
National Institute of Diabetes and Digestive and Kidney Diseases (NIDDK).

\newpage

\appendix

\begin{appendices}

\section{Standard Row and Column Algorithms}\label{app:row_col}

\begin{algorithm}
  \begin{algorithmic}[1]

    \For{$j:2 \textrm{ to } N$}

      \State $i \gets 1$

      \While{$i < j$}

        \If{$\text{low}(j) = \text{pivot}(i)$}

            \State  column $j \gets \text{column } j \oplus \text{column } i$ 

            \If{column $j$ is empty}

              \State $\text{low}(j) \gets -1$\Comment{this column is reduced to $\mathbf{0}$}

              \State break

            \Else
              \State  $i \gets 1 $
            \EndIf

        \Else

            \State  $i \gets i+1$

        \EndIf

      \EndWhile

    \State $\text{pivot}(j) \gets \text{low}(j)$

    \EndFor

  \end{algorithmic}
  \caption{Standard column algorithm}
  \label{alg:std_col}
\end{algorithm}

\begin{algorithm}
  \begin{algorithmic}[1]

    \For{$i:N \textrm{ to } 1$}

      \For{$j: 1 \textrm{ to } N$}
          \If{$\text{low}(j) = i$}
            \text{break}
          \EndIf
      \EndFor

      \If{$j = N + 1$}
        \text{continue}
      \EndIf

      \State $\text{pivot}(j) = \text{low}(j)$

      \For{$k = j+1 \textrm{ to } N$}
        \If{$\text{low}(k) = i$}
            \State $\text{column } k \gets \text{column } k \oplus \text{column } j$
            \If{column $k$ is empty}

              \State $\text{low}(k) \gets -1$\Comment{this column is reduced to $\mathbf{0}$}
              \State $\text{pivot}(k) \gets -1$
            \EndIf

        \EndIf
      \EndFor
    \EndFor

  \end{algorithmic}
  \caption{Standard row algorithm}
  \label{alg:std_row}
\end{algorithm}

\newpage
\section{Algorithms to Compute Cohomology for Edges}\label{app:cob_edge}

\begin{algorithm}
  \begin{algorithmic}[1]
    \State \textbf{Input:} $(e, i_a, i_b, \delta_e)$
    \State \textbf{Output:} $(e, i_a, i_b, \delta_e)$

    \State $\{a, b\} \gets f_1^{-1}(e)$

    \While{$i_a < N(a)$ AND $i_b < N(b)$}
        \If{$f_0(n^a_{i_a}) < f_0(n^b_{i_b})$}
        \State $i_a \gets i_a + 1$
        \ElsIf{$f_0(n^a_{i_a}) > f_0(n^b_{i_b})$}
        \State $i_b \gets i_b + 1$
        \Else
        \State \Return $(e, i_a, i_b, \langle f_1(\{a, b\}), n^a_{i_a}\rangle)$
        \EndIf

    \EndWhile

    \State \Return $(e, i_a, i_b, \text{Empty})$ \Comment{Move to case 2}

  \end{algorithmic}
  \caption{Case 1 for edges}
  \label{alg_case1_edge}
\end{algorithm}

\begin{algorithm}
  \begin{algorithmic}[1]
    \State \textbf{Input:} $(e, i_a, i_b, \delta_e)$
    \State \textbf{Output:} $(e, i_a, i_b, \delta_e)$

    \State $\{a, b\} \gets f_1^{-1}(e)$

    \While{$i_a < N(a)$ AND $i_b < N(b)$}
        \If{$f_1(e^a_{i_a}) < f_1(e^b_{i_b})$}
          \State $\{a, d\} \gets f_1^{-1}(e^a_{i_a})$
          \If{$d \in N^b$ AND $f_1(\{b, d\}) < f_1(e^a_{i_a})$}
              \State \Return $(e, i_a, i_b, \langle f_1(e^a_{i_a}), d \rangle)$
          \Else
              \State $i_a \gets i_a + 1$
          \EndIf
        \Else
          \State $\{b, d\} \gets f_1^{-1}(e^b_{i_b})$
          \If{$d \in N^a$ AND $f_1(\{a, d\}) < f_1(e^b_{i_b})$}
              \State \Return $(e, i_a, i_b, \langle f_1(e^b_{i_b}), d \rangle)$
          \Else
              \State $i_b \gets i_b + 1$
          \EndIf
        \EndIf
    \EndWhile

    \While{$i_a < N(a)$}
          \State $\{a, d\} \gets f_1^{-1}(e^a_{i_a})$
          \If{$d \in N^b$ AND $f_1(\{b, d\}) < f_1(e^a_{i_a})$}
              \State \Return $(e, i_a, i_b, \langle f_1(e^a_{i_a}), d \rangle)$
          \Else
              \State $i_a \gets i_a + 1$
          \EndIf
    \EndWhile

    \While{$i_b < N(b)$}
          \State $\{b, d\} \gets f_1^{-1}(e^b_{i_b})$
          \If{$d \in N^a$ AND $f_1(\{a, d\}) < f_1(e^b_{i_b})$}
              \State \Return $(e, i_a, i_b, \langle f_1(e^b_{i_b}), d \rangle)$
          \Else
              \State $i_b \gets i_b + 1$
          \EndIf
    \EndWhile

    \State \Return $(e, i_a, i_b, \text{Empty})$

  \end{algorithmic}
  \caption{Case 2 for edges}
  \label{alg_case2_edge}
\end{algorithm}

\begin{algorithm}
  \begin{algorithmic}[1]
    \State \textbf{Input:} $e$
    \State \textbf{Output:} $(e, i_a, i_b, \delta_e)$

    \State $\{a, b\} \gets f_1^{-1}(e)$
    \State $i_a \gets 0$
    \State $i_b \gets 0$
    \State $\delta^e_0 \gets $ Empty

    \State  $(e, i_a, i_b, \delta_e) \gets$ \textbf{Function} \texttt{Case1}$((e, i_a, i_b, \delta_e))$

    \If{$low$ is Empty}
      \State $i_a \gets $ smallest index s.t. $f_1(e^a_{i_a}) > f_1(e)$
      \State $i_b \gets $ smallest index s.t. $f_1(e^b_{i_b}) > f_1(e)$
      \State $(e, i_a, i_b, \delta_e) \gets$ \textbf{Function} \texttt{Case2}$((e, i_a, i_b, \delta_e))$
    \EndIf

    \State \Return $(e, i_a, i_b, \delta_e)$

  \end{algorithmic}
  \caption{\texttt{FindSmallestt}}
  \label{alg_findsmallest_edge}
\end{algorithm}

\begin{algorithm}
  \begin{algorithmic}[1]
    \State \textbf{Input:} $(e, i_a, i_b, \delta_e)$
    \State \textbf{Output:} $(e, i_a, i_b, \delta_e)$

    \State $\{a, b\} \gets f_1^{-1}(e)$

    \State $\langle k^p, k^s \rangle \gets \delta^e_i$
    \If{$k^p = f_1(e)$}
        
      \State Increment $i_a, i_b$ by 1
      \State  $(e, i_a, i_b, \delta^e_{i+1}) \gets$ \textbf{Function} \texttt{Case1}$((e, i_a, i_b, \delta^e_i))$

      \If{$\delta^e_{i+1}$ is Empty}
        \State $i_a \gets $ smallest index s.t. $f_1(e^a_{i_a}) > f_1(e)$
        \State $i_b \gets $ smallest index s.t. $f_1(e^b_{i_b}) > f_1(e)$
      \Else
        \State \Return $(e, i_a, i_b, \delta_e)$
      \EndIf
    \Else
      \State $f_1(e^a_{i_a}) < f_1(e^a_{i_b})\, ? \, \text{increment } i_a : \text{increment } i_b$
    \EndIf

    \State $(e, i_a, i_b, \delta_e) \gets$ \textbf{Function} \texttt{Case2}$((e, i_a, i_b, \delta_e))$

    \State \Return $(e, i_a, i_b, \delta_e)$

  \end{algorithmic}
  \caption{FindNextt}
  \label{alg_findnext_edge}
\end{algorithm}

\begin{algorithm}
  \begin{algorithmic}[1]
    \State \textbf{Input:} $e, \delta_{\#}$
    \State \textbf{Output:} $(e, i_a, i_b, \delta_e)$

    \State $\{a, b\} \gets f_1^{-1}(e)$
    \State $\langle k^p, k^s \rangle \gets \delta_{\#}$

    \If{$k^p < f_1(e)$}
        
      \State  $(e, i_a, i_b, \delta_e) \gets$ \textbf{Function} \texttt{FindSmallestt}$(e)$
      \State \Return $(e, i_a, i_b, \delta_e)$

    \ElsIf{$k^p = f_1(e)$}

        \State $i_a \gets $ smallest index s.t. $f_0(n^a_{i_a}) \geq k^s$
        \State $i_b \gets $ smallest index s.t. $f_0(n^b_{i_b}) \geq k^s$

        \If{$f_0(n^a_{i_a}) = f_0(n^b_{i_b})$}
            \State  \Return $(e, i_a, i_b, \langle k^p, f_0(n^a_{i_a})\rangle))$
        \Else
            \State  $(e, i_a, i_b, \delta_e) \gets$ \textbf{Function} \texttt{FindNextt}$((e, i_a, i_b,
            \text{Empty}))$
            \If{$\delta_e$ is not Empty}
                 \State \Return $(e, i_a, i_b, \delta_e)$
            \EndIf
        \EndIf

    \EndIf

    \State $i_a \gets $ smallest index s.t. $f_1(e^a_{i_a}) \geq k^p$
    \State $i_b \gets $ smallest index s.t. $f_1(e^b_{i_b}) \geq k^p$

    \If{$\langle k^p, k^s \rangle$ is in $\delta e$}
      \State \Return $(e, i_a, i_b, \langle k^p, k^s \rangle)$
    \EndIf

    \State  $(e, i_a, i_b, \delta_e) \gets$ \textbf{Function} \texttt{Case2}$((e, i_a, i_b, \delta_e))$

    \State \Return $(e, i_a, i_b, \delta_e)$

  \end{algorithmic}
  \caption{FindGEQt}
  \label{alg_findgeq_edge}
\end{algorithm}

\newpage
\section{Algorithms to Compute Coboundary of Triangles}\label{app:cob_triangle}

\begin{algorithm}
  \begin{algorithmic}[1]

    \State \textbf{Input:} $(t, i_a, i_b, i_c, f, \delta_t)$
    \State \textbf{Output:} 1 or 0

    \State $\langle k^p, k^s \rangle \gets f_2(t)$
    \State $a \gets $ min$\{f_1^{-1}(k^p)\}$
    \State $b \gets $ max$\{f_1^{-1}(k^p)\}$
    \State $c \gets k^s$

    \While{$i_c < N(a)$ AND $f_1(e^c_{i_c}) < k^p$}

        \State $\{c, d\} \gets e^c_{i_c}$

        \If{$f_1(\{a, d\})$ and $f_1(\{b, d\})$ are less than $k^p$}

            \State $f \gets 0$
            \State $\delta_t \gets \langle k^p, f_1(e^c_{i_c})\rangle$
            \State \textbf{return} 1
            
        \EndIf

        \State $i_c \gets i_c + 1$

    \EndWhile

    \State \textbf{return} 0

  \end{algorithmic}
  \caption{Case1 for triangles}
  \label{alg:case1_triangles}
\end{algorithm}

\begin{algorithm}
  \begin{algorithmic}[1]

    \State \textbf{Input:} $(t, i_a, i_b, i_c, f, \delta_t)$

    \State $\langle k^p, k^s \rangle \gets f_2(t)$
    \State $a \gets $ min$\{f_1^{-1}(k^p)\}$
    \State $b \gets $ max$\{f_1^{-1}(k^p)\}$
    \State $c \gets k^s$

    \While{not reached end of all $E^a, E^b, E^c$}

        \State $o_* \gets \text{min}\{f_1(e^a_{i_a}), f_1(e^b_{i_b}),
        f_1(e^c_{i_c})\}$\Comment{Exclude indices that reach end of edge-nbd}

        \State $\{v_1, d\} \gets f_1^{-1}(o_*)$

        \State $\{v_2, v_3\} \gets \{a, b, c\} \setminus \{v_1\}$

        \If{$f_1(\{v_2, d\})$ and $f_1(\{v_3, d\})$ are less than $o_*$}
            \State $\delta_t \gets \langle o_*, f_1(\{v_2, v3_\})\rangle$
            \State $f \gets $ 1 if $v_1 = a$; 2 if $v_1 = b$; 3 if $v_1 = c$
            \State \textbf{return}
        \Else
            \State Increment $i_{v_1}$
        \EndIf

    \EndWhile

    \State $\delta_t \gets$ Empty
    \State $f \gets -1$

  \end{algorithmic}
  \caption{Case 2 for triangles}
  \label{alg:case2_triangles}
\end{algorithm}

\begin{algorithm}
  \begin{algorithmic}[1]

    \State \textbf{Input:} $(t, i_a, i_b, i_c, f, \delta_t)$

    \State $i_c \gets 0$

    \If{\textbf{Function} \texttt{Case1}$(t, i_a, i_b, i_c, f, \delta_t)$}
        \State \textbf{return}
    \EndIf

    \State $\langle k^p, k^s \rangle \gets t$
    \State $\{a, b\} \gets f_1^{-1}(k^p)$

    \State $i_a \gets \argmin\limits_{i_a}(f_1(e^a_{i_a}) \geq k^p)$ 
    \State $i_b \gets \argmin\limits_{i_b}(f_1(e^b_{i_b}) \geq k^p)$

    \State Increment $i_a, i_b$

    \State \textbf{Function} \texttt{Case2}$(t, i_a, i_b, i_c, f, \delta_t)$

  \end{algorithmic}
  \caption{FindSmallesth}
  \label{alg:findsmallest_triangle}
\end{algorithm}

\begin{algorithm}
  \begin{algorithmic}[1]

    \State \textbf{Input:} $(t, i_a, i_b, i_c, f, \delta_t)$
    \State $\langle k^p, c \rangle \gets f_2(t)$

    \State $a \gets $ min$\{f_1^{-1}(k^p)\}$
    \State $b \gets $ max$\{f_1^{-1}(k^p)\}$
    \State $c \gets k^s$

    \If{$f = 0$}
        
        \State Increment $i_c$

        \If{\textbf{Function} \texttt{Case1} $(t, i_a, i_b, i_c, f, \delta_t)$}
            \State \textbf{return}
        \EndIf

        \State $i_a \gets \argmin\limits_{i_a}(f_1(e^a_{i_a}) \geq k^p)$ 
        \State $i_b \gets \argmin\limits_{i_b}(f_1(e^b_{i_b}) \geq k^p)$ 

        \State Increment $i_a, i_b$

    \EndIf

    \If{Not moving from \texttt{Case1}}

        \If{$f = 1$}
        \State $i_a \gets i_a + 1$
        \ElsIf{$f = 2$}
        \State $i_b \gets i_b + 1$
        \ElsIf{$f = 3$}
        \State $i_c \gets i_c + 1$
        \EndIf
        
    \EndIf

    \State \textbf{Function} \texttt{Case2}$(t, i_a, i_b, i_c, f, \delta_t)$

  \end{algorithmic}
  \caption{FindNexth for triangles}
  \label{alg:findnext_triangle}
\end{algorithm}

\begin{algorithm}
  \begin{algorithmic}[1]

    \State \textbf{Input:} $(t, i_a, i_b, i_c, f, \delta_t), \delta_{\#}$

    \State $\langle k^p_{\#}, k^s_{\#} \rangle \gets \delta_{\#}$

    \State $\langle k^p, c \rangle \gets t$

    \State $\{a, b\} \gets f_1^{-1}(k^p)$

    \If{$k^p_{\#} < k^p$}

        \State \textbf{Function} \texttt{FindSmallesth}$(t, i_a, i_b, i_c, f, \delta_t)$
        \State \textbf{return}
    \ElsIf{$k^p_{\#} = k^p$}
        
        \State $i_c \gets \argmin\limits_{i_c}(f_1(e^c_{i_c}) \geq k^s_{\#})$ 

        \If{$f_1(e^c_{i_c}) = k^s_{\#}$}
            \State $\delta_t \gets \delta_{\#}$
            \State $v \gets 0$
            \State \textbf{return}
        \EndIf

        \If{\textbf{Function} \texttt{Case1}$(t, i_a, i_b, i_c, f, \delta_t)$}
            \State \textbf{return}
        \EndIf

        \State $i_a \gets \argmin\limits_{i_a}(f_1(e^a_{i_a}) \geq k^p)$ 
        \State $i_b \gets \argmin\limits_{i_b}(f_1(e^b_{i_b}) \geq k^p)$ 

        \State Increment $i_a, i_b$

    \Else

        \State $i_a \gets \argmin\limits_{i_a}(f_1(e^a_{i_a}) \geq k^p_o)$ 
        \State $i_b \gets \argmin\limits_{i_b}(f_1(e^b_{i_b}) \geq k^p_o)$ 
        \State $i_c \gets \argmin\limits_{i_c}(f_1(e^c_{i_c}) \geq k^p_o)$

    \EndIf

    \While{1}

        \State \textbf{Function} \texttt{Case2}$(t, i_a, i_b, i_c, f, \delta_t)$

        \If{$\delta_t \geq \delta_{\#}$ OR $\delta_t$ is Empty}
            \State \textbf{break}
        \EndIf

        \If{$f = 1$}
        \State $i_a \gets i_a + 1$
        \ElsIf{$f = 2$}
        \State $i_b \gets i_b + 1$
        \ElsIf{$f = 3$}
        \State $i_c \gets i_c + 1$
        \EndIf

    \EndWhile

  \end{algorithmic}
  \caption{FindGEQh for triangles}
  \label{alg:findgeq_triangle}
\end{algorithm}

\newpage
\section{Algorithms for Serial-parallel Cohomology Reduction of Triangles}\label{app:serial_parallel}

%
%
%
%
%
%
%
%
%
%
%
%
%

\begin{algorithm}
  \begin{algorithmic}[1]

    \State \textbf{Function} \texttt{SerialParallelReduce}
    \Indent

        \State  \textbf{Function} \texttt{ParallelReduce}
        \State  \textbf{Function} \texttt{SerialReduce}
        \State \textbf{Function} \texttt{Clearance}
    \EndIndent
    \State \textbf{EndFunction}

    \item[]

    \State \textbf{Input} $F_1,$ batch-size
    \State $V^\bot, \mathbf{v} \gets [\,]$
    \State $i \gets 0$

    \For{$e$ in $F^{-1}_1$}

        \State Compute $\delta e$ and sort it along $F^{-1}_2$

        \For{$t$ in $\delta e$ with $d(t) = e$}

            \State $(t, i_a, i_b, i_c, f, \delta_t) \gets $ \textbf{Function} \texttt{FindSmallesth}$(t)$
            \If{$\delta^t_1$ is Empty}
              \State \textbf{continue}
            \EndIf
            
            \State $v_i \gets [(t, i_a, i_b, i_c, f, \delta_t)]$
            \State $\delta^{v_i}_* \gets \delta_t$
            \State $f_e \gets 0$
            \State Append $(t, v_i, \delta^{v_i}_* , f_v, f_r, f_a, f_e)$ to $\mathbf{v}$
            \State $i \gets i + 1$

            \If{$i = $ batch-size}
                \State \textbf{Function} \texttt{SerialParallelReduce} 
            \EndIf

        \EndFor
        
    \EndFor

    \item[]

      \While{$i > 0$}
        \State \textbf{Function} \texttt{SerialParallelReduce} 

    \EndWhile

  \end{algorithmic}
  \caption{SerialParallel: Reduction of triangles}
  \label{alg:cohom_reduce_serial_parallel}
\end{algorithm}

\begin{algorithm}
  \begin{algorithmic}[1]

    \State \textbf{Input:} $\mathbf{v}, V^\bot, p^\bot$
    \State \textbf{Output:} $\mathbf{v}$

    \For{$(t, v_i, \delta^{v_i}_*, f_v, f_r, f_a, f_e)$ in $\mathbf{v}$}\Comment{Embarrassingly
    parallel}

          \State $f_r \gets 0$
          \State $f_a \gets 1$

          \If{$(\delta^{v_i}_*, t)$ is a trivial persistence pair}
              \State \textbf{continue}
          \EndIf

          \While{$(\delta^{v_i}_*, t')$ is a trivial persistence pair OR is in $p^\bot$}

            \State $(t', i_a, i_b, i_c, v, \delta_{t'}) \gets $ \textbf{Function}
            \texttt{FindGEQt}$(t', \delta^{v_i}_*)$

            \State Append $(t', i_a, i_b, i_c, f, \delta_{t'})$ to $v_i$

            \For{$t_k$ in $V^\bot(t')$}

                \State $(t_k, i_a, i_b, i_c, f, \delta_{t_k}) \gets $ \textbf{Function}
                \texttt{FindGEQt}$(t_k, \delta^{v_i}_*)$

                \State Append $[(t_k, i_a, i_b, i_c, f, \delta_{t_k})]$ to $v_i$

            \EndFor

            \State Implicit column algorithm to update $\delta_*^{v_i}$

            \If{$\delta^{v_i}_*$ is Empty}
                \State $f_e \gets 1$
                \State \textbf{return}
            \EndIf
            \EndWhile

  \EndFor

  \end{algorithmic}
  \caption{Parallel reduction}
  \label{alg:parallelreduce}
\end{algorithm}

\begin{algorithm}
  \begin{algorithmic}[1]

    \For{$(t, v_i, \delta^{v_i}_*, f^i_r, f^i_v, f^i_a, f^i_e)$ in $\mathbf{v}$}

      \If{$f^i_e$ OR $f^i_v$}
        \State \textbf{continue}
      \EndIf

      \State $j \gets 1$

      \While{$j < i$}

          \If{$f^j_e$}
            \State $j \gets j + 1$
            \State \textbf{continue}
          \EndIf

          \If{$\delta^{v_j}_* > \delta^{v_i}_*$}

               \State $j \gets j + 1$
               \State \textbf{continue}
            
          \EndIf
              
          \If{$\delta^{v_j}_* < \delta^{v_i}_*$}
               \If{$f^j_r$ OR $f^j_v$}
                  \State $f^i_a \gets 0$
                  \State \textbf{break}
               \EndIf

               \State $j \gets j + 1$
              \State \textbf{continue}
          \EndIf

          \State Append $v_j$ to $v_i$
          \State Fast implicit column algorithm to update $\delta_*^{v_i}$

          \If{$\delta^{v_i}_*$ is Empty}
              \State $f^i_e \gets 1$
              \State \textbf{break}
          \EndIf

          \If{$\exists t' \text{ s.t. } (\delta^{v_i}_*, t')$ is a trivial pair}
              \State $f^i_v \gets 1$
              \State $f^i_r \gets 0$
              \State $f^i_a \gets 0$
              \State \textbf{break}
          \EndIf

          \If{$(\delta^{v_i}_*, t')$ is a persistence pair in $p^\bot$}
              \State $f^i_v \gets 0$
              \State $f^i_r \gets 1$
              \State $f^i_a \gets 0$
              \State \textbf{break}
          \EndIf

          \State $j \gets 1$

      \EndWhile

    \EndFor

  \end{algorithmic}
  \caption{Serial reduction}
  \label{alg:serialreduce}
\end{algorithm}

\begin{algorithm}
  \begin{algorithmic}[1]

    \For{$(t, v_i, \delta^{v_i}_*, f_v, f_r, f_a, f_e)$ in $\mathbf{v}$}

        \If{$f_e = 0$}
            \State Non-contractible cycle born at $t$
            \State Remove $(t, v_i, \delta^{v_i}_*, f_v, f_r, f_a, f_e)$ from $\mathbf{v}$
            \State \textbf{continue}
        \EndIf

        \If{$f_a = 1$}

            \State Store $(\delta^{v_i}_*, t)$ in $p^\bot(t)$

            \State Define $T \gets [t_k \text{ in } v_i = [(t_k, i_a, i_b, i_c, f, \delta^{t_k}_*)]
            \text{ with } t_k \neq t]$

            \State Remove triangles from $T$ that occur even number of times (modulo 2 sum)
            \State $V^\bot(t) \gets T$
            \State Remove $(t, v_i, \delta^{v_i}_*, f_v, f_r, f_a, f_e)$ from $\mathbf{v}$
            \EndIf

    \EndFor

  \end{algorithmic}
  \caption{Clearance}
  \label{alg:updateV}
\end{algorithm}

\newpage
\section{Computation and Benchmarks}\label{app:tab_benchmarks}

\textbf{Base memory used by Dory} \\
$F_1: $ (two vertices per edge) $+$ (pointer to every edge) $ +$ (length of every edge)
$= 2n_e + n_e + n_e = 4n_e.$\\
Lengths of neighborhoods: one for every vertex  $= n.$\\
Vertex-neighborhood:(every edge is in two neighborhoods)*(neighbor and order are
stored) + (pointer for every vertex)  $= (2n_e)*2 + n$  $ = 4n_e + n.$\\
Edge-neighborhood $=$ Vertex-Neighborhood $=4n_e + n.$\\

\textbf{Fast implicit col and implicit row algorithm} \\

\begin{table}
\begin{center}
  \resizebox{0.5\textwidth}{!}{
  \begin{tabular}{|c|c|c|} 
 \hline
    Data set    &Fast Imp. col. & Imp. row \\
 \hline
    dragon    &  ( 2.859 s, 262 MB )  & (2.887 s, 270 MB ) \\
   \hline
    fractal   &  ( 32.1 s, 695 MB)  & (29.848 s, 670 MB ) \\
   \hline
    o3        &  ( 4.989 s,  157 MB) & ( 22.7 s, 140 MB ) \\
   \hline
    torus4(1) &  ( 9.4 s, 328 MB) & ( 44.47 s, 295 MB ) \\
   \hline
    torus4(2) &  (41.2 s,  1 GB)  & ( 74.6 s, 1 GB  )  \\
   \hline
    Hi-C (control) &(263 s, 6.23 GB)  & ( 595 s, 7.19 GB )    \\
   \hline
    Hi-C (Auxin) (102 s, 3.98 GB  ) &  ( 123 s,  3.83 GB ) \\
   \hline

  \end{tabular}}

\end{center}

  \caption{Comparing fast implicit column algorithm (Dory compiled with \texttt{-D SAVEPD}) and
  implicit row algorithm (both serial-parallel over 4 threads). Time taken (seconds) was measured
  using the command \texttt{time} in terminal of macOS. Peak memory usage is recorded using
  application Instruments in macOS.}

\label{tab:imp_row_vs_col}
\end{table}

\textbf{Gudhi and Eirene}\\ The peak memory usage by Gudhi (v.\ 3.4.0)was recorded using the package
\texttt{memory-profiler} (v.\ 0.57.0) in Python (v.\ 3.8.5), and the memory used by Eirene (v.\
1.3.5) was estimated by observing \texttt{activity monitor} of macOS because the memory profiling
tools in Julia (v.\ 1.5) report cumulative memory allocations instead of the net memory in use.

\begin{table}
\begin{center}
  \resizebox{0.6\textwidth}{!}{
  \begin{tabular}{|c|c|c|c|c|c|} 
 \hline
    Data set  &$n$ &$\tau_m$  & $d$ &Gudhi&Eirene \\
 \hline
    dragon	& 2000	&  $\infty$	     &  1& NA &	7.4 GB\\
   \hline
    fractal	& 512	  &  $\infty$	      &  2& NA &	9.13 GB\\
   \hline
    o3 & 8192	&  1	     &  2& 2.3 GB &	9.34 GB	\\
   \hline
    torus4(1)	& 50000	&  0.15  & 	1& 3 GB & 	126 GB \\
   \hline
    torus4(2)	& 50000	&  0.15  & 	2& 30 GB&	NA  \\
   \hline

  \end{tabular}}
\end{center}
  \caption{Benchmarks for Gudhi and Eirene. NA means that the data set was not processed within 10
  mins. or the system ran out of memory.}
  \label{tab:gudhi_eirene_benchmarks}
\end{table}

\newpage
\section{Persistence Diagrams for Data Sets}\label{app:PDs}

\begin{figure}[tbhp]
  \centering
  \begin{subfigure}{.48\textwidth}
    \includegraphics[width=\linewidth]{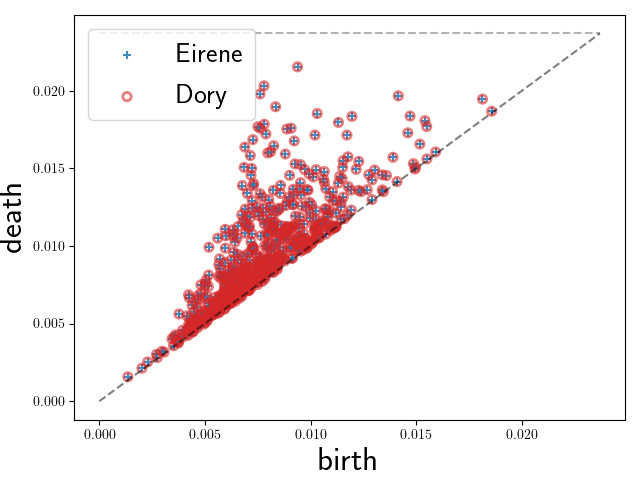}  
    \caption{Eirene}
    \label{fig:drag_eirene}
  \end{subfigure}
  \centering
  \begin{subfigure}{.48\textwidth}
    \includegraphics[width=\linewidth]{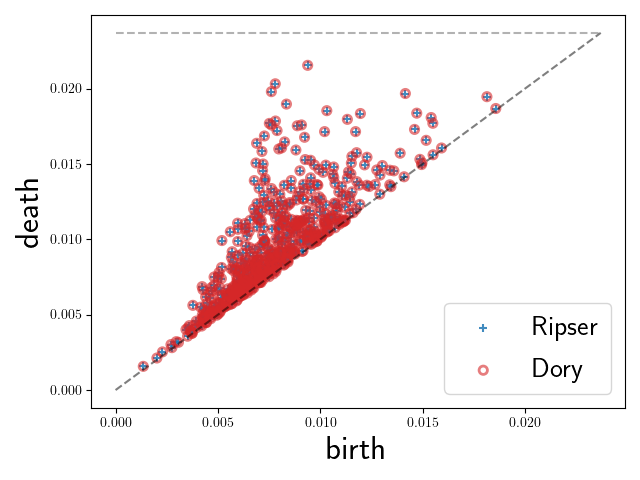}  
    \caption{Ripser}
    \label{fig:drag_ripser}
  \end{subfigure}
    \caption{Dragon H$_1$ PD}
    \label{fig:dragon_PD}
\end{figure}

\begin{figure}[tbhp]
  \centering
  \begin{subfigure}{.48\textwidth}
    \includegraphics[width=\linewidth]{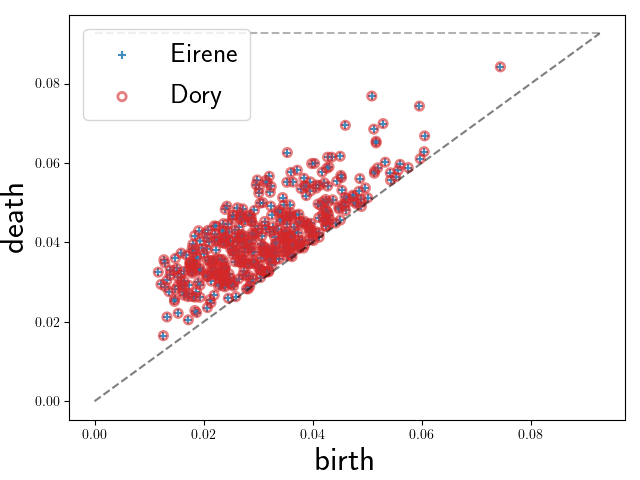}  
    \caption{Eirene}
    \label{fig:fracth1_eirene}
  \end{subfigure}
  \centering
  \begin{subfigure}{.48\textwidth}
    \includegraphics[width=\linewidth]{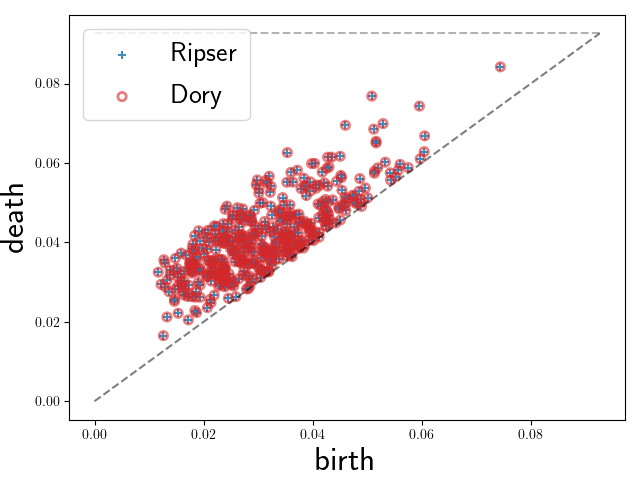}  
    \caption{Ripser}
    \label{fig:fracth1_ripser}
  \end{subfigure}
    \caption{Fractal H$_1$ PD}
    \label{fig:fracth1_PD}
\end{figure}

\begin{figure}[tbhp]
  \centering
  \begin{subfigure}{.48\textwidth}
    \includegraphics[width=\linewidth]{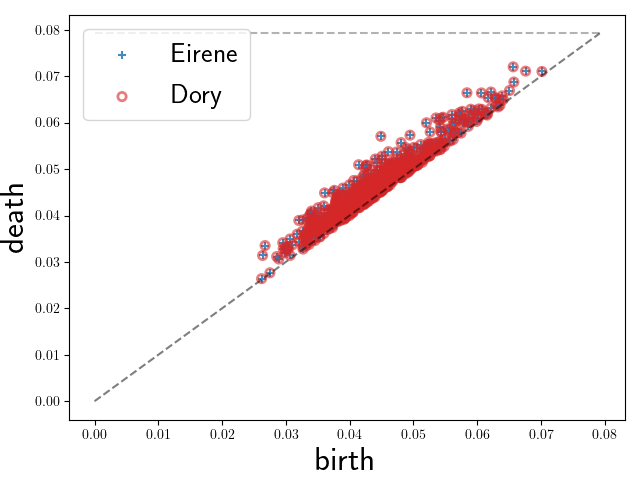}  
    \caption{Eirene}
    \label{fig:fracth2_eirene}
  \end{subfigure}
  \centering
  \begin{subfigure}{.48\textwidth}
    \includegraphics[width=\linewidth]{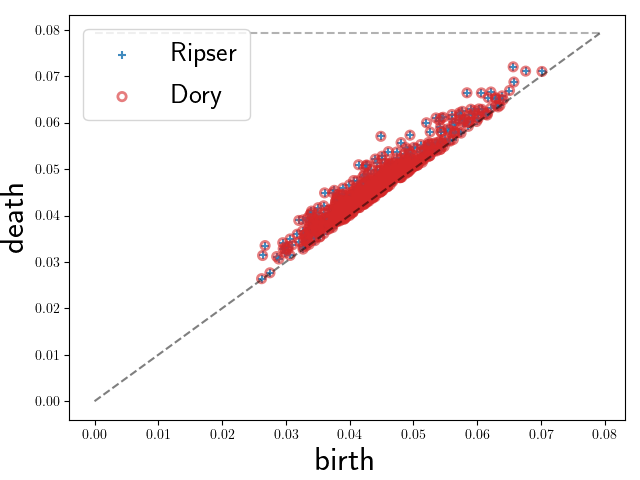}  
    \caption{Ripser}
    \label{fig:fracth2_ripser}
  \end{subfigure}
    \caption{Fractal H$_2$ PD}
    \label{fig:fracth2_PD}
\end{figure}

\begin{figure}[tbhp]
  \centering
  \begin{subfigure}{.3\textwidth}
    \includegraphics[width=\linewidth]{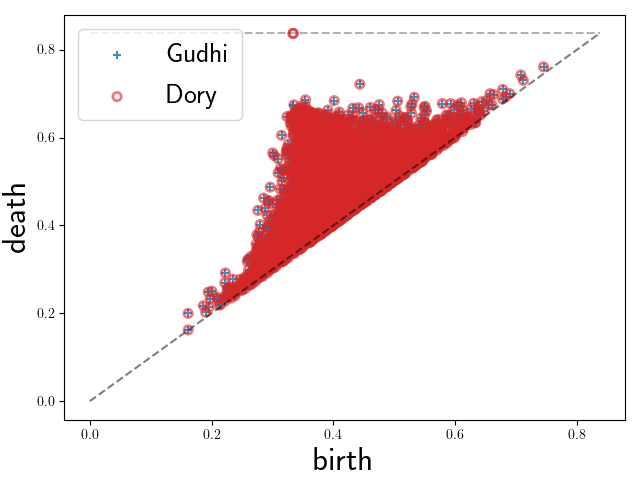}  
    \caption{Gudhi}
    \label{fig:o3_H1_gudhi}
  \end{subfigure}
  \centering
  \begin{subfigure}{.3\textwidth}
    \includegraphics[width=\linewidth]{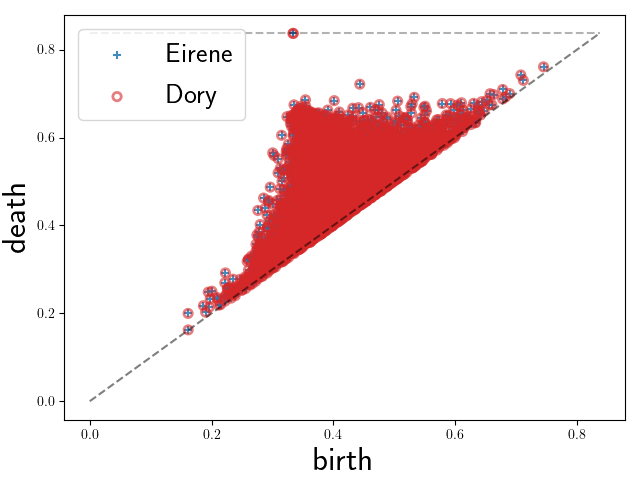}  
    \caption{Eirene}
    \label{fig:o3_H1_eirene}
  \end{subfigure}
  \centering
  \begin{subfigure}{.3\textwidth}
    \includegraphics[width=\linewidth]{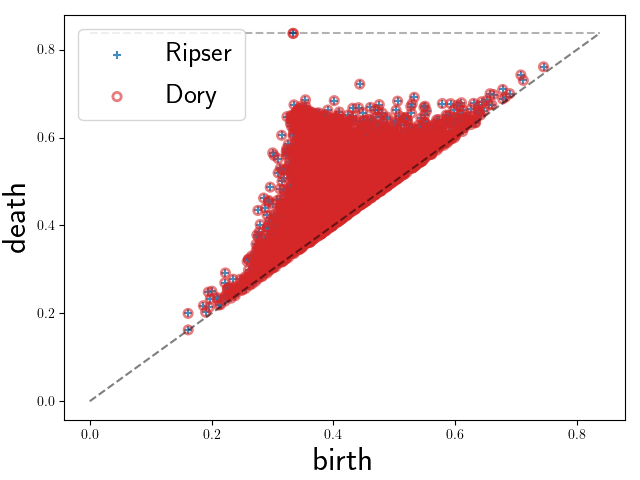}  
    \caption{Ripser}
    \label{fig:o3_H1_ripser}
  \end{subfigure}
    \caption{o3 H$_1$ PD}
    \label{fig:o3_H1_PD}
\end{figure}

\begin{figure}[tbhp]
  \centering
  \begin{subfigure}{.3\textwidth}
    \includegraphics[width=\linewidth]{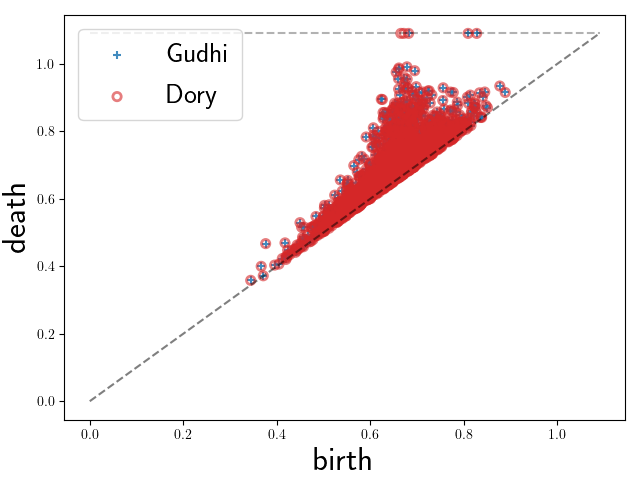}  
    \caption{Gudhi}
    \label{fig:o3_H2_gudhi}
  \end{subfigure}
  \centering
  \begin{subfigure}{.3\textwidth}
    \includegraphics[width=\linewidth]{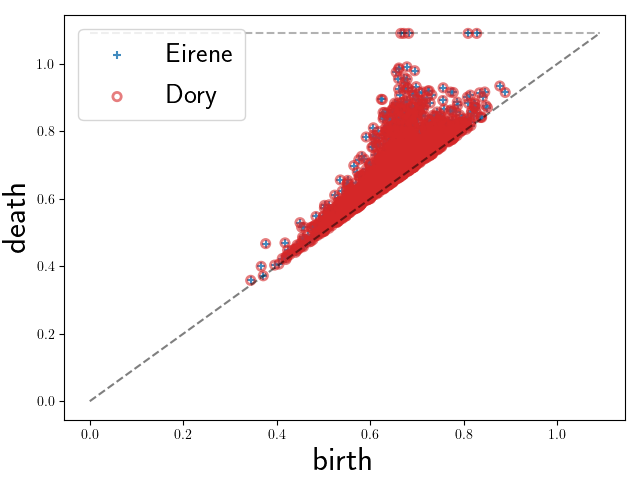}  
    \caption{Eirene}
    \label{fig:o3_H2_eirene}
  \end{subfigure}
  \centering
  \begin{subfigure}{.3\textwidth}
    \includegraphics[width=\linewidth]{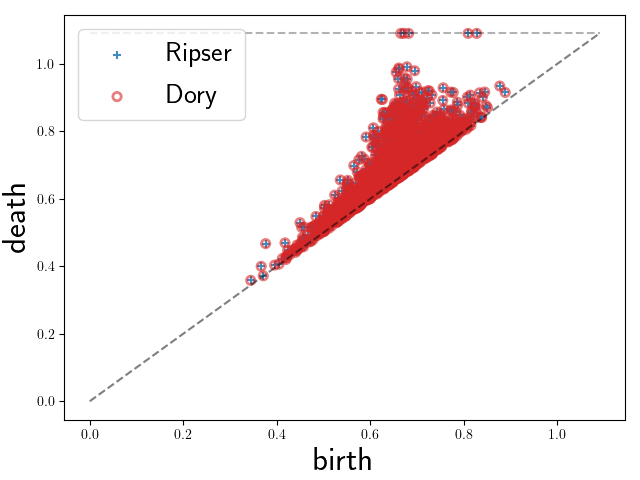}  
    \caption{Ripser}
    \label{fig:o3_H2_ripser}
  \end{subfigure}
    \caption{o3 H$_2$ PD}
    \label{fig:o3_21_PD}
\end{figure}

\begin{figure}[tbhp]
  \centering
  \begin{subfigure}{.3\textwidth}
    \includegraphics[width=\linewidth]{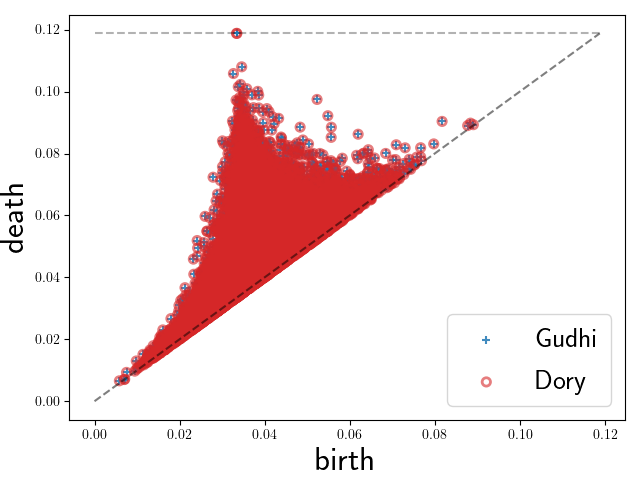}  
    \caption{Gudhi}
    \label{fig:torus_H1_gudhi}
  \end{subfigure}
  \centering
  \begin{subfigure}{.3\textwidth}
    \includegraphics[width=\linewidth]{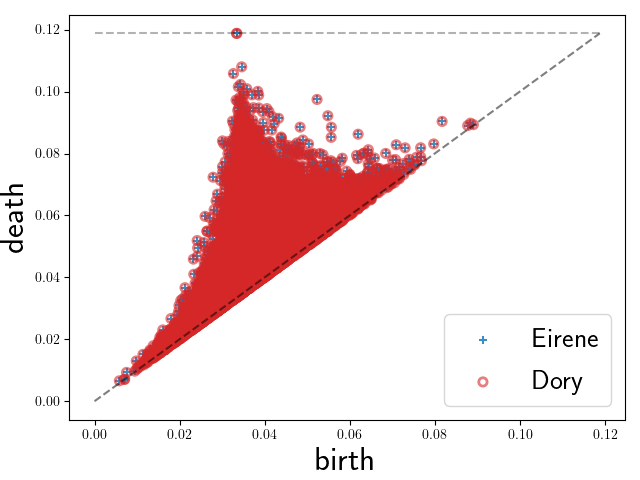}  
    \caption{Eirene}
    \label{fig:torus_H1_eirene}
  \end{subfigure}
  \centering
  \begin{subfigure}{.3\textwidth}
    \includegraphics[width=\linewidth]{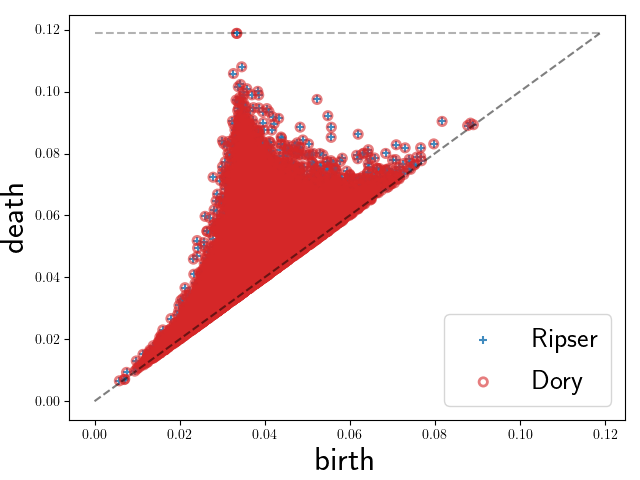}  
    \caption{Ripser}
    \label{fig:torus_H1_ripser}
  \end{subfigure}
    \caption{torus H$_2$ PD}
    \label{fig:torus_H1_PD}
\end{figure}

\begin{figure}[tbhp]
  \centering
  \begin{subfigure}{.45\textwidth}
    \includegraphics[width=\linewidth]{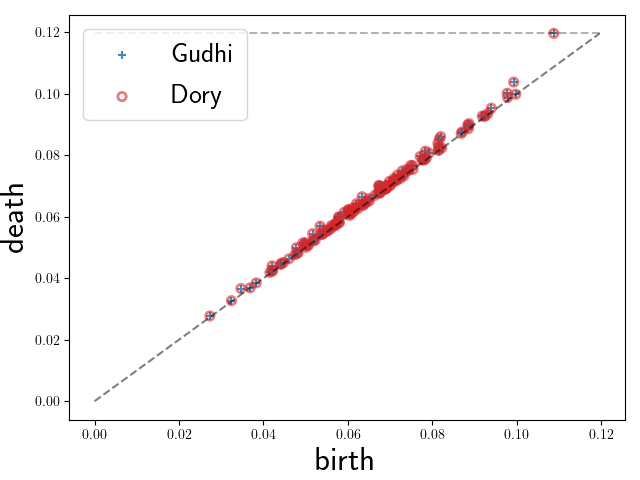}  
    \caption{Gudhi}
    \label{fig:torus_H2_gudhi}
  \end{subfigure}
  \centering
  \begin{subfigure}{.45\textwidth}
    \includegraphics[width=\linewidth]{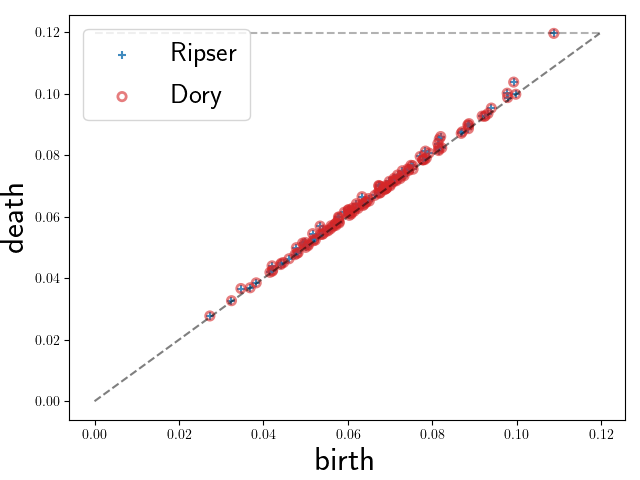}  
    \caption{Ripser}
    \label{fig:torus_H2_ripser}
  \end{subfigure}
    \caption{torus H$_2$ PD}
    \label{fig:torus_H2_PD}
\end{figure}

\newpage
\section{Persistence Diagrams for Hi-C}\label{app:pd_hic}

\begin{figure}[tbhp]
  \centering
\begin{subfigure}{.48\textwidth}
  \centering
  \includegraphics[width=\linewidth]{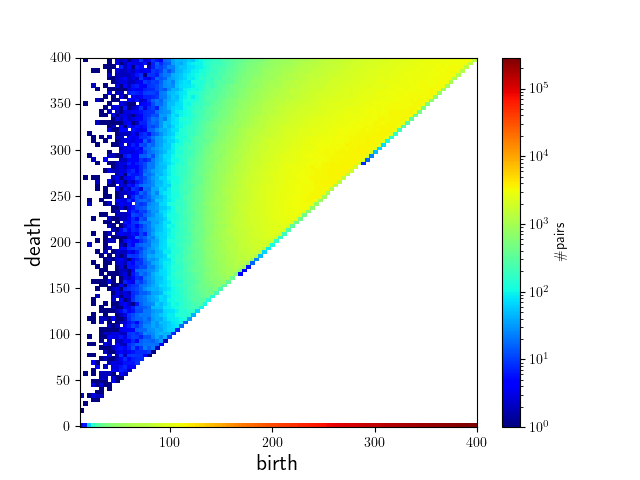}  
  \caption{Control}
  \label{fig:HiC_control_pd1}
\end{subfigure}
  \centering
\begin{subfigure}{.48\textwidth}
  \centering
  \includegraphics[width=\linewidth]{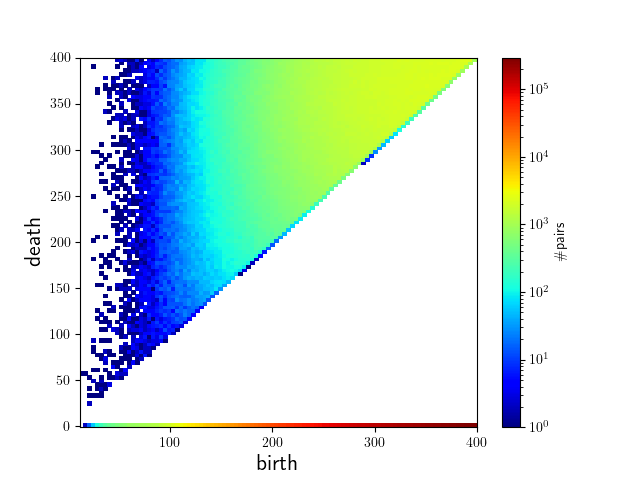}  
  \caption{Auxin}
  \label{fig:HiC_aux_pd1}
\end{subfigure}
  \caption{Hi-C data sets H$_1$ PD}
\label{fig:HiC_PD}
\end{figure}

\begin{figure}[tbhp]
  \centering
\begin{subfigure}{.48\textwidth}
  \centering
  \includegraphics[width=\linewidth]{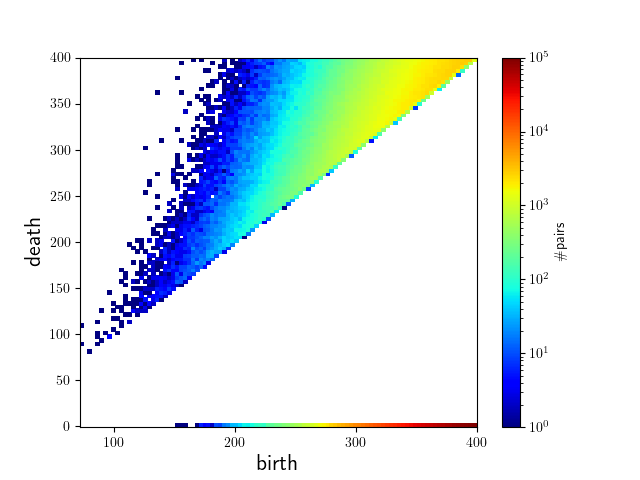}  
  \caption{Control}
  \label{fig:HiC_control_pd2}
\end{subfigure}
  \centering
\begin{subfigure}{.48\textwidth}
  \centering
  \includegraphics[width=\linewidth]{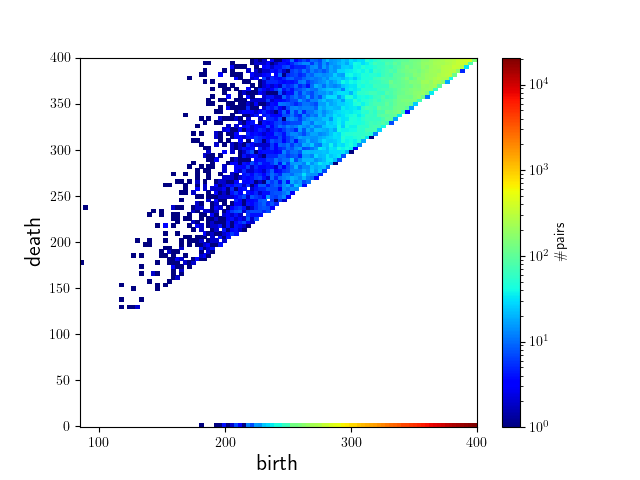}  
  \caption{Auxin}
  \label{fig:HiC_aux_pd2}
\end{subfigure}
  \caption{Hi-C data sets H$_2$ PD}
\label{fig:HiC_aux_PD}
\end{figure}

\vskip 0.2in
\end{appendices}

\newpage
\bibliographystyle{unsrtnat}
\bibliography{PH_refs}

\end{document}